\documentclass[twoside,11pt]{article}
\usepackage{jair,rawfonts}
\usepackage[apaciteclassic,nodoi]{apacite}

\usepackage{microtype}
\usepackage{graphicx}
\usepackage{subfigure}
\usepackage{booktabs} 
\usepackage[implicit=false]{hyperref}

\usepackage{array}
\usepackage{wrapfig}
\usepackage{mathrsfs}
\usepackage{tabu}
\usepackage{mathtools}
\usepackage{amsmath}
\usepackage{amsthm}
\usepackage{bm}
\usepackage{amssymb}
\usepackage{cancel}
\usepackage{bbm,dsfont}
\usepackage{soul}
\usepackage{tikz-cd}
\usepackage{quiver}

\usepackage{multicol, blindtext}
\usepackage[algo2e,linesnumbered,ruled,lined]{algorithm2e}

\usepackage[utf8]{inputenc}     
\usepackage[english]{babel}    
\usepackage{tikz}
\usetikzlibrary{cd}
\usepackage{graphicx}
\newcolumntype{L}{>{$}l<{$}}
\newcommand{\dg}{\textsl g}
\newcommand{\br}{\mathbf{r}}

\newcommand{\ba}{\boldsymbol{\alpha}}

\DeclareMathOperator*{\argmax}{argmax}

\DeclareMathOperator*{\argmin}{argmin}
\newcommand\mydots{\hbox to 1em{.\hss.\hss.\hss}}

\usepackage{xcolor}

\newcommand{\mc}{\mathcal}
\DeclarePairedDelimiterX{\infdivx}[2]{(}{)}{%
  #1\;\delimsize\|\;#2%
}

\DeclareMathOperator*{\expec}{\mathop{\mathbb{E}}}
\usepackage{float}

\newcommand{\xoverbrace}[2][\vphantom{\dfrac{A}{A}}]{\overbrace{#1#2}}
\newcommand{\xunderbrace}[2][\vphantom{\dfrac{A}{A}}]{\underbrace{#1#2}}
\usepackage{graphicx}
\usepackage{tcolorbox}
\tcbuselibrary{theorems}

\newtcbtheorem[number within=section]{mydef}{Definition}%
{colback=blue!5,colframe=blue!35!black,fonttitle=\bfseries}{th}

\newtcbtheorem[number within=section]{mytheo}{Theorem}%
{colback=green!5,colframe=green!35!black,fonttitle=\bfseries}{th}

\jairheading{x}{x}{1-15}{x/xx}{x/xx}
\ShortHeadings{Reward is not Necessary}
{Ringstrom}
\firstpageno{1}

\begin{document}
\title{Reward is not Necessary: How to Create a Modular \& Compositional Self-Preserving Agent for Life-Long Learning}


\author{\name Thomas J. Ringstrom \email rings034@umn.edu \\
\addr Department of Computer Science, University of Minnesota,\\
Minneapolis Minnesota, USA, 55455}

\maketitle

\begin{abstract}
    The theoretical tradition of Reinforcement Learning (RL) views the maximization of rewards and avoidance of punishments or costs as central to explaining motivated goal-directed behavior.  However, over the course of a life, organisms, as agents, will need incorporate knowledge about many different aspects of the world's structure, what are called the \textit{states} of the world and the \textit{state-vector transition dynamics}---how a collection of states, organized as a vector, can change over time. The number of combinations of possible states of the world grows exponentially as an agent incorporates new knowledge of states. This creates an epistemic problem for life-long agents: there is no obvious weighted combination of pre-existing rewards or costs defined for a given \textit{combination} of states, because such a weighting would need to encode information about all good and bad combinations prior to an agent's experience in the world. Traditional RL does not accommodate this complexity, and so we must develop more naturalistic accounts of behavior and motivation in large state-spaces. We show that it is possible to use only the intrinsic motivation metric of \textit{empowerment}, which is a function of the agent's state-vector and possible transition dynamics, and measures the agent's capacity to realize many possible futures. We show how to scale empowerment to large hierarchical spaces of many individual state-spaces. To achieve this, we propose using Operator Bellman Equations. These reward-free equations produce \textit{state-time feasibility functions}, which are abstract and compositional hierarchical state-time transition operators. This means that the feasibility functions map the initial state and time when an agent begins a course of action to the final states and times of completing a goal as a result of the actions, and feasibility functions for single time-dependent sub-goals can be sequentially \textit{composed} to complete multiple sub-goals in a hierarchical state-space. Because feasibility functions form hierarchical transition operators, we can define hierarchical empowerment measures on them. An agent can then optimize plans to distant states and times to maximize its hierarchical \textit{empowerment gain} (i.e. ``valence") by changing the state, transition structure, and affordances of the world. This optimization allows an agent to \textit{interpret} which state-vectors, as goals, are more favorable to the coupling of its internal structure (e.g. hunger, hydration \& temperature states, and skills) to its external environment (e.g. world structure, spatial state, and items). Embodied life-long agents could therefore be primarily animated by principles of compositionality and empowerment for planning and goal setting, exhibiting self-concern for the growth and maintenance of their own structural integrity without recourse to reward-maximization. 
\end{abstract}

\newpage
\section*{Author's Note}
(Sept. 2023): This manuscript is going under review and has been updated from a previous version posted (Nov. 2022). The sequence of arguments made in the introduction have been rearranged and expanded by about a couple extra pages to discuss compositionality (along with a new figure (\ref{fig:composition})), The Reward Hypothesis, Reward-is-Enough, Active Inference and preference formation in more detail, in addition to some more references. The technical content of the document is the same, and \textit{nothing significant has been removed} (technical or discussion). However, I rearranged some of the content in the Self-preserving Agent chapter 5. As it is now, I first introduce the TG-CMDP planning factorization, followed by the bi-directional TG-CMDP planning factorization theorem, then followed by the subsection on computing the aggregate feasibility function. The previous version included all of these steps into one section, which risked confusing the reader with too many details simultaneously. The TG-CMDP decomposition theorem and proof has been improved for clarity and the theorem has been added to the main text, which should make things easier to understand. The manuscript also contains an experimental section that compares methods. The document has been reformatted, extra figures have been added for better comprehension, and some grammatical and notational errors have been fixed. I have improved some notation, for example: high-level actions (goals) used to be simply $\dg$, but is now written as $\alpha_{\dg}$, where $\dg$ is now an index appended to an action $\alpha$. There is a small new subsection on how to compute the value of an \textit{item} (such as a key). I also added some visual influence diagrams to illustrate transition operator compositions  corresponding to the equations. I will likely not make additional changes before this is published, but should I do so, I will post the notes here.

\tableofcontents
\newpage
\textit{This paper is organized into two parts: an introduction and high-level overview (Chapters \ref{sec:intro} \& \ref{sec: Motivation}), and technical sections (Chapters \ref{sec:Empowerment} to \ref{chapter:lifelong}). In the first part, we aim to provide a high-level overview of the motivation and technical content of the paper, to theorists and non-specialists alike.}
\section{Introduction}\label{sec:intro}
Stoffel the Honey Badger was the star of a 2014 PBS documentary called \href{https://www.youtube.com/watch?v=c36UNSoJenI}{\textit{Honey Badgers: Masters of Mayhem}}, in which he is shown performing impressive escape routines from his pen, Badger Alcatraz \cite{mayhem}. This was all to the astonishment and annoyance of the caretaker Brian, who constantly had to remove the items and resources that Stoffel used to open gates and jump over walls. If there was a tree in the middle of the pen, Stoffel would climb up it and sway it in the direction of the wall to coordinate an escape. Remove the tree and Stoffel would find novel objects like a branch or a rake, or he would unearth stones to position next to the wall to climb up. And if those were taken away, Stoffel would pack mud into balls and stack them into a climbable pyramid. What else can honey badgers do? If there is food in a box, they can move objects under it to climb up close enough to reach it \cite{fearless}; and, if there is a gate with a latch, Stoffel and his girlfriend Hammie can coordinate to undo the latch mechanism and open the door. Not only do honey badgers complete these tasks with clever reasoning, they do so potentially with a variety of possible motivations: for satisfying hunger, or for expanding the capacity to move into new external territory, or perhaps, much more speculatively, for the pleasure of \textit{trolling} Brian by intentionally acting in a way that defies his preferences. It is not too hard to imagine the later, after all, Stoffel can be heard letting out what appears to be a sound of satisfaction when he escapes down the side of the wall and runs away from an employee; Brian claims that Stoffel resented his presence and treated all of his challenges as a game.

How does Stoffel achieve these tasks? When solving the problems, the honey badger is presumably performing complex sub-goals to manipulate key abstract variables to \textit{change the allowable dynamics of the world} in order to solve a task. For example, Stoffel repositions an item from one location to another in order to climb up the wall at a specific location. Or, Stoffel removes the fastening wire, which binds and immobilizes the latch to the side of the fence, in order to slide the latch from the binary state of closed to open. By performing this sub-task Stoffel can then pass through a gate, which is a new mode of possible dynamics. 

An even more perplexing set of question are: what purpose motivated Stoffel to perform the tasks?  Why should Stoffel choose one purpose over another? And what is a purpose? These questions get at some of the core issues at the heart of Artificial Intelligence, the interaction between three interlinked facets of intelligence discussed in this paper: skills, abstractions, and intrinsic motivation. By skills, we mean complex sequences of polices to attain some desired transformation \cite{konidaris2009skill,shu2017hierarchical,konidaris2010constructing}. By abstractions, we mean two things: \textit{abstract transition functions}, which help us simplify low-level details in order to reason quickly at a higher-level of abstraction \cite{ringstrom2020jump}, and \textit{high-level state abstraction} in the form of higher-order state-spaces or automata \cite{icarte2022reward, hasanbeig2020deep} for representing a modular and remappable task \cite{ringstrom2020jump}; for example, a gate latch is a continuous object with many (infinite) physical configurations corresponding to both open and closed, but we can represent those two abstract states as one bit. This bit can be changed from a given low-level state, and ideally for modular agents, the mechanisms and rules of such a task should be able to be \textit{remapped} to the features of many different environments \cite{ringstrom2020jump} (see figure \ref{fig:composition}). Lastly, by intrinsic motivation, we mean that an agent's internal representations are the sole source of its justification for doing tasks \cite{oudeyer2007intrinsic, oudeyer2009intrinsic}, which is also related to an emerging concept in intrinsic motivation is known as \textit{autotelic agency}, the ability for an agent to use its own representations and knowledge to set goals \cite{forestier2017intrinsically,sigaud2021towards,khazatsky2021can,colas2022autotelic,akakzia2022help}. What representations and computations help intelligent organisms autonomously survive in a dynamic and complex world? Is there a mutual dependency and interplay between these facets? And, what optimization principles are at play? 

\begin{figure}[h]
\centering
\includegraphics[width=\linewidth]{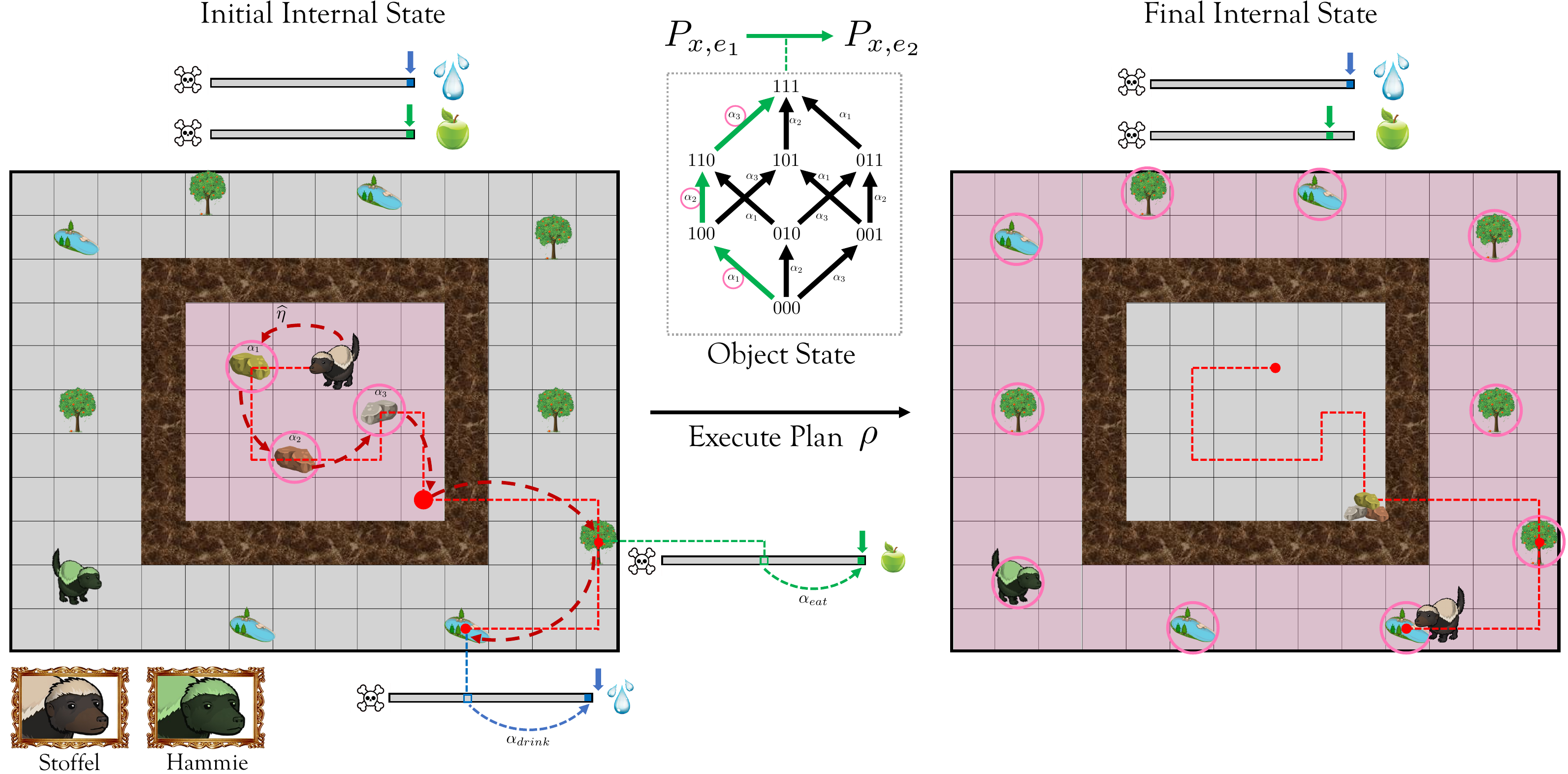}
\caption{Escape from Badger Alcatraz:  Stoffel collects three stones, recorded in a binary vector object space by flipping the bit corresponding to each object when collected. Once the three objects are acquired, Stoffel can place them in the corner, thereby re-parameterizing the low-level transition operator from $P_{x,e_1}$ to $P_{x,e_2}$ by transitioning the indexing variable $e$. The new transition structure enables Stoffel to climb over the wall. Before placing the stones, Stoffel can only reach states (pink squares) and perform tasks (pink circles) on the interior of the pen, but can access more states and tasks (such as eating, drinking, and mating) once he climbs over the wall (the example assumes Stoffel can not climb back in). The difference in reachability before and after the plan $\rho$ (sequence of policies) can be computed as a specific instance of empowerment gain (valence) to justify the semi-Markov plan. Small dashed red lines denote a path from the plan, dark red dashed arcs denote abstract, initial state-time to final state-time transitions under feasibility function ($\widehat{\eta}$) transitions for each of the $6$ policies in the plan, which allows the agent to reason at the level of goals and time. Plans are proposed and evaluated by forward sampling polices under a factorized  product-space operator that avoids representing the otherwise intractable product-space.}
\label{fig:escape}
\end{figure}

In a recent article, Vamplew et al. \citeyear{vamplew2021scalar} termed a speculative set of innate capacities that help guide an agent's behavior the \textit{intellectual phenotype}. To an agent, and to Stoffel, the internal and external world may consist of many coupled state-spaces, such as (but not limited to) hydration, caloric, temperature, and object-possession state-spaces, along with the world's state-space, forming a complex combinatorial space (i.e. a Cartesian product-space) of state-variables which interact with each other in a dynamic and non-stationary manner. For example, Stoffel may need to drink at a specific location in the world to change his hydration state, but a dehydration state in a his physiological state-space, which evolves independently over time, will arrest his dynamics in the world. Or, Stoffel may need to obtain a key in order to enter a different part of the state-space where he can drink water. If Stoffel is an organism that coordinates the expectations of multiple signals affecting each other in a complex manner, what intellectual phenotype might Stoffel have to support computations of this nature? 

Recently, Eppe et al. \citeyear{eppe2022intelligent} have called for the development of a unified architecture based on hierarchical reinforcement learning (RL) \cite{sutton1998introduction}, which utilizes model-based reasoning and compositional abstractions to tackle the breadth of advanced biological problem solving. While new methods are indeed necessary, it is unclear if reward-maximization objectives, such as infinite horizon discounted reward-maximization (IHDR), admit natural factorizations that can be used to solve problems in large spaces of variables in real-time---that is, it might not support the best intellectual phenotype. This should be a cause of reflection for theorists: if modern learning frameworks such as RL are built upon optimization foundations which do not admit a natural decomposition to exploit, then it may be unlikely that the learning framework will scale to the large hierarchical spaces naturally intelligent agents are faced with\footnote{Meta-RL could be made to act as a meta optimization which invents new optimization paradigms that engender flexibility and low-sample complexity, and in such a case reward could be considered \textit{sufficient} in an instrumental sense.}. Given these considerations, it is important that we thoroughly examine the assumptions behind the hypothesis of reward-maximization as a theory of rational action and assess its limitations, as we will now proceed to do.

\section{Overview of Motivation and Theory}\label{sec: Motivation}

\subsection{A Critique of Reward-Maximization and the Reward is Enough Hypothesis}

A theoretical position taken in the field of RL, recently proposed by Silver et al., is that scalar reward-maximization is \textit{sufficient} as a means of generating purpose driven behavior characteristic of general intelligence \cite{silver2021reward}. This position can be summarized by the Reward Is Enough Hypothesis (RIEH):
\begin{quote}
    ``Intelligence, and its associated abilities, can be understood as subserving the maximisation of reward by an agent acting in its environment."
\end{quote}
RIEH takes the stance that all attendant behaviors and abilities exhibited by intelligent systems (such as model-based planning and representation learning) are acting in the service of accumulating reward signals, even if there are many different kinds of reward signals in the world.

However, in their article, Vamplew et al. \citeyear{vamplew2021scalar} critique RIEH arguing that it cannot account for the breadth of real-world problems, which are fundamentally multi-objective in nature from moral reasoning down to physiological regulation.  Of the later they state, ``In mammals, as in most other organisms, the bio-computational processes that constitute that intellectual phenotype have no single objective; rather, they include multiple including hunger satiation, thirst quenching, social bonding, and sexual contact", and they contend that these factors are intrinsically multi-objective \cite{vamplew2021scalar}. Their critique of RIEH ranges over a number of arguments, centering on the point that training polices on a single scalar reward limits the space of good polices under a change in an agent's priorities in multi-objective RL (MORL). 

In this paper, we take the position that maximizing a scalar is indeed sensible (this is the essence of optimization after all), but we challenge the assumption of RIEH that agents need to act to accumulate reward, at least insofar as it is a signal ``received" at a given state which needs to be represented and stored as an \textit{accumulated quantity}.  Of course, one can always make the case that this is simply a semantic preference and that all positive scalars that are maximised could be called ``reward." But, we argue against this case, as we do not find it appropriate for the two Bellman equations we will introduce in this paper. Both of the Bellman equations will have analogues of the reward function that are \textit{derived} from the structure of the agent's control architecture itself, and not \textit{received} from some other source. Just because a given optimized quantity is a scalar doesn't mean that it should be considered a reward; semantic distinctions can take into account how and when the scalar is derived and how it is used. Semantically, RIEH also regards reward-maximization as being \textit{the goal}, where sub-goals are anything that support this objective, such as dynamics learning or value prediction. This is in contrast to a goal being (for instance) a variable representing state-action attainment, which we argue for in this paper---we take the view that goals are more primary. These semantic distinctions have practical consequences for how we interpret objective functions and organize concepts. 

Sutton and Barto's \citeyear{sutton1998introduction} \textit{Reward Hypothesis} (RH) is related to RIEH and formalizes \textit{goals and purposes}: ``all of what we mean by goals and purposes can be expressed as the accumulation of a scalar reward function." RH is about the \textit{expression} of ``goals and purposes" as reward accumulation, whereas RIEH is a hypothesis about the sufficiency of reward-maximization to realize a theory of general intelligence. Recently, Bowling et al. \citeyear{bowling2022settling} have given the precise conditions in which RH is true. The argument is that if one \textit{defines} ``goals and purposes" to be a preference relation on trajectory histories, then reward-maximization leads to policies which satisfy a set of Von Neumann-Morgenstern (vNM) axioms on these preference relations, and RH will hold. Therefore, reward signals effectively contribute to a preference relation, which is not necessarily \textit{represented} by an agent.
If we accept this argument (as we do), and also accept RIEH, we are led to believe that the purpose of an intelligent organism is to maximize a received signal which \textit{encodes} the preference relation that constitutes, as they call it, the organism's ``subjective purpose".  In RIEH, the reward signal is the primary interest of the agent. If a reward signal is a proxy for a subjective purpose and the reward signal is what is used to \textit{learn}, then nothing an intelligent agent has done or will do via \textit{reasoning} contributes to the formation of the preference relation. This is because RIEH asks us to believe that all of the ancillary intelligent activity of an agent, whether it be the formation of abstractions and representations, or model-based planning or reasoning, is done \textit{in the service} of obtaining reward. RIEH does not explicitly make statements about the necessity of respecting vNM axioms, and acknowledges that multiple reward-signals could be relevant to maximize in complex environments. However, if the agent were to act in a way that explicitly violates a given preference relation on trajectory histories, or develop a new affinity for a specific kind of reward, what then would instigate the formation of a \textit{new} preference? \textbf{We argue that the ability to rationally choose which signals to respond to and when to act against established preferences is a central mechanism of intelligence.} RIEH is agnostic to this question because if an agent were to form new preferences, the mechanism for this capability would be contributing to maximizing reward regardless of the answer. But, this hypothetical mechanism \textit{would be an ancillary mechanism of reward maximization}, and thus requires an explanation. By the logic of RIEH, the mechanism in question would have to be shaped by reward-maximization itself. There are two obvious possibilities that could shape this mechanism: the self-consistent (but unsatisfying) explanation that there is a more universal and prioritized reward signal that guides the process (a proposal that would shift the burden of explanation to a new level of abstraction), or the explanation that the mechanism of preference formation is shaped by evolution to collect more reward in order to avoid death (where evolution would need to produce a mechanism general enough for humans to care about abstractions removed from physiological reality). Either explanation would have to produce a mechanism that would be able to \textit{interpret} why any new reward signal for some new abstraction is \textit{salient} to the agent during life-long learning, which is challenging given that abstract states cannot be anticipated before they are created as part of an agent's model of the world. 
 

The assumption that a signal corresponding to a choice ``carries" or ``communicates" value has been critiqued in the psychological decision sciences by Srivistava and Schrater \citeyear{srivastava2015learning}. They point out that expected utility theory does not account for the emergence of new preferences and that ``existing formal models of preference formation end up assuming what they are supposed to generate---that value ‘signals’ associated with an option pre-exist in the environment, and that the goal of a theory of preference formation is to specify how to efficiently separate out these signals from ambient noise introduced by probability." They further argue, ``if how much something is valued is already a signal though, then a person’s experience cannot have a role in shaping it, which contradicts the primary role that psychological theories assign to experience."  To account for the emergence of new preferences, their solution is to assume a latent \textit{acceptability function} which returns whether or not a choice was acceptable \textit{to the agent}. In doing so, they are able to show that an agent can perform Bayesian inference over memories of choice acceptability and context and infer the relative desirability of options to make choices without recourse to economic judgements and hedonic utility maximization principles that rely on a ``common currency," and they also replicate well-known irrational choice biases in psychology.
The main insight relevant to our work is that new preferences can simply be accounted for by the information processing mechanisms of an organism, where the acceptability function can be thought of as a function of the agent itself, indicating what is acceptable \textit{to the agent}, an insight which is one of our main inspirations. This is in sharp contrast to RL where the reward signal or cost is a normative representation whose content is \textit{pre-interpreted} as good or bad independent of any knowledge about the effect it has on the agent. 

To bring these intuitions into the realm of embodied planning, one must have a measure of the impact of a choice on an agent's \textit{ontology}, where for convenience we define ontology to be the agent's entire hierarchical planning transition operators. We propose an optimization framework for an autotelic agent called the \textit{self-preserving agent} (SPA) that addresses the problem of autonomously setting goals, where SPA has an intrinsically self-undermining ontology with states that can kill the agent (permanently immobilizing it). AI agents that play GO or Chess may develop representations that represent important processes and dynamics relevant to a task, but they do not have a core ontology that efferent processes depend on, and they do not develop the internal representations of the processes that \textit{sustain} it, such as the state of the energy source used to power the volatile memory and GPU computations required to represent the agent and to generate plans. However, an animal might form internal representations of an interoceptive domain representing energetic processes that sustain its functioning, an ontology to which symbolic content can be grounded. Sims \citeyear{sims2022self} has recently called for the creation artificial agents that exhibit \textit{self-concern} by anticipating the needs of these sustaining processes. These themes have also been extensively discussed within the intellectual traditions of enactivism and embodiment that emphasize the precarious coupling of an agent to its umwelt \cite{varela2017embodied, varela1979principles, froese2023autopoiesis}, in addition to the Predictive Processing research that emphasizes physiological regulation, including the Free Energy Principle \cite{seth2016active, allen2018cognitivism, ramstead2020tale, seth2021being}. While we draw inspiration from these traditions, we depart from them by centering the controllability of agent's product-space as a fundamental objective rather than sensory prediction. Whereas RIEH asks us to believe that model-based planning and reasoning is something that occurs in order to accumulate more reward by \textit{occupying} states where the agent \textit{receives} a reward signal, we are arguing that model-based planning and reasoning can \textit{generate} actionable value signals---\textit{valence}, as it will be called---when applied to an agent's ontology to assess its integrity. When valence is optimized, it results in an agent which exhibits self-concern. Unlike reward, valence is never ``received" in transit or at the terminal state, rather, it is derived as a computational process on an agent's ontology in order to \textit{initiate} action, instigated by potential future changes in its hierarchical structure. Our agent will always be lead by an understanding of the future, anticipating how a long course of action can improve its internal organization\footnote{As the Canadian phenom Drake once rapped: ``Started from the bottom now we're here!" One could argue that his song evokes the valence arising from the change in circumstances of his life.}.

``Reward," as the word is typically used in everyday life, often refers to a \textit{change in state or capability}, whether it be a change in wealth, a gifted object, or the granting of a privilege. Reward maximization in RL, however, is an objective that produces accumulated quantities of reward in the value function that is fundamentally non-fungible\footnote{Fungibility can be thought of as a property of how a state-space (e.g. monetary states) operates with other state-spaces, however we will not address this in this paper.} nor exchangeable---one does not \textit{spend} accumulated reward as if it were a state of the world. In practice, information about \textit{where} and \textit{when} something is rewarded is lost when rewards are aggregated as discount-weighted quantities into the value function---you cannot extract this information from it once rewards are accumulated.  However, as we will show, information such as the location and timing of an object availability is critical for compositional planning. For example, a reward-function for a dollar bill could encode information about when and where it is located in space, and the act of picking up a dollar bill could increment a wealth state, which is typically not represented by reward functions; however, we need all of this information for composing solutions to sequential tasks. However, value functions for individual problems are not very useful for sequential tasks because they do not represent a state of the world and therefore they cannot condition policies \textit{as variables}, and this means that they essentially play a \textit{passive} normative role to the agent and not a functional one, they do not serve as reusable and remappable representations that can be used across a variety of tasks in the future. In this paper, our objective functions preserve the state, time, and goal availability information in the representations that are optimized, and following an optimized policy results in final state-vectors that represent states, such as object-possession states for a key, that could be used, modified, or exchanged in subsequent tasks. This is also important because, when combined with an intrinsic motivation metric, agents can \textit{bestow} value to actions or obtained objects in the world (like a key) based on how the actions or object affects the agent's ontology (discussed in section \ref{sec:item value}), similar to work by Kolchinsky and Wolpert \citeyear{kolchinsky2018semantic} who have identified information-theoretic measures for quantifying the intrinsic \textit{semantic} information of an agent/environment coupling, which is information that is salient to a system's persistence.

\textbf{As our title suggests, our claim is that reward is not \textit{necessary} in the definition and function of a basic self-preserving agent that can perform complex tasks and create abstract hierarchical actions. We do not \textit{prove} that a self-preserving agent cannot be constructed on a foundation of reward-maximization, but we do cast doubt that reward-maximization is appropriate for open-ended agency due to the problem of defining rewards on combinatorially large spaces of variables.} We also do not dispute the usefulness reward-maximization as a general method for a variety of model-based or model-free AI algorithms that have made tremendous progress in a variety of areas, from matrix multiplication algorithm discovery to plasma control \cite{fawzi2022discovering,degrave2022magnetic}.  


Lastly, in addition to contesting parts of RIEH, we also challenge the position of Vamplew et al. \citeyear{vamplew2021scalar} that many problems should be framed as multi-objective in the first place. MORL requires us to pick a policy from a space of multi-dimensional value trade-offs called a Pareto Frontier, and the selection requires a utility function that takes weight parameters which prioritize some objectives over others. The introduction of weight parameters introduces the new problem of needing to explain the weights. However, as we show, problems which \textit{appear} to require a multi-objective treatment, such as multi-dimensional physiological regulation, need not be formalized this way; rather, we argue in chapter \ref{sec:MORL}, that these problems can be formalized as single-objective multi-goal problems defined on multi-dimensional state-spaces, thereby avoiding Pareto Frontier weighting altogether.

\subsection{The Problem with Reward in Cartesian Product Spaces}
Goal-directed behavior has typically been framed by RL in terms of the maximization of accumulated rewards and minimization of punishments over time. However, in a life-long learning setting an agent will not have access to knowledge about reward functions as mathematical objects, only reward signals, and it is unclear how reward signals could be conditioned on multiple variables in a high-dimensional spaces---where does the information about state-combinations come from? Therefore, for clarity, it is worth approaching the problem from a full model-based perspective, where we consider what reward-maximization problems look like when the state-space and transition operator dynamics of an agent expands over time.

\begin{figure}[h!]
    \centering
    \includegraphics[width=\columnwidth]{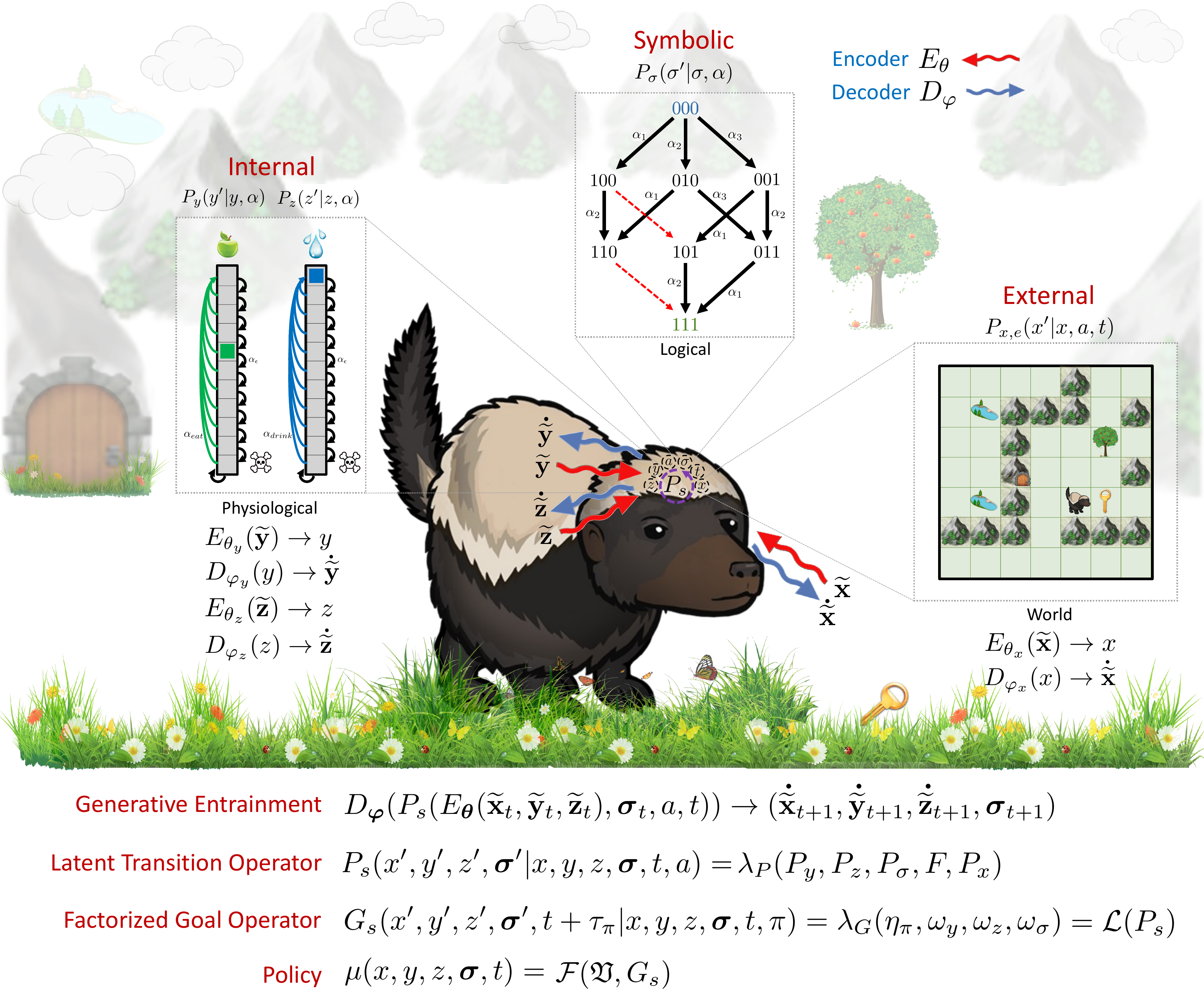}
    \caption{\small Stoffel, a generatively entertained self-preserving agent, lives in a world of high-dimensional signals that are made comprehensible and amenable to computation by compressing them down to discrete states through an encoder $E_{\theta}$. A generative model $D_{\varphi}$ (decoder) can anticipate the high-dimensional signals influenced by other sensory modalities, symbolic variables $\boldsymbol{\sigma}$, and action through a latent transition operator $P_s$. Thus, the agent is \textit{entrained} to the umwelt of coupled internal and external domains over time. This paper does not address the optimization of $D_{\theta}$ and $E_{\varphi}$, but rather asks two central questions: 1) If a factorization were possible for $P_s$ through a composition operator $\lambda_P$, what objective function $\mc L$ can optimize a planning factorization $(\eta_{\pi},\omega_y,\omega_z,\omega_{\sigma})$ so the agent can efficiently plan in a large non-stationary Cartesian product-space? And, 2) how can we optimize a policy $\mu$ with an objective function $\mc F$ and a reward-free intrinsic motivation metric $\mathfrak{V}$ to create a self-preserving agent that grows or maintains the integrity of its own structure? We argue that $\mc L$ can be the Operator Bellman Equations, $\mathfrak{V}$ can be a valence function (empowerment gain), and $\mathcal{F}$ can be the Valence Bellman Equation.}\label{fig:dynamic_entrainment}
\end{figure}

Transition operators are objects that describe the dynamics of states of a system under action. For example, a transition operator $P_x(x'|x,a)$ maps a state $x$ to the next-state $x'$ under an action $a$ with a given probability. If we consider transition operators as a central object of learning, then over a life-time the accumulated knowledge of a large set of transition operators and their couplings necessarily induces a combinatorially large space of state-combinations. The space of state-combinations is called a \textit{Cartesian-product space}. For two state-spaces $\mc X = \{x_1,...,x_n\}$ and $\mc Y= \{y_1,...,y_m\}$, the Cartesian product space is written as,
\begin{align*}
    \mc X \times \mc Y := \{(x_1,y_1),(x_1,y_2),...,(x_n,y_{m-1}),(x_n,y_m)\},
\end{align*}
which is of size $nm$. Product-spaces are combinatorially large in the number of sets used. Consider the scenario depicted in figure \ref{fig:dynamic_entrainment}, where Stoffel has $4$ sets of states representing spatial location ($\mc X$), physiological states $(\mc Y, \mc Z)$ and logical states ($\Sigma$). Each of these state-spaces can have its own dynamics; for example, $P_x$ mentioned earlier, is an operator local to a set of states $\mc X$, along with $P_y$ $P_z$, and $P_\sigma$ for the other spaces. However, a transition operator $P_s$ that probabilistically maps state-vectors $(x,y,z,\boldsymbol{\sigma})$ to next state vectors $(x',y',z',\boldsymbol{\sigma}')$, will be enormous and impractical to \textit{explicitly} represent even for (non-trivially small) grid-world environments, as it is defined on the product-space $\mc X \times \mc Y \times \mc Z \times \Sigma$. Furthermore, there can be conditional interactions between the dynamics of state-variables, for example, an apple can induce conditional dynamics on an object-possession state $\boldsymbol{\sigma}$, which will depend on its location $x$ and action $a$ for picking it up. Thus, a product-space operator would require some kind of factorization using individual components bound together by a function $\lambda_P$ (defined later in equation \ref{eq:hierarchical_trans_op}), along with a function $F$ which couples the individual component transition operators together, 
\begin{align*}
P_s(x',y',z',\boldsymbol{\sigma}'|x,y,z,\boldsymbol{\sigma},a,t)=\lambda_P(P_y,P_z,P_\sigma,F,P_x).
\end{align*}
In general, the product space of a life-long agent will be enormous considering the exponential increase in possible state-vectors when incorporating additional knowledge of state-spaces over a life-time. This complexity---what Richard Bellman calls the curse of dimensionality \cite{bellman1966dynamic}---poses significant challenges for intelligent systems, which need to use some kind of intellectual phenotype or task-decomposition to reason about actions and their consequences in real time without extensive trial-and-error feedback, often for the purpose of survival.

From the RL perspective, a physiological agent like Stoffel that acts in hierarchical Cartesian product space would also require some kind of reward-signal to incentivize the completion of a task and to avoid death. One might intuit that fully satiated states are maximally rewarding, or perhaps these  states produce no reward, but the bad ``death" states (starvation, dehydration) incur arbitrarily high costs. \textbf{While these are reasonable assumptions for physiological states, it is not obvious how to define a reward-function on \textit{any} given Cartesian product-space that encodes context-sensitive goals that are relevant at a given moment}. Should the product-space reward function $\mc R$ be a linear or non-linear function, $h$, of individual reward functions $r$ on each space,
\begin{align*}
    \mc R(x,w,y,z,...) = h(r_x(x),r_w(w),r_y(y),r_z(z),...)~~?
\end{align*}
Where would $h$ come from? Would nature furnish the product-space with reward? What information about the agent or the world would this composite reward function communicate? And, how would an agent decide that this reward signal is \textit{salient} to it? Some authors have suggested that reward functions could have an evolutionary explanation, showing that they can be \textit{evolved} on a given space to keep an agent alive \cite{singh2009rewards}. However, there are countless possible product spaces that an agent could experience in a lifespan, and it is unclear how an evolution-derived reward function should generalize to any new product space on the fast time-scale of an agent's life. If a Cartesian product-space operator $P_s$ defined on the product-spaces $\mc X \times \mc W \times \mc Y \times \mc Z$ were to expand to accommodate new states and transition structure ($P_s'$ incorporating the state $\boldsymbol{\sigma}$ in $\Sigma$), there is no naturalistic process that informs the agent how the reward function, $\mc R$ (as well as $h$), of an expanded state-vector should be expanded to a new function $\mc R'$:
\begin{align*}
    P_s(x',w',y',z'|x,w,y,z,t,a) &\rightarrow P_s'(x',w',y',z',\boldsymbol{\sigma}'|x,w,y,z,\boldsymbol{\sigma},t,a)\\
    \implies \mc R(x,w,y,z) &\xrightarrow{?} \mc R'(x,w,y,z,\boldsymbol{\sigma}), 
\end{align*}
To be an explanation, a mechanism for how to expand $\mc R$ must have an explainable origin---what information about \textit{normativity} is the reward \textit{communicating}? This is a critical question that theorists of open-ended reinforcement learning would need to answer.

One might suggest that agents will not have reward functions defined on specific state \textit{combinations}. Rather, when an agent encounters a new state-space $\Sigma$ there is simply a function $r_{\boldsymbol{\sigma}}$ not conditional on $(x,w,y,z)$, which is added to the rewards of other states. Another possibility is that new abstract states do not actually have rewards or costs, but they contribute indirectly to the total reward accumulated (i.e. in the value function) from a more privileged set of state-spaces for which nature \textit{has} provided reward functions, such as hedonic physiological states. This suggestion is extreme because it implies that new forms of value for abstractions are always instrumental to accumulating some fixed core reward function; it is exemplified by a mathematician who discovers that the value of learning a new theorem or analytic technique is indirectly due to the ability to satisfy rewarded physiological variables like hunger because the mathematician will remain employable and materially compensated to buy food. Both of these suggestions imply that the contextual information about state-combinations arises implicitly when computing the optimal value function of the long run reward accumulation. The first suggestion doesn't explain where the new reward function comes from, and worse, the second suggestion assumes that reward functions are static and unchanging and that new forms of normative signals never appear in the world. Neither of these possibilities are satisfactory answers that would allow agents to flexibly \textit{reason} about new forms of value. Furthermore, while we are focusing our critique on reinforcement learning, it should be noted that other paradigms, such as Active Inference (ActI) \cite{friston2009reinforcement}, are burdened by the same problem: ActI does not account for how the generative observation model in the free-energy functional should be re-defined to accommodate new knowledge about state-spaces and transition dynamics. We contend that the problem of product-space rewards or generative models for ActI, in fact, does not need to be solved. Instead, we will argue that we can compute value from intrinsic controllability measures on the agent's entire hierarchical structure.

To address this question in a different way, we must consider the possible ontology of life-long agents along with the computational properties and processes they might employ to tackle the problem of product-space planning and valuation.

\subsection{Hypothesis: The Ontology of a Self-Preserving Agent}

Inspired by Stoffel, we provide a hypothetical structure of an embodied agent, illustrated in Figure \ref{fig:dynamic_entrainment}, to frame our discussion of the representations and mechanisms for defining fully autonomous systems. Consider the possibility that world presents many high-dimensional internal and external signals to Stoffel's sensory systems, where the signals are mapped down to a discrete state-space through an encoder, $E_{\theta}(\widetilde{\textbf{s}})\rightarrow s$, and generatively mapped back to the high-dimensional space through a decoder, $D_{\varphi}(s)\rightarrow \dot{\widetilde{\textbf{s}}}$, accurately reproducing the original signal. An agent that is constantly conditioned by these signals and engaged in their reproduction would be \textit{generatively entrained} to its umwelt through time. Let us limit the theory to $4$ spaces for now, while recognizing that their could be many more. For instance, Stoffel could be engaged in the continual generative production of the entirety of his interoceptive and proprioceptive domains through time, but we will simplify this intuition and consider only hydration and caloric physiological domains. The purpose of discussing this generative entrainment is simply to frame the problem of real-world agents: it is imperative that organisms which plan and reason across many different continuous domains will need computational techniques to mitigate the induced product-space of state-variables. The benefit of discrete latent states is that we can define discrete planning algorithms on them, and they can also evolve along with symbolic logic task states (such as a binary vector $\boldsymbol{\sigma}$, e.g. $\boldsymbol{\sigma} = [1,0,...,1]$) that mediate the coupling of internal and external spaces through cross-domain conditioning and action. An encoder and decoder could map all spaces to and from discrete and continuous domains (e.g. $E_{\boldsymbol{\theta}}(\widetilde{\mathbf{x}}_t,\widetilde{\mathbf{y}}_t,\widetilde{\mathbf{z}}_t)\rightarrow (x_t,y_t,z_t)$), and the latent coupling of dynamics could be described by an operator $P_s$ (defined in \eqref{eq:hierarchical_trans_op}), which dictates the evolution of the entrainment,
\begin{align*}
    &\text{Transition Operator:}&&P_s(x',y',z',\boldsymbol{\sigma}'|x,y,z,\boldsymbol{\sigma},a,t)=\lambda_P(P_y,P_z,P_\sigma,F,P_x),\\
    &\text{Gen. Entertainment:}&&D_{\boldsymbol{\varphi}}(P_s(E_{\boldsymbol{\theta}}(\widetilde{\mathbf{x}}_t,\widetilde{\mathbf{y}}_t,\widetilde{\mathbf{z}}_t),\boldsymbol{\sigma}_t,a_\pi,t)) \rightarrow (\dot{\widetilde{\mathbf{x}}}_{t+1},\dot{\widetilde{\mathbf{y}}}_{t+1},\dot{\widetilde{\mathbf{z}}}_{t+1},\boldsymbol{\sigma}_{t+1})\\
    &\text{Agent Time-series:}&&(\dot{\widetilde{\mathbf{x}}}_{t_0},\dot{\widetilde{\mathbf{y}}}_{t_0},\dot{\widetilde{\mathbf{z}}}_{t_0},\boldsymbol{\sigma}_{t_0})\xrightarrow[\pi]{DP_sE}(\dot{\widetilde{\mathbf{x}}}_{t_1},\dot{\widetilde{\mathbf{y}}}_{t_1},\dot{\widetilde{\mathbf{z}}}_{t_1},\boldsymbol{\sigma}_{t_1}) \xrightarrow[\pi]{DP_sE}(\dot{\widetilde{\mathbf{x}}}_{t_2},\dot{\widetilde{\mathbf{y}}}_{t_2},\dot{\widetilde{\mathbf{z}}}_{t_2},\boldsymbol{\sigma}_{t_2})...
\end{align*}

The role of a binary vector $\boldsymbol{\sigma}$ is to encode complex non-Markovian logical conditions of behavior on the external state-space $\mc X$ necessary to induce transformations on the internal spaces $\mc Y$ and $\mc Z$, such as the requirement to obtain multiple items in order to cook and eat food. While $\boldsymbol{\sigma}$ is a binary vector space in our example, it could also represent the state of any automata-like state-machine. By non-Markovian, we simply mean that an agent's policy $\pi$ (which outputs actions to control on space $\mc X$) is conditioned by its history on $\mc X$ recorded as a state in another space.  For example, obtaining three necessary items for a single task requires history dependence on $\mc X$ which is registered in a bit-vector space $\Sigma$ so that the problem is Markovian on the full product-space.


\subsection{Physiological Regulation}
The constraint of physiological regulation---the need for agents to regulate critical internal states corresponding to hunger, hydration, and temperature which underpin their normal functioning---is an important but underdeveloped test-case for assessing the power of planning and intrinsic motivation algorithms. These constraints can shed light on the qualitative features and computational challenges facing biological organisms and artificial agents, namely, the problem of planning in non-stationary Cartesian product spaces of variables. There is intrinsic structure to physiological problems that is not exploited by standard model-based Bellman formalisms, which points towards ideal representations that \textit{should} be computed in order to deal with the real complexity of the world. Physiological regulation is also a good test-case because there are well-defined conditions in which the agent is \textit{alive} and has the capacity to influence the world, and when it is \textit{dead}, and does not. A number of other researchers have identified physiological regulation and self-preservation as key constraints to address \cite{roli2022organisms}. Sennesh et al. \citeyear{sennesh2022interoception} have developed control models for allostasis, Kiverstein et al. \citeyear{kiverstein2022problem} have suggested that the free-energy principle could address physiological self-preservation \cite{friston2009reinforcement,friston2010free}, and Man and Damasio \citeyear{man2019homeostasis} argue that today's current robots ``lack self-hood and `aboutness,'" and that meaning may emerge when homeostatic consequences impinge upon an agent's capacity for information processing. Additionally, a pioneering avenue of research called Homeostatically Regulated Reinforcement Learning (HRRL) was put forward by Keramati and Gutkin \citeyear{keramati2011reinforcement} who formalized a theory of internal drive reduction (inspired by the early work of Hull \citeyear{hull1943principles}) and showed that closing the gap between a current state and a physiological set-point could be equivalently cast as a reward maximization problem by formalizing the reward as being proportional to the distance reduced to the set-point \cite{keramati2014homeostatic, laurenccon2021continuous}. 

From the behavioral- and neuroscience perspective, Juechems and Summerfield \citeyear{juechems2019does} provide a valuable critique of RL. They argue that there is no known external entity that can determine the reward of an action, nor is there a dedicated channel distinct from the classical senses that registers a receipt of rewards. Rewards are just scalars, so how is an agent supposed to distinguish a reward from any other kind of (non-rewarding) sensory data? Similar to Srivastava and Schrater, they conclude: ``rewards and punishments ‘are’ sensory observations [...] and so stimulus value must be inferred by the agent, not conferred by the world." 
Drawing on the strengths of HRRL, they conclude
that organisms may set their own goals which act as set-points within the organism. This idea is similar to the previously mentioned concept of autotelic agency, which is built on the foundation of goal-conditioned RL.
Juechems and Summerfield conclude by asking the question about the intellectual phenotype: what mechanisms and representations would make it possible?  We seek to address this question, in addition to a couple other shortcomings of HRRL; namely, 1) HRRL does not explain why an undesirable state, such as starvation, dehydration, or thermal disregulation is bad in terms of a \textit{functional consequence} to the agent, it assumes undesirable states are bad simply by being far from a set-point, and 2) it does not explain how agents could reason in a multi-dimensional physiological space in real-time. We will show how these challenges can be met with Bellman equations that produce reusable and remappable compositional operators rather than value functions, and which decompose over a hierarchical space by exploiting implicit structure of the hierarchy.

\subsection{The Value of Valence}
Given the difficulty of defining product-space rewards for large hierarchical spaces, is there a different approach we could take to develop agents which reason about goals and motivation in novel circumstances in real time? Important insights can be found in a recent paper by Roli et al. \citeyear{roli2022organisms}, where they argue that an artificial agent with true agency must \textit{want} something against the background of limitless possibilities, and moreover, they need to be able to leverage new affordances in the service of their desires. They argue that current AIs are framed as input-output machines, or, \textit{algorithms} in the classic sense, but that this is a limited perspective because one cannot enumerate all of the possible goals, actions, and affordances prior to the computation of a solution. They proceed to argue that true autonomous agents capable of wanting must be ``Kantian wholes," meaning, agents that are organized such that their parts ``exist for and by means of the whole," as originally articulated by the philosopher Immanuel Kant. A Kantian agent, Roli et al. suggest, could exhibit an open-ended form of \textit{organismal agency}: an agent that is ``able to perceive its environment and to select from a repertoire of alternative actions when responding to environmental circumstances based on its \textit{internal organization} [emphasis added]." Our work is in agreement with this perspective on agency. We believe that instead of using rewards, advanced Kantian agents should ideally be able to reason about the vast space of possible state-vectors that have not necessarily been experienced in the past, and in a naturalistic and model-based teleological fashion, organize their behavior around the realization of a spatiotemporally distant state-vector that has been \textit{justified} by its impact on an agent's future internal organization. 

We argue that the problem of \textit{machine wanting}, and the process goal-justification, can be addressed by \textit{empowerment gain} maximization in the Cartesian product space of SPA's coupled internal and external transition operators (which we call product-space \textit{valence}), where the controllability of the product space must be maintained or expanded. Formally, empowerment (reviewed in section \ref{sec:Empowerment}) is defined as,
$$\mathfrak{E}_n(P|x)=\max_{p(\mathbf{a}|x)}I(A_n;X),$$
which is the $n$-step channel capacity of a transition operator $P$, and quantifies the maximum mutual information $I$ between sequences of $n$ actions (random variable $A_n$) and resulting states (random variable $X$) under $P$ starting from state $x$---this can be thought of as controllable optionality \cite{salge2014empowerment}.  Higher empowerment means an agent has a greater predictive capacity to realize a wide range of possible futures.

\begin{figure}
    \centering
    \includegraphics[width=\linewidth]{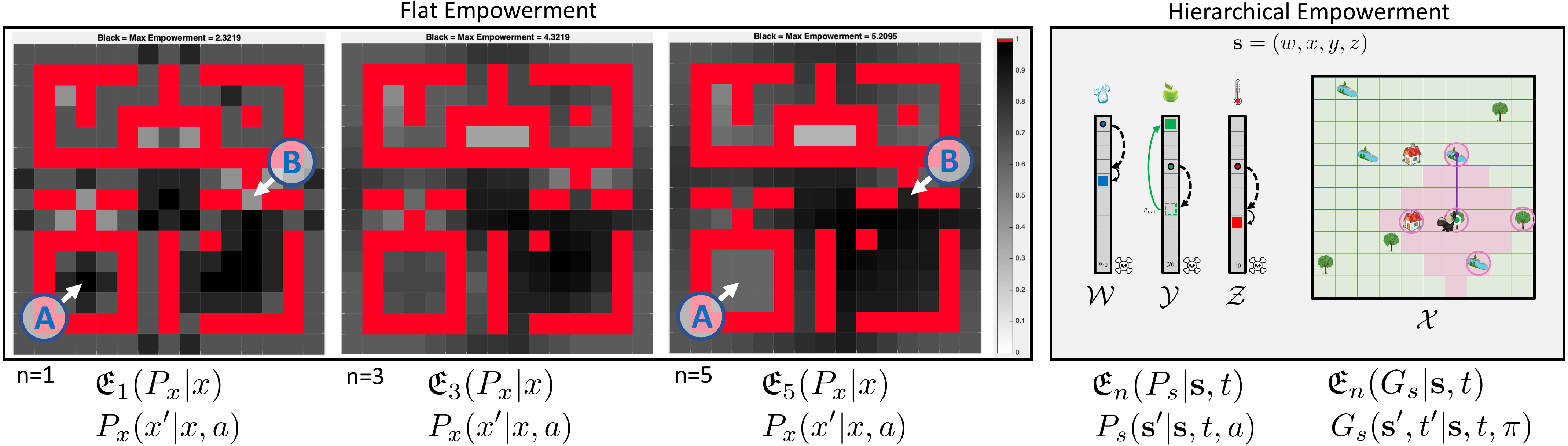}
    \caption{The first three maps show empowerment computed on a low-level state-space under an operator $P_x$ at different horizons $n$, where the walls are red and the darker states indicate higher empowerment. High-empowerment agents are at states where they have high optionality, but here the agent does not perform complex tasks. The right panel shows the focus of this paper: how to compute empowerment in the product-space of Stoffel's \textit{ontology}, which could collapse over time. While computing empowerment on $P_s$ is possible (pink squares indicate Stoffel's optionality), it operates over a limited time-span, so we also compute empowerment where actions are goal-conditioned \textit{policies} for an abstract spatiotemporal goal-operator $G_s$ parameterized by feasibility functions. Stoffel performs \textit{tasks} (pink circles) to regulate his internal and external states to increase task-empowerment.}
    \label{fig:empower}
\end{figure}

To illustrate empowerment, figure \ref{fig:empower} shows normalized empowerment plotted in a gridworld at different horizons. Notice that at a short action-horizon $n=1$, the square $A$ is high empowerment (black), whereas $B$ has low empowerment (light grey) because it is next to walls which limit its possible one-action futures. However, at a longer horizon $n=5$ the situation flips: $A$ has low empowerment because the it is confined to the room, but $B$ has high-empowerment because it is near a wide-open space. Empowerment can also be used as an objective to balance a cart-pole or invert a double pendulum, because those are states from which the largest range of possible future states can be predictably realized \cite{salge2017empowerment}. 

A potential criticism of empowerment is that it only measures optionality, and it does not dictate what tasks to work on. This might be true if empowerment is computed on a single flat state-space when reward functions are considered to constitute a \textit{task} (a semantic convention which we reject). \textit{Flat empowerment} in the gridworld is computed only on the low-level actions of a transition operator $P(x'|x,a)$. Flat empowerment is also only confined to a short spatiotemporal range, because increasing horizon parameter $n$ increases the computation-time exponentially due to the combinatorial increase in possible action sequences.  A key innovation of this paper is to scale empowerment to all state-spaces that an agent represents and cares about. We show how to compute empowerment on abstract product-space Goal Operators $G_s$ acting on state-vectors $\mathbf{s}=(x,w,y,z,...,\boldsymbol{\sigma})$,
\begin{align*}
    &G_s(\textbf{s}',t+\tau_{\pi}|\textbf{s},t,\pi),\\
    &\mathfrak{E}_n(G_s|\mathbf{s},t),
\end{align*}
and have \textit{policies} $\pi$ as actions which map the agent over long spatial and temporal ($\tau_{\pi}$) distances in an agent's Cartesian product-space. In this paper, the \textit{difference} in empowerment over a long course of action $\rho$ will be quantified by a \textit{valence function}, 
$\mathfrak{V}$:
$$\mathfrak{V}(G_s',G_s,Q,\mathbf{s}_0,t_0, \rho) = \expec_{\rho\sim Q}\left[\mathfrak{E}_n(G_s'|\mathbf{s}_\rho,t_\rho)-\mathfrak{E}_n(G_s|\mathbf{s}_0,t_0)\right],$$ 
which is a function of the change in an agent's state-vector $\mathbf{s}$ and hierarchical transition operator $G_s$ (the details of this function can be found in section \ref{sec:self-preserving}). We will show that unlike flat empowerment, in hierarchical state-spaces reward-free goal-directed motivation is entailed by optimizing empowerment-gain especially when the agent's empowerment in the hierarchical state-space can collapse over time in the absence of planning. For instance, if an agent gets hungrier over time, eventually there will come a point in which the agent dies and cannot move around to perform other tasks, effectively arresting its capacity to control all efferent state-spaces which could resuscitate it. Putting the computational work into achieving goals that transform some other (physiological) state-space then becomes an imperative. Agents which maximize valence will seek out state-vectors, as goals, which facilitate a favorable coupling between its internal states, skills, and the external states of the environment, and therefore these agents exhibit self-concern for the maintenance and growth of their own structural integrity\footnote{To anticipate a potential misconception, valence maximization does not directly optimize the quantity of time-from-death. If death is inevitable, one could in principle have low-empowerment agents with long life-spans and high-empowerment agents that have shorter life-spans. And, one could have agents which are equally ``close-to-death" in the absence of control, but attain different levels of empowerment (e.g. compare a prisoner to a billionaire CEO, both of which are close to death in the absence of control but with different levels of empowerment). Death state-vectors (and state-vectors that invariably lead to them) are simply avoided if better options are available and the choices between empowerment and life-span are contingent on look-ahead hyper-parameters, which is an interesting topic worthy of further discussion that we will not address in this paper.}.

It is important to keep in mind that many state-elements of a large state-vector can be quasi--death-states (or simply ``bad-states") within a sub-space of a Cartesian product space. Consider an agent that enters the state of being fired from a job. In such a case, the agent would still be alive, but restricted in how it can use its skills to acquire money to buy food and shelter. The important capability of empowerment is that it can discover, without supervision, the goodness or badness of states as intrinsic properties of the product-space transition operator. Valence in a hierarchical space can be thought of the contraction or expansion of an agent's capacity to control other internal and external state spaces, in the same way an army lieutenant might sense the internal contraction of his or her capacity to perform learned skills and tasks in the world after a demotion, or how Stoffel might sense the expansion of his capacity to access parts of the world and procure objects, food, and mating opportunities external to Badger Alcatraz if he were to escape---these are computations that propagate information \textit{across} a hierarchy of state-spaces and transition operators. 

Having explained hierarchical empowerment and its role in normative valuation, we can now define a form of agency we will refer to as \textit{teleological agency}: 
\begin{quote}
    An organism possesses teleological agency if it is capable of generating a spatially and temporally distant goal-state, quantifying its normative value, and can attempt to realize it, wherein the normative valuation is derived (via computation) from the impact of the goal on the capacity of the agent's entire control architecture to affect the internal and external world under its domain.
\end{quote}
The notion of \textit{purposefulness} associated with the word teleology is distinctly present in this definition, as goal-states have the potential to causally bring about or preserve environmental affordances associated with the capabilities of the control architecture---purpose, here, comes from the functional significance of the components of the control architecture in relation to the goal state. This is contrast from a weaker form of agency in which an agent, such as a squirrel, takes actions to navigate to a spatial goal-state to obtain and eat an acorn without such deliberative considerations. The acorn could still be significant to sustaining the functioning of a squirrel's control architecture, and navigational computations may still be required, but simpler mechanisms which map a stimulus to a goal-state could suffice in lieu of the deliberative processes underlying teleological agency. What teleological agency provides is a means to evaluate and organize behavior for realizing complex states of the world that have yet to be experienced, but have a purposeful, functional significance to the agent's structure. The theory we develop in this paper is intended to advance our understanding of this form of agency.


    

\subsection{The Value of Modular, Compositional, and Factorized Representations}

\begin{figure*}[h]
    \centering\includegraphics[width=\linewidth]{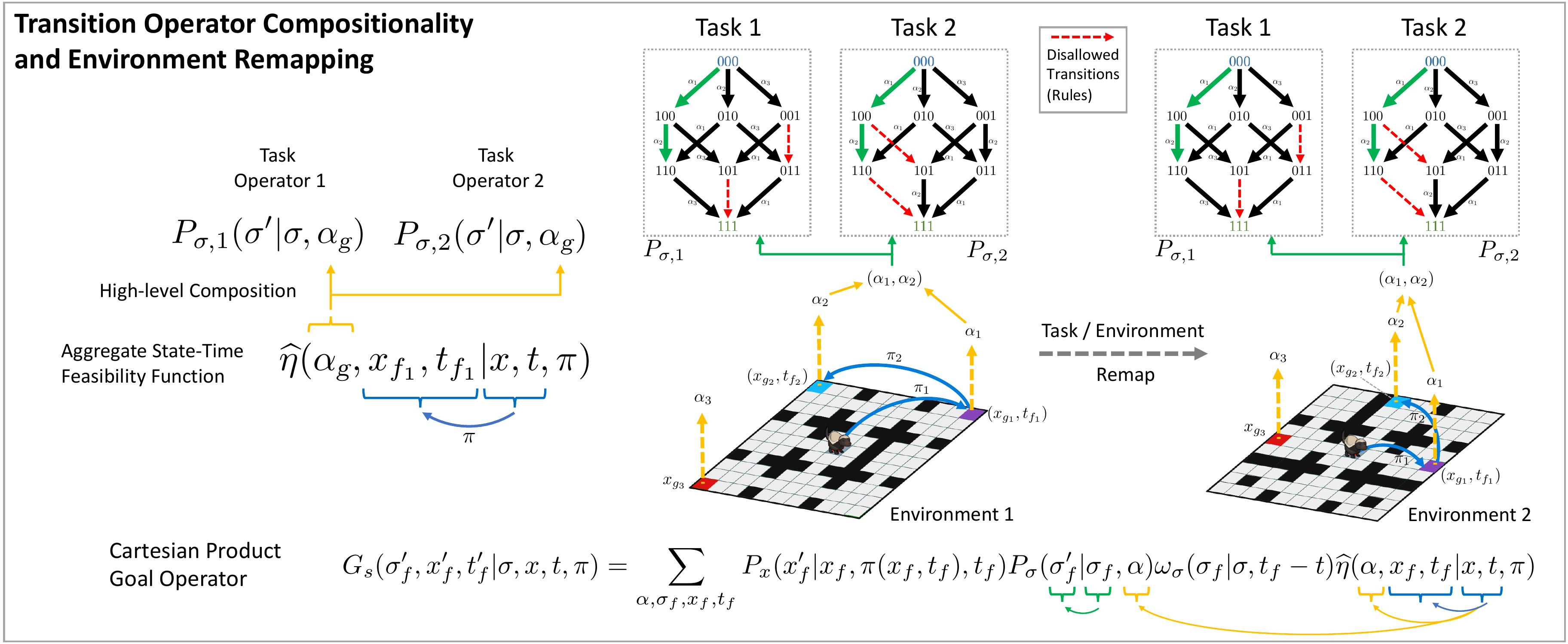}
    \caption{The Hierarchical TG-CMDP presented in section \ref{sec:Hierarchical OBEs} facilitates key features of flexible intelligent, such as the ability to compose and remap representations. STFFs can be sequentially composed to update the final state-times of following a sequence of policies (e.g. ($\pi_1,\pi_2$)) where $\hat{\eta}$ is an \textit{aggregate} function of multiple individual problem solutions (see section \ref{sec:forward}), and the goal variables $\alpha_\dg$ permits composition with higher-order state-spaces such as the binary vector task modules. Importantly, multiple modules with different structure can be can be mapped to the same goal variable within an environment, or modules can be \textit{remapped} to new low-level TG-MDP solutions in a new environment, where state features such as color correspond to different high-level actions $\alpha$.}
    \label{fig:composition}
\end{figure*}

We have introduced the idea that we can construct factorized abstract operators $G_s$ for planning and empowerment computations within a product space, but have not yet described how these operators are actually constructed. The answer is by the computation and aggregation of compositional factorized representations called State-Time Feasibility Functions (STFF), which we will describe shortly---this factorization can be seen in Figure \ref{fig:composition} and defined in equation \eqref{eq:jump-op} in the paper. Without factorized representations a hierarchical state-space is difficult to \textit{plan} within due to the size of the space. Planning in non-stationary product-spaces is a primary challenge to both natural and artificial agents alike, and agents that plan in a product space will find two extremes untenable: dynamic programming \cite{bertsekas2012dynamic}, a recursive method for computing the exact solution, is impractical due to the problem of representing the solution, and forward sampling sequences of low-level actions from a starting state is also impractical due to the size of the action tree. 

In Deep RL, agents forward sample actions, but most of the true complexity and structure underlying a problem such as a video game is not explicitly represented by the agent, rather, the agent constructs inner representations from experienced rewards \cite{franccois2018introduction}. This often takes the form of a recurrent neural network, trained by rewards to contextualize a policy on agent history. In doing so, researchers are essentially baking in implicit information about the solutions to a non-stationary and non-Markovian task into the \textit{weights} of a network, and weights are static until experience is used to update them. The downside of this approach, arguably, is that it requires lots of trial-and-error feedback, which may contribute significantly to the high sample complexity of Deep RL and renders the agent incapable of \textit{conjecturing and reasoning} about task structure in real time.  We believe that the way to address the central problem of sample complexity is to develop and use model-based theory for hierarchical decomposition and abstraction that entails a reusable goal-conditioned planning factorization for structured and remappable non-stationary non-Markovian tasks (see figure \ref{fig:composition})---tasks which can be composed hierarchically and be mapped to new environments. We advocate for creating artificial agents which represent time explicitly in planning operators instead of implicitly in network weights; such an approach allows agents to flexibly forward sample at the level of goal-conditioned policies in a non-stationary product-space. 

\begin{figure}[h]
\centering
\includegraphics[width=\linewidth]{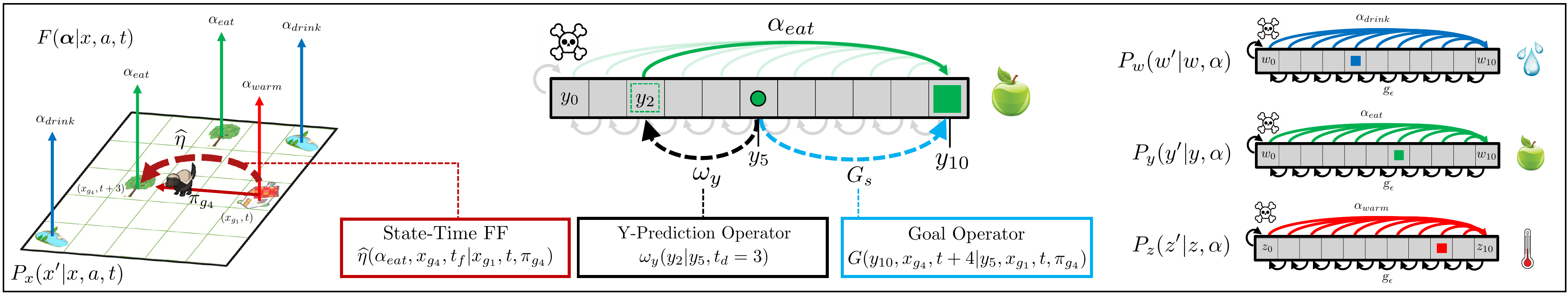}
\caption{State-time Feasibility Functions, Goal-Operators, and Internal State Transition Operators: A central theoretical innovation of this paper is the development of Bellman equations which produce abstract spatiotemporal transition operators $\eta$, which map initial state-times to final state-times under a policy. We can use the time difference $(t_d=t_f-t)$ encoded by $\eta$ under a give policy $\pi$ to update all other higher-order state-spaces with operator $\omega$, in parallel, that evolve for a given period of time under the policy, and update the state-spaces influenced by the goal by one step to form the goal-operator $G_s$ (blue dashed arc). This factorization allows us to reason with abstract jumps on the low-level state-space (red dashed arc), but also in (but not limited to) physiological state-spaces (black dashed arc) to forward-plan in an otherwise intractable Cartesian product space of variables.}
\label{fig:Homeostatic Control 1}
\end{figure}

To this end, in section \ref{sec:TG-MDP} we introduce a new class of decision process called the Temporal Goal Markov Decision Process (TG-MDP), and the hierarchical version, Temporal Goal Compositional MDP (TG-CMDP). These decision process have Bellman equations called Operator Bellman Equations (OBEs, see section \ref{sec:OBEs}), a reachability optimization that, rather than optimizing a value function, optimizes a feasibility function $\kappa_{\dg}(x,t)$, which is the cumulative probability of achieving a \textit{goal variable} $\alpha_\dg$ from state-time $(x,t)$ under policy $\pi_\dg$. The OBEs do not use a reward function, rather, they use an availability function ($f_{\dg}$ in figure \ref{fig:Homeostatic Control 1}) that specifies the probability that a goal is available, where this availability signal is \textit{intrinsic} to the connections between state-spaces in a hierarchical transition operator $P_s$. By using the availability function, the OBEs can also generate the aforementioned spatiotemporal transition operators called state-time feasibility functions, which probabilistically maps initial state-times to both the final state-time and goal completed under a policy $\pi$ (see Figure \ref{fig:Homeostatic Control 1}),
$$\eta_{\pi}(\alpha_\dg,x_f,t_f|x,t).$$
The goal variable here plays the functional role of an \textit{action} $\alpha$ on another transition operator for a different state-space. For example, in $P_\sigma(\boldsymbol{\sigma}'|\boldsymbol{\sigma},\alpha_\dg)$ a goal-action $\alpha_\dg$ flips the $\dg^{th}$ bit in a bit-vector $\boldsymbol{\sigma}$, or in $P_y(y'|y,\alpha_{eat})$ the goal-action $\alpha_{eat}$ transitions the agent to the top of the caloric energy state-space. Thus, these STFFs are inherently compositional, as they can sequentially compose with each other, and also vertically compose with higher-order transition operators by conditioning their action space. Doing so permits compositionality at different levels of representation, shown in figure \ref{fig:composition} (this was demonstrated by Ringstrom et al. for stationary state-to-state feasibility functions \cite{ringstrom2020jump}). In the hierarchical setting we can compute STFFs only on the base state-space $\mc X$ and combine them with prediction operators $\omega$ of the higher-order state-spaces to form factorized feasibility functions which update the full hierarchical state-vector after following a policy.  This factorization is equivalent to computing the intractable hierarchical STFF $\bar{\eta}_{\pi}$ on the full space:
\begin{align*}
    &\bar{\eta}_{\pi}(\alpha_\dg,\boldsymbol{\sigma}_f,y_f,z_f,x_f,t_f|\boldsymbol{\sigma},y,z,x,t)\\
&\quad\quad\quad\quad=\omega_{\sigma}(\boldsymbol{\sigma}_f|\boldsymbol{\sigma},t_f-t)\omega_y(y_f|y,t_f-t)\omega_z(z_f|z,t_f-t)\eta_{\pi}(\alpha_\dg,x_f,t_f|x,t).
\end{align*}
The previously mentioned Goal Operator $G_s$ which we will compute empowerment on (discussed in the last section) is the product-space feasibility function $\bar{\eta}_{\pi}$ evolved one extra step after achieving the goal to update the resulting state. This is illustrated in figure \ref{fig:Homeostatic Control 1} and formally defined by equation \eqref{eq:jump-op}.

Crucially, the state-time feasibility function factorization allows us to forward sample policies and predict the resulting state-vectors in a high-dimensional space in parallel without representing a full product-space transition operator, where we can advance each of the components of a state-vector by pushing them through the factorization when applying sequences of policies ($\pi^{k_1},\pi^{k_2},...,\pi^{k_f}$) (see 
figure \ref{fig:Valcence_comb} for illustration), 
$$(\boldsymbol{\sigma},y,z,x,t)_{k_0}\xrightarrow{\pi^{k_1}}(\boldsymbol{\sigma}',y',z',x',t')_{k_1}\xrightarrow{\pi^{k_2}}...\xrightarrow{\pi^{k_f}}(\boldsymbol{\sigma}_f,y_f,z_f,x_f,t_f)_{k_f}.$$

Compositionality, which has been argued to be a key ingredient of human intelligence \cite{lake2015human}, is a core attribute of our agent, where feasibility functions can compose across a transition operator hierarchy and be remapped within the same environment or to new environments (see figure \ref{fig:composition})---we hypothesize that agents which achieve human-level sample complexity will need to employ these principles. Ideally, advanced agents should be able to rapidly synthesize \textit{theories} of how the world works in the form of composed state-spaces and transition operators, similar to various other constructivist approaches that have been advocated for in cognitive science \cite{tenenbaum2011grow, ullman2012theory, tsividis2021human}. Though we will not be performing inference over theories in this paper, a formalism for transition operator composition, and theory to plan with them is provided. While transition operator composition is a quick and easy computation (provided that we represent it in a factorized form and do not construct the full explicit operator), it implies an exponentially large space of state-vectors one could occupy (in this paper we compose with $\lambda_P$ and $\lambda_G$ operators). This is the regime that both natural and artificial intelligent systems inevitably find themselves in, and it is characterized by \textit{the problem of the implicit product space}: an agent that incorporates new transition structure into its knowledge-base through implicit composition does not have a corresponding normative representation to optimize for the composition. That is, the agent does not immediately understand how to control, and the \textit{consequences} of controlling, through the new space with respect to all other state-spaces the agent cares about maintaining. 
The trade-offs, conflicts, and synergism between individual sate-spaces in an exponentially large product-space are all left to be discovered through evaluation, which requires what is known as \textit{the freedom from immediacy} \cite{gold2007neural,roli2022organisms}, the possibility of introspective rumination---theorizing, planning, and evaluating---without the expectation of immediate action. If an agent composes a transition model, the value of any given state is not obvious and so the space will need be evaluated with the appropriate representations and an intrinsic motivation metric $\mathfrak{V}$ (valence, empowerment-gain) to optimize a policy: 
\begin{align*}
    &\text{Transition Model Composition (Easy): } && \xoverbrace{P_s}^{\mathclap{\text{Explicit}}} = ~\xoverbrace{\lambda_P(P_y,P_z,...,P_\sigma,F,P_x)}^{\text{Implicit}}\\
    &\text{Policy Optimization (Hard): } &&~~\pi = \mc F(P_s,\mathfrak{V})
\end{align*}
When OBEs are defined on a hierarchical composition of transition operators, $P_s$, a derived STFF factorization will constitute a corresponding goal-conditioned policy and feasibility function \textit{decomposition} which can be used to break down a problem into pieces and evaluate the structure of a product space with semi-Markov planning and empowerment computations via forward sampling, making the optimization (and state justification) possible. 

State-time feasibility functions therefore play a dual role for planning and intrinsic motivation: 1) they allow us to create flexible planning factorizations to build up an abstract control architecture for planning with goal-conditioned transitions around a product-space, and 2) we can compute empowerment in a hierarchical product space with them because feasibility functions are transition operators and empowerment is a function of transition operators.  These two points leads us to a profound conclusion: \textbf{a life-long agent such as SPA, that optimizes valence within an expanding product space of transition knowledge, does not need to \textit{represent} any running total of accumulated reward as many RL agents might, rather, for SPA, empowerment gain is reflected the changes of the ontology and is thus \textit{implicitly} accumulated, registered in both the full hierarchical state-vector \textit{and} changes to the structure of its hierarchical planning operator. These changes to the state and ontology thereby serve as a baseline for future plans to optimize against as the agent learns more about the world's dynamics}. 
It is for this reason that we do not consider valence to be a ``reward signal," as it is not a \textit{received quantity} from a state to be stored, rather it is a quantity which is \textit{derived} and \textit{revealed} through computational work applied to the system as a whole. Put another way, valence is a derived quantity used to initially justify a plan, but it is not a quantity that is ever \textit{experienced} as a received signal at the final state or along the path, since it simply reflects changes to the state of the agent and its structure. And, because OBEs produce abstract spatiotemporal transition operators, we can assess the product space by evaluating both empowerment and valence \textit{at long time-scales} through forward sampling\footnote{As we will explain in the paper, the long time-spans are considered \textit{within} a single empowerment calculation and also the time-span that separate initial and final empowerment calculations (valence).}.

\textbf{In the context of hierarchical control, optimizations centered around \textit{reachability} are more general than reward-maximization because reachability optimizations can produce reusable and remappable factorizations.} If the model-based foundations of an optimization, such as infinite horizon discounted control, cannot solve challenging problems with dynamic programming when everything is known, then we can expect that there will be limitations when it comes to corresponding theories of learning, such as RL. Our framework puts forward a different model-based perspective: that forgoing reward in favor of goal-availability signals is precisely what facilitates the factorization of an agent's control architecture through the production of abstract feasibility function transition operators, which makes both forward-planning in non-stationary product spaces efficient, and the evaluation of empowerment at long time-scales possible. 
\subsection{Additional Contributions}

Throughout this paper, we will also demonstrate a few other key properties of SPA, TG-MDPs and OBEs. In section \ref{sec:afford} we discuss how feasibility functions naturally fit into the philosophy of \textit{Gibsonian affordances} \cite{gibson1977theory}, in which the agent actively plans to change its environmental affordances to solve problems. In section \ref{sec:VBE} we define and study the properties of the Valence Bellman Equation (VBE), which is a finite horizon Bellman equation that uses a valence function instead of a reward function. The VBE is the model-based foundation SPA is formalized on and can be parameterized by OBE solutions. In section \ref{sec:compositionality} we demonstrate how multiple levels of transition operators can be composed \textit{vertically} so that an agent can plan when non-stationary non-Markovian tasks induce internal (physiological) state-transformations (such as retrieving multiple items to prepare and eat food). Also, in section \ref{sec:sublimation} we prove the Sublimation Theorem for hierarchical OBEs which bounds the product space feasibility from above with a feasibility function computed exclusively on a higher space. This theorem allows for abstract \textit{sublimated} reasoning on other (often symbolic) state-spaces, which permits the agent to rule out low-level policy samples which cannot contribute to the successful achievement of a task. 
Lastly, in section \ref{sec:lifelong}, we demonstrate how a product-space operator can be built up from experience for life-long learning and transfer across environments, where new information about high-level transition structure can be anticipated with priors on the valence of associated state-features. By doing so, an agent can learn transferable representations which can generalize across environments by \textit{remapping} known structure to state-features of new environments \cite{ringstrom2020jump}. 


Finally, it is important to note that though we framed the self-preserving agent as being an entrainment of generative faculties in the form of encoders and decoders, these objects will play no formal normative role in the theory of this paper. We use the notion of generative entrainment as a proposal to frame the problem as a plausible biological system and to inspire a new perspective, but we will only develop algorithms applied to the latent transition operator $P_s$, and not train any networks. However, we invite the reader to consider the importance of a generative entrainment for our theory and its relationship to other theories. Paradigms such as Active Inference \cite{friston2009reinforcement,friston2017active} have taken the Free Energy Principal \cite{friston2010free} in the direction of control, where the objective is to optimize a policy that minimizes a Free Energy functional. The expected free-energy functional is the negative model evidence (surprise) of incoming sensory data, which can be decomposed into the agent's generative prior expectation of incoming sensory observations plus an information gain term. In the setting of control, the agent will choose actions to ensure, in part, that incoming sensory data accords with its expectations while balancing the epistemic incentives of the information gain term. Thus, in ActI the generative model is in the objective function and therefore takes on a \textit{normative} role, where it encodes assumptions about what the agent should want in a way similar to that of a utility function or reward function. As we have argued, this creates significant challenges in explaining normativity when an agent's product space needs to expand over a life-time, because the generative model has no concrete definition in such an event.
In contrast, in our proposal of a generatively entrained self-preserving agent, the generative model constitutes the \textit{medium} that SPA can be defined on, where our intrinsic motivation functions will measure the internal \textit{integrity} (i.e. controllability) of the entire latent hierarchical space described by $P_s$. Our theory is therefore complimentary with Sims' \citeyear{sims2022self} call to create artificial agents that exhibit \textit{self-concern}. Although one can imagine that the quality of an encoder and decoder could impact the algorithm, we will not address this in our theory at this time. Indeed, good generative models are important and there may be possible theoretical connections or potential hybrid theories between ActI and empowerment, especially considering the similarities and connections that have been made between them \cite{biehl2018expanding, hafner2020action}. The important aspect of a generative entrainment is that it clarifies the role of sensory prediction: encoding high-dimensional signals down to discrete states and predicting their evolution over time is important for computing empowerment-gain and thus sensory prediction would play an \textit{instrumental} role in the computation of value, not a primary role as it does in ActI. 
Our intention in introducing the frame of a generative entrainment is to inspire the idea that, in an open-ended life-long learning scenario, an agent can scaffold new knowledge off of a persistent core ontology and certify the value of its acquired knowledge in relation to its effect on the ontology's intrinsic integrity.

\newpage
\subsection{Summary of Contributions}
In this paper we make several contributions:
\begin{itemize}
    \item We define a Temporal Goal Markov Decision Process, a process for formalizing jointly non-stationary non-Markovian problems, and which parameterizes the functions of the Operator Bellman Equation. 
    \item We formalize Operator Bellman Equations which produce abstract spatiotemporal transition operators for hierarchical planning called state-time feasibility functions. We make the case that representations of initial-to-final state or state-time transition operators is the appropriate kind of abstractions for flexible hierarchical planning and goal justification.
    \item We define \textit{Temporally Extended Semi-Markov Empowerment}, which is empowerment defined only on the final state-times of a policy. We show how state-time feasibility functions are good for computing empowerment at long time-scales in a Cartesian product space.
    \item We formalize a Valence Bellman Equation for maximizing total empowerment gain (valence) and show how it can be parameterized by feasibility functions.
    \item We demonstrate \textit{sublimated reasoning}, a property of hierarchical OBEs that allows agents to reason at a higher level of abstraction to rule out lower-level policies during forward sampling.
    \item  We discuss connections between our theory and the concept of \textit{Gibsonian affordances}
    \item We contrast our theory in relation to multi-objective RL.
    \item We argue SPA is well-suited for life-long learning of hierarchical transition operators, where learned task modules and operators can be remapped to new environments.
\end{itemize}
\newpage
\section{Terminology}

    Goals and tasks have a similar meaning in this paper. In general, \textit{goals} will be indicated by $\dg$, which are indices that can be attached to any state-time $(x,t)_{\dg}$, or action variable $a_{\dg}$ or $\alpha_{\dg}$. A task is simply all of the acceptable state-times for satisfying a goal (i.e. the set of acceptable goals).

\begin{mydef}{Goal and Task Definitions}{}
\begin{itemize}
    \item Goal: an index $\dg$, which indicates an acceptable state-time-action of a task.  Often it will be appended to a goal-action $\alpha_{\dg}$ which is an action on a higher-order state-space which can be induced from another lower-order state-space.
    \item Task: The state-time conditions $\mathscr{T}$ (goal-states and availability times) that a task-goal variable will be induced from a state-space. 
    \item Goal-variable: a variable, (e.g. $x_\dg$, $a_\dg$, $t_\dg$, $\alpha_\dg$), indicating that the variable is one of many variables that can satisfy a task.
    \item Null-variable: an action-variable which is not encoded by a task, but which conditions the "default" dynamics of another state-space as the agent acts to complete a task. This will often be referred to as a \textit{null goal-action} $\alpha_{\epsilon}$.
    \item Higher-order goal variable: A goal variable, e.g. $\bar{\alpha}_{\dg}$ (denoted with a bar), on a higher-order state-space, which is induced from an intermediate state-space, and not induced by the base state-space $\mc X$.
    \item Subgoal: Any lower-order goal which can be induced to satisfy a task encoded for an higher-order task-goal $\bar{\alpha}_{\dg}$.
\end{itemize}

\end{mydef}




\newpage
\section{Review: Empowerment}\label{sec:Empowerment}
 Empowerment \cite{salge2014changing, salge2017empowerment}, $\mathfrak{E}_n(P|x_t)$, is formally defined as the channel capacity $C$ of a channel derived from all open-loop $n$-step action sequences, $\mathbf{a} = (a^{(1)},...,a^{(n)})$, applied to the transition operator $P$, where an action sequence R.V. $A^{\tau-1}_t$ (with probability distribution $p(\mathbf{a}_t^{\tau-1})$ on action sequences) is the input from time $t$ to $\tau-1$ starting from $x_t$, $X_\tau$ is the final state R.V. with distribution
$p(x_\tau|x_{t},\mathbf{a}_t^{\tau-1})$, 
and $n=\tau-t$ is the horizon:
\begin{align*}
    \mathfrak{E}_n(P|x_t) &=C(A^{\tau-1}_t;X_\tau|x_t)\\
    &=\max_{p(\mathbf{a}_t^{\tau-1}|x_t)}I(A^{\tau-1}_t;X_\tau|x_t)=\max_{p(\mathbf{a}_t^{\tau-1}|x_t)}\left[H(X_\tau|x_t)- H(X_\tau|A^{\tau-1}_t,x_t)\right].
\end{align*}
The definition of channel capacity is the maximum mutual information across the channel over all possible input distributions. Here, $I$ is the mutual information between two random variables, and $H$ is the entropy of a random variable. 

Empowerment is a measure of an agent's capacity to predictably realize a variety of future state outcomes from a given starting state $x_t$, a kind of optionality. 
There are two extreme cases of empowerment. If $P(x'|x,a)$ has the same next-state distribution for each action, e.g. a uniform distribution, then $H(X_\tau|A^{\tau-1}_t,x_t)=H(X_\tau|,x_t)$ and so empowerment must be zero. Alternatively, if $P(x'|x,a)$ is deterministic, then the conditional entropy $H(X_\tau|A^{\tau-1}_t,x_t)$ is zero. This means that empowerment is the maximum possible final state entropy, $\max_{p(\mathbf{a}^{\tau-1}_t|x_t)}H(X_\tau|x_t)=\log_2(|\mc X_{\mathbf{a}_n}|)$ \cite{salge2014empowerment}, which is the log of the number of possible reachable states under open-loop plans $\mathbf{a}_n$ of length $n$. When these two conditions do not hold, empowerment is  computationally expensive for large $n$ because the row-space of the channel grows exponentially in $n$. So, while channel capacity is usually computed with the Blahut–Arimoto algorithm \cite{blahut1972computation,arimoto1972algorithm}, the advantage of determinism is that we only need to count the number of reachable states, which is the number of states with non-zero probability after a (tractable) $n$-step forward diffusion under a uniform action distribution (See A.\ref{appx:det_emp}). We will assume determinism for the examples of this paper to simplify the concepts without loss of generality. 

An agent with high empowerment in a single low-level state-space is in locations with many possible futures to exploit, and so maximizing empowerment can appear to be at odds with goal-directed behavior. However, as depicted in figure \ref{fig:empower}, empowerment can be a measure of not just a low-level stationary operator, $P_x(x'|x,a)$, but any operator, for example a state-time \textit{goal operator},
\begin{align}
    G_s(\mathbf{r}_f',x_f',t_f'|\mathbf{r},x,t,\pi),
\end{align}
which maps initial state-times to a state-time \textit{after} completing a goal variable under a policy in a hierarchical state-time product space $\mc S\times \mc T$ where $\mc S = \mc X \times \mc R$. As we will see later, in a hierarchical state-spaces under the dynamics of $G_s$, where goals induce dynamics on other state-spaces, goal-directed behavior on a low-level state-space emerges as a consequence of optimizing hierarchical empowerment. In such a case the \textit{Temporally-Extended Semi-Markov} \textit{Empowerment},
\begin{align*}
    \mathfrak{E}_n(G_s|\textbf{s}_t,t) = C(\Pi_{n};ST_n|\textbf{s}_t,t),
\end{align*}
is the maximum mutual information between a policy sequence R.V., $\Pi_{n}$, and the resulting $n$-step state-time R.V., $ST_n$. This empowerment only considers the final state-times under each policy and is thus defined on the \textit{task sub-space} of the agent's full product-space.

Empowerment has been proposed as a means of self-preservation, since $\mathfrak{E}_n=0$ implies ``death" at that horizon $n$, however it has been pointed out by Turner et al. that there is not necessarily a convergent limiting behavior when increasing the parameter $n$ \cite{turner2019optimal}. They argue that their metric, POWER, has the advantage that it takes into account an agent's discount factors (and therefore converges), and they explicitly tie POWER to an agent with infinite horizon discounted reward maximization as the agent's objective, showing that most reward-functions imply POWER seeking policies when optimizing for the objective. We do not make reward-maximization assumptions in our work, and do not see the absence of limiting behavior as problematic. Instead, future theory can be directed towards rationally choosing the horizon parameters. Empowerment has recently also been demonstrated in a task-centric setting help to robustly acquire reward under changing tasks \cite{volpi2020goal}. To our knowledge, our work is the first to combine empowerment with abstract transition operators and to explicitly optimize empowerment gain in the objective over long time-scales. 

\section{Temporal Goal Markov Decision Process}\label{sec:TG-MDP}

We begin be defining the Temporal Goal Markov Decision Process (TG-MDP), which is a Markov decision process \cite{puterman1990markov} for maximizing the probability of achieving a goal under finite-horizon non-stationary conditions. The TG-MDP will extend to hierarchical settings and will have nice properties for goal-conditioned policy decomposition which we will use to build up transition operator factorizations. The TG-MDP definition is given as:

\begin{mydef}{\textbf{Temporal Goal Markov Decision Process (TG-MDP)}}{}

A TG-MDP is a five-tuple $\mathscr{M}:=(\mc X,\mc A,\mc T, P_x,f_{\dg})$ where $\mc X$ is a discrete state-space, $\mc A$ is a discrete action space, $P_x:(\mc X\times \mc A \times \mc T)\times \mc X \rightarrow [0,1]$ is a transition operator, $f_{\dg}:(\mc X\times\mc A \times \mc T) \times \{\alpha_\dg\} \rightarrow [0,1]$ is a goal-availability function, and $\mc T=\{t_0,t_1,...,T_f\}$ is a set of discrete times with a horizon $T_f$.\label{def:TG-MDP}
\end{mydef}
The goal-availability function, $f_{\dg}(x,a,t)$, is the probability that a \textit{task-goal} variable $\alpha_\dg$ is \textit{available} to be completed at a given state, action, and time. In this paper, we will use the word \textit{goal} to refer to a \textit{variable} $\dg$, and a state this variable is associated with through $f_{\dg}$ is a \textit{goal-state} $x_{g}$. The task-goal is a variable, which in the context of hierarchical planning will be an action on a higher-order state-space. The TG-MDP is a decision process which is based on   \textit{reachability}, rather than cumulative reward-maximization.  Unlike a first-exit Markov decision process, in which the value function is defined to be the cumulative cost up until hitting a boundary state, we will instead compute the cumulative probability up until the event of being at a state and inducing the goal variable.
\subsection{Operator Bellman Equations}\label{sec:OBEs}
The TG-MDP, $\mathscr{M}$, parameterizes the three Operator Bellman Equations (OBE) given as,
\begin{align}
    &\underbrace{\kappa_{\dg}^*(x,t)}_{\mathclap{\text{Cumulative FF}}} = \max_{a\in \mc A} \left[f_{\dg}(x,a,t) + (1-f_{\dg}(x,a,t)) \sum_{x'} P_x(x'|x,a,t)\kappa_{\dg}^*(x',t+1) \right], \label{obs-C-Bellman-kappa}\\
    &\underbrace{\pi^{**}_{\dg}(x,t)}_{\mathclap{\text{Time-min Policy}}} = \argmin_{a\in \mc A^*_{x,t}}\left[tf_{\dg}(x,a,t)+(1-f_{\dg}(x,a,t))\expec_{~~\mathclap{x'\sim P_x}~~}~\sum_{x_f,t_f}t_f \eta^{**}_{\pi_g}(\alpha_\dg,x_f,t_f|x',t+1) \right],\label{eq:obs-c-Bellman-pol}\\
    &\underbrace{\eta_{\pi_g}^{**}(\alpha_\dg,x_f,t_f|x,t)}_{\mathclap{\text{State-time FF}}} = (1-f_{\dg}(x,a_{xt}^{**},t))\sum_{x'} \eta_{\pi_g}^{**}(\alpha_\dg,x_f,t_f|x',t+1)P_x(x'|x,a_{xt}^{**},t),\label{graveyard-marg-eta}
\end{align}
where $\kappa$ is the \textit{cumulative feasibility function} (the cumulative probability of achieving the goal $\alpha_\dg$ from $(x,t)$ under $\pi$), $\eta$ is the \textit{state-time feasibility function} (STFF) (the probability of achieving the goal at state-time $(x_f,t_f)$ when starting from $(x,t)$ under $\pi$), $\alpha_\dg$ is a single fixed variable called the \textit{task goal}, and $a_{xt}^{**}=\pi_{\dg}^{**}(x,t)$. Equation \eqref{graveyard-marg-eta} is defined for all $t_f$ such that $t<t_f<T_f$, and when $t_f=t$ it is defined as, $$\eta_{\pi_g}^{**}(\alpha_\dg,x_f,t_f|x,t)=f_{\dg}(x,a_{xt}^{**},t).$$ The boundary conditions for OBEs (\ref{obs-C-Bellman-kappa}-\ref{graveyard-marg-eta}) at the horizon time $T_f$ are given in A.\ref{appx:boundary}. The relationship between cumulative and state-time feasibility is intuitive, where the cumulative feasibility is the sum of all individual final state-time probabilities (see A.\ref{appx:kappaeta} for derivation): 
\begin{align*}
    \kappa_{\dg}^*(x,t) = \sum_{x_f,t_f}\eta_{\pi_\dg}^{**}(\alpha_\dg,x_f,t_f|x,t).
\end{align*}
The set $\mc A^*_{x,t}$ in equation \eqref{eq:obs-c-Bellman-pol} is the action set of equally maximizing arguments of equation (\ref{obs-C-Bellman-kappa}). Since it is possible to maximize cumulative feasibility by arriving at the goal either as soon as possible or later when it is still available, the policy $\pi^{**}$ is optimized with an additional conditional expected-time minimization (denoted with two stars $**$) over the set $\mc A^*_{x,t}$, which means the agent will take the shortest possible time that maximizes cumulative feasibility. Note that the time-minimization expectation is over the next-state distribution $P_x$ conditioned as $P_x(\cdot|x,a,t)$. The OBEs are finite-horizon and can be solved via backwards recursion with a simple algorithm called \textit{feasibility iteration} (\ref{alg:feasibility_iteration}), which is a direct analogue of finite-horizon value iteration \cite{bellman1957markovian}. The STFF represents the state-time-goal probability at policy termination, which is dictated by either task completion (inducing $\alpha_\dg$) or task-failure, inducing the \textit{null goal} $\alpha_\epsilon$. A task-failure termination event is defined as the last time the agent has a positive probability of completing the goal $\alpha_\dg$ under the policy. If $P_x$ and $f_{\dg}$ are deterministic and there is a single goal state $x_g$, we can set the task-failure probability as (See A.\ref{sec:task_failure} for derivation and details),
\begin{align*}
    \eta_{\pi_\dg}^{**}(\alpha_{\epsilon},x_f,t_f|x,t)=\{1-\kappa_{\dg}^*(x,t), ~~\text{if:}~~ (t_f = t_f^+) \land (x_f=x_g), ~~0~~ \text{otherwise}\},
\end{align*}
where $t_f^+$, which is obtained from $\eta^{**}_{\pi_g}(\alpha_\dg,x_f,t_f|x,t)$, is the last time the goal $\alpha_\dg$ can possibly be achieved under $\pi_{\dg}$ from $(x,t)$, inducing the null goal $\alpha_\epsilon$. If the task is impossible ($\kappa_{\dg}(x,t)=0$), then definitionally, $\eta_{\pi_\dg}(\alpha_{\epsilon},x,t|x,t)=1$, mapping the agent to its current state-time. With the failure probability computed, $\eta$ sums to $1$ over states, times, and goals, making it a transition operator with \textit{one} action, $\pi^{**}$, and we will see shortly how several STFFs can be aggregated together to create a transition operator with multiple actions (policies). Thus, unlike the infinite horizon discounted reward Bellman equations, the STFF OBE explicitly retains and propagate only the critical information for compositionality, both within a state-space (state-time mappings) and up a hierarchy of state-spaces (goals). Also, note that the termination conditions result from the optimization, unlike the Options framework \cite{sutton1999between} in which initiation sets and termination sets for an option are fixed parameters of the problem. Furthermore, unlike the Options framework, the TG-MDP is designed for non-stationarity and hierarchical planning.

\begin{algorithm2e}[h]
\DontPrintSemicolon
\SetKw{return}{return}
\SetKwRepeat{Do}{do}{while}
\SetKwData{conflict}{conflict}
\SetKwData{safe}{safe}
\SetKwData{sat}{sat}
\SetKwData{unsafe}{unsafe}
\SetKwData{unknown}{unknown}
\SetKwData{true}{true}
\SetKwInOut{Input}{input}
\SetKwInOut{Output}{output}
\SetKwFor{Loop}{Loop}{}{}
\SetKw{KwNot}{not}
\begin{small}
	\Input{Dynamics $P_x(x'|x,a,t)$, goal-availability function $f_{\dg}(x,a,t)$, final time $T_f$}
	\Output{Cumulative feasibility function $\kappa$, Policy $\pi$, State-time feasibility function $\eta$}
	define $nx$ as the number of states\;
	initialize $\eta_\dg,\eta_-\leftarrow zeros([nx*T_f,nx*T_f])$\;
    initialize $\kappa\leftarrow zeros([nx,T_f])$\;
    $\kappa(x_f,T_f) \leftarrow \max_a f_{\dg}(x_f,a,T_f)~\forall x_f \in \mc X,~~\# \text{CFF Boundary Conditions}$\;
    $\pi(x_f,T_f) \leftarrow \argmax_a f_{\dg}(x_f,a,T_f)~\forall x_f \in \mc X,~~\# \text{Policy Boundary Conditions}$\;
    $\eta_{\dg}(x_f,T_f|x_f,T_f) \leftarrow f_{\dg}(x_f,\pi(x_f,T_f),T_f),~\forall x_f \in \mc X~~\# \text{Success Boundary Conditions}$\;
    $\eta_{-}(x_f,T_f|x_f,T_f) \leftarrow 1-f_{\dg}(x_f,\pi(x_f,T_f),T_f),~\forall x_f \in \mc X~~\# \text{Failure Boundary Conditions}$\;
	\For{$t$ from $T_f-1$ to $0$}{
    	\For{$x \in \mc X$}{
    	$(max\_\kappa,\mc A^*_{x,t}) \leftarrow \argmax_{\mc A}\left[ f_{\dg}(x,a,t) + (1-f_{\dg}(x,a,t))\expec_{x'\sim P_x}\kappa(x',t+1)\right]$\;
    	$\kappa(x,t) \leftarrow max\_\kappa$\;
    	$\pi(x,t) \leftarrow a^{**}_{xt}\leftarrow \argmin_{a \in \mc A^*_{x,t}}\left[tf_{\dg}(x,a,t)+(1-f_{\dg}(x,a,t))\expec_{ P_x}\expec_{\eta_\alpha}t_f\right]$\;
        $\eta_{\dg}(x_f,t|x_f,t) \leftarrow f_{\dg}(x,a^{**}_{xt},t), ~~ \forall x_f \in \mc X$\;
        $\eta_\dg(x_f,t_f|x,t) \leftarrow (1-f_{\dg}(x,a^{**}_{xt},t))\expec_{x'\sim P_x}\eta_\alpha(x_f,t_f|x',t), \forall (x_f,t_f) \in \mc X \times \mc T_{>t}$\;
        $\# \text{Failure probability}$\;
        \uIf{$\kappa(x,t)>0$}{
        $\eta_{-}(x_f,t_f|x,t) \leftarrow \expec_{x'\sim P_x(\cdot|x,a^{**}_{xt},t)}\eta_-(x_f,t_f|x',t+1), \forall (x_f,t_f) \in \mc X \times \mc T_{>t}$\;
        }
        \Else{
         $\eta_{-}(x,t|x,t) \leftarrow 1$\;
        }
        
    	}
	}
   $\eta \gets \texttt{Combine}(\eta_\dg, \eta_-)$\;
    \texttt{return~}$\kappa$,$\pi$,$\eta$\;
\end{small}
\caption{Feasibility Iteration}
\label{alg:feasibility_iteration}
\end{algorithm2e}

The TG-MDP builds on the feasibility function introduced by Ringstrom et al. in their modular Goal-Operator Planning theory (2019, 2020) \cite{ringstrom2019constraint,ringstrom2020jump} which uses goal-conditioned transition goal-operators that transition the agent from initial-to-final state-times under a policy, and compliments work in RL with timed-subgoals \cite{gurtler2021hierarchical}. It differs from Ringstrom et. al (2020) in that the feasibility function is computed directly from the Bellman equation, rather than derived from a policy. It also shares similarities with first-occupancy RL which terminates on the first goal-state occupancy \cite{moskovitz2021first}. The TG-MDP, especially in the context of hierarchical state-spaces, is a new contribution to the area of goal-centric policy synthesis of which there are numerous examples in the literature \cite{kaelbling1993learning, ghosh2018learning, schaul2015universal, andrychowicz2017hindsight}, often going under the name goal-conditioned RL \cite{hoang2021successor, liu2022goal, chane2021goal,tang2021hindsight}, multi-goal RL \cite{plappert2018multi,colas2019curious}, reward-machines \cite{icarte2018using, camacho2019ltl, icarte2022reward}, and goal-operator planning \cite{ringstrom2019constraint, ringstrom2020jump}. We also believe that the OBEs could prove to be an important objective function generally for temporal logic problems such as Linear Temporal Logic (LTL) \cite{yuan2019modular,hasanbeig2019reinforcement,hasanbeig2020towards}, especially considering the limitations of infinite-horizon objectives for LTL \cite{yangtractability}, and that LTL formulae can be expressed as high-level state-spaces. We do not address LTL here, but we will demonstrate examples of Boolean logic tasks which have time-constraints in our hierarchical TG-CMDP formulation, which is not as expressive as LTL, but can be considered to be a kind of temporal logic. The "C" in TG-CMDP stands for \textit{composition}, and indicates that the TG-MDP is defined on a composition of transition operators.
 
A TG-CMDP has a strong connection to program synthesis \cite{yang2021program,nye2019learning,ellis2021dreamcoder} and policy sketching \cite{andreas2017modular} given that it is formed on compositions of transition functions. There is also a new emerging area of research called \textit{Compositional RL} (CRL) which emphasizes the composition of task modular high-level tasks for policy synthesis\cite{jothimurugan2021compositional, neary2022verifiable}. However, CRL tasks do not have time-varying goals or environments, nor do they provide global model-based Bellman equation foundations with compositional hierarchical transition operators. In this paper, a TG-CMDP serves this very purpose for synthesizing non-stationary non-Markovian problems with a composition operator and a corresponding goal-conditioned policy decomposition. Lastly, the TG-MDP provides new representations for reasoning about time, which is growing area of interest in deep RL \cite{lampinen2021towards}. 
\subsection{Hierarchical TG-CMDPs for a Self-Preserving Agent}\label{sec:Hierarchical OBEs}
We can extend the TG-MDP to hierarchical state-spaces by composing transition operators into a Cartesian product transition operator $P_s$, and we will refer to this as the Temporal-Goal Compositional-MDP, (TG-CMDP). Let $\mc W = \{w_0,...,w_{max}\}$ be a hydration state space, $\mc Y = \{y_0,...,y_{max}\}$ be a caloric energy state-space, and $\mc Z = \{z_0,...,z_{max}\}$ be a temperature space, with each space paired with an action space $\mc A_w = \{\alpha_{drink},\alpha_{\epsilon}\}$, $\mc A_y=\{\alpha_{eat},\alpha_{\epsilon}\}$, $\mc A_z=\{\alpha_{warm},\alpha_{\epsilon}\}$, where each set has an active action (e.g. $\alpha_{eat}$) that modifies an internal state (e.g. drinking water increases the state to the max value), and a null action $\alpha_{\epsilon}$, which induces the natural uncontrolled dynamics of the space when the agent does not directly influence it (e.g. the agent gets hungrier over time). Also, let the transition operators $P_w$, $P_y$, and $P_z$ have the same chain-like structure shown in Figure \ref{fig:Modes-and-time} and \ref{fig:Homeostatic Control 1}, where (taking $P_y$ as an example) the task-goal causes the agent's state to jump to the top state $y_{max}$, and the null goal causes the agent to descend from $y_{i}$ to $y_{i-1}$ until $y_0$.  The set of states $\mc D = \{w_0,y_0,z_0\}$ are \textit{defective states}, which will inhibit the agent's low-level movement.

\begin{figure}[h]
\centering
\includegraphics[scale=0.5]{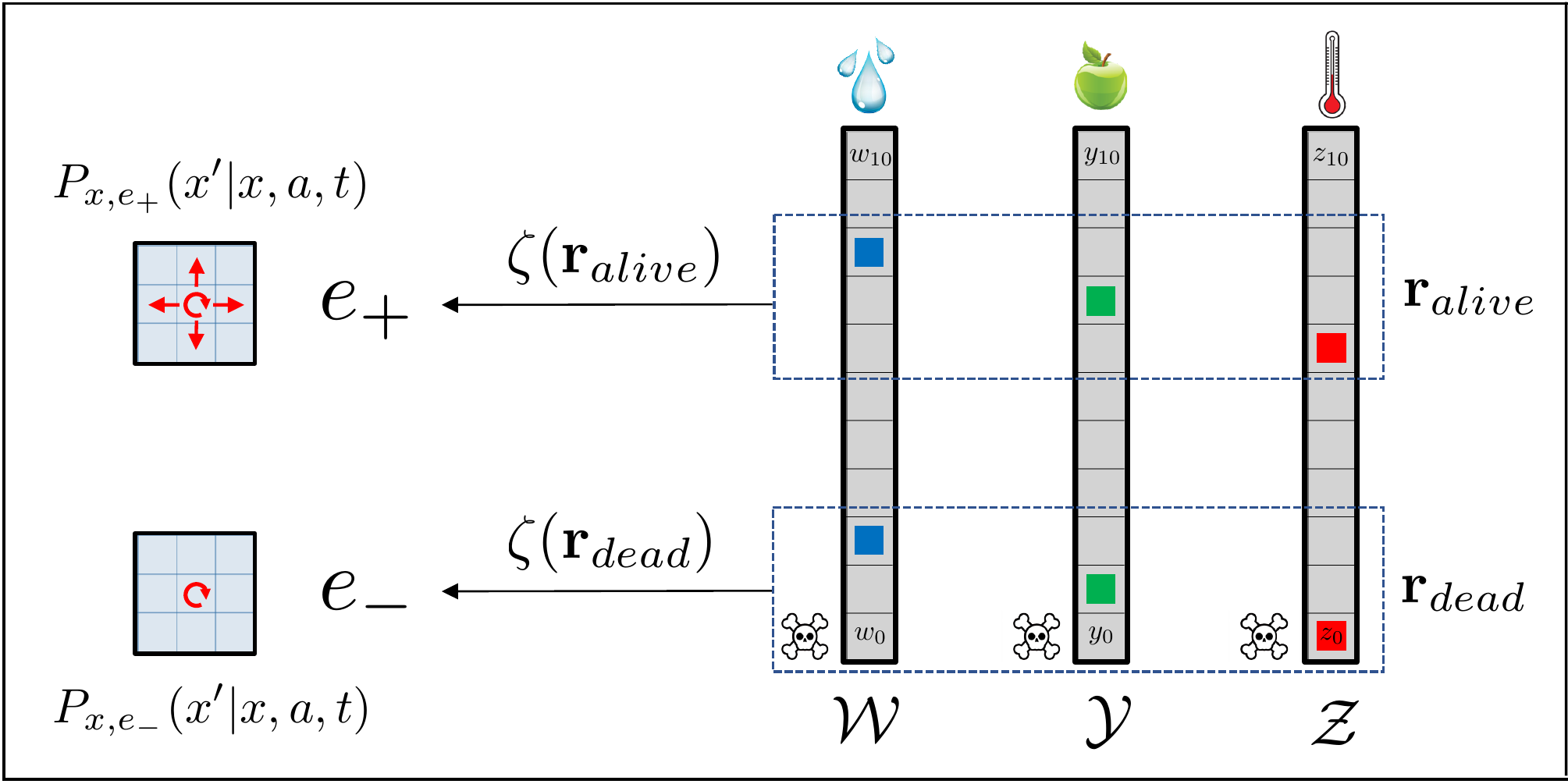}
\caption{Mode Mapping: State-vectors that have no defective components map to the normal grid-world dynamics mode. If a state-vector has a defective component, it maps to the defective dynamics mode in which all actions are identity transitions.}
\label{fig:Modes-and-time}
\end{figure}

Now let $P_x: (\mc X \times \mc A \times \mc T \times \mc E)\times \mc X \rightarrow [0,1]$ be a transition operator which is indexed by a dynamics mode variable $e \in \{e_{+},e_{-}\} = \mc E$.  The modes partition the $x-$dynamics into a normal mode indexed by $e_{+}$ (e.g. gridworld dynamics) and a defective mode indexed by $e_{-}$ where the dynamics will be a simple identity map $x_i\xrightarrow{a} x_i$ for all state-actions, meaning the agent cannot move. We then can split the operator $P_x$ into $P_{xe_+}(x'|x,a,t)$ and $P_{xe_-}(x'|x,a,t)$.  We provide a \textit{mode function}, $\zeta:\mc R \rightarrow \mc E$, where $\mathbf{r}$ is an internal state vector, $\mathbf{r}\in \mc R =\mc W \times \mc Y \times \mc Z$; the mode function partitions the internal vectors into disjoint subsets which condition low-level dynamics by a mode variable $e$ (see figure \ref{fig:Modes-and-time} for an illustration). We call a mode function \textit{element-invariant} if all states (e.g. $w \in \mc W$) of a state vector $\br=(w,y,...)$, map to a consistent mode $e$ when other states in the vector change, meaning there is no pairwise interaction between states in the vector that determine the mode. The variable $e$ transitions from $e_+$ to $e_-$ when the agent enters any defective state in $\mc D$ and transitions from $e_-$ to $e_+$ when an agent stops occupying a defective state. If the agent hits any defective state, it is immobilized on $\mc X$ and effectively becomes a self-absorbed Markov chain, where it cannot complete tasks to transition itself out of the defective state unless it is already at a goal state. 

For compactness let $P_r: (\mc R \times \mc A^{1}_{r})\times \mc R \rightarrow [0,1]$, be defined as:
\begin{align*}
    P_r(\mathbf{r}'|\mathbf{r},\boldsymbol{\alpha}) := P_w(w'|w,\alpha_w)P_y(y'|y,\alpha_y)P_z(z'|z,\alpha_z),~~~~\mathbf{r}=[w,y,z], ~\boldsymbol{\alpha}=(\alpha_w,\alpha_y,\alpha_z).
\end{align*}
    where $\mc A^{1}_r$ is a set of action vectors that induce dynamics on $\mc R$, where each vector $\boldsymbol{\alpha}\in \mc A^{1}_r$ has elements that are actions for a higher-level space. We use the convention that the superscript, $1$, refers to when the action set belongs to a state-space linked to the base-state space $\mc X$ (a ``distance" of one); later, $\mc A^{2}$ will belong to a state-space a distance of two from $\mc X$.

Let $\mc S = \mc R \times \mc X$, let $\mathbf{s} =\mathbf{r}\circ x$ be a concatenated vector in $\mc S$, and let $F:(\mc X \times \mc A \times \mc T) \times \mc A_r^1 \rightarrow [0,1]$ be a \textit{action-availability function}, which is a probability mass function where $\sum_{\boldsymbol{\alpha}\in \mc A_r^1}F(\boldsymbol{\alpha}|x,a,t)=1$. The action-availability function dictates the state-action-time to goal probabilities for all goals that parameterize a higher-level transition system (see figure \ref{fig:Homeostatic Control 1}). We can then use it to link together low- and high-level spaces, the full fine-grained product dynamics $P_s$ are defined by a composition operator $\lambda_P$ applied to the functions $(P_r,F,\zeta,P_x)$:
\begin{align}
    P_s(\mathbf{s}'|\mathbf{s},a,t)&\equiv P_s(\mathbf{r}',x'|\mathbf{r},x,a,t)\label{eq:hierarchical_trans_op}\\    \nonumber&=\lambda_P(P_r,F,\zeta,P_x):=\sum_{\boldsymbol{\alpha}}P_r(\mathbf{r}'|\mathbf{r},\boldsymbol{\alpha})F(\boldsymbol{\alpha}|x,a,t)P_x(x'|x,a,\zeta(\mathbf{r}),t).
\end{align}

Notice, due to the bidirectional coupling, this transition system is intrinsically self-undermining in the absence of computational work applied to computing a goal directed control policy, and it will serve as the agent's ontology, which it can ground it's value judgements in without reward. The composition operator $\lambda_P$ should have the form $\lambda_P((secondary),(linking), base)$ where, if there are multiple secondary operators, they are assumed to be independent (e.g. as we did with $P_r$).

We can formalize a hierarchical variant of the TG-CMDP, $\bar{\mathscr{M}}$, using (\ref{def:TG-MDP}) along with the composition function that builds a hierarchical transition operator: 
\begin{align*}
    \bar{\mathscr{M}}=(\mc S, \mc A, \lambda_P(P_r,F,\zeta,P_x),\bar{f}_{\dg})=(\mc S,\mc A,P_s,\bar{f}_{\dg}),
\end{align*}
where $\bar{f}_{\dg}$ encodes the acceptable task conditions. The functions of $\bar{\mathscr{M}}$ can parameterize a hierarchical OBE (along with the omitted policy and state-time feasibility function equations):
\begin{align}
    &\bar{\kappa}_{\dg}^*(\textbf{r},x,t) = \max_{a\in \mc A} \left[\bar{f}_{\dg}(\textbf{r},x,a,t) + (1-\bar{f}_{\dg}(\textbf{r},x,a,t)) \sum_{\textbf{r}',x'} P_s(\textbf{r}',x'|\textbf{r},x,a,t)\bar{\kappa}_{\dg}^*(\textbf{r}',x',t+1) \right]. \label{eq:h-Bellman-kappa}
\end{align}
However, because the size of $\mc S$ grows exponentially as we add additional higher-order state-spaces, two problems arise when we model problems in the TG-CMDP formalism: (A) computing closed-loop policies on these spaces quickly becomes intractable due to the prohibitive space- and time-complexity, and (B) $\bar{f}_{\dg}$ becomes challenging to intuitively \textit{define} over a product space because, unlike a human engineer, an autonomous agent without a capacity of \textit{reasoning and justification} does not know \textit{a priori} which state-vectors in a product space are good and which are bad. However, with regards to the former problem, we show in the next section (\ref{sec:forward}) that we can derive an efficient planning factorization. Then we can use the factorization to forecast final state-times $(\mathbf{r}_{t_f},x_{t_f},t_f)$ produced from \textit{plans} (sequences of polices) starting from an initial state-time $(\mathbf{r},x,t)$. The later problem (B) will be addressed in sections \ref{sec:self-preserving} and section \ref{sec:VBE} where we show how, rather than formalising the problem as the TG-CMDP above, it is better formalized as a \textit{valence maximization} objective parameterized by TG-CMDP solutions $\bar{\eta}$ of equation \eqref{eq:h-Bellman-kappa}, using the operator factorization that we will now discuss.

\subsubsection{TG-CMDP Planning Factorization}\label{sec:factor}
Consider equation \eqref{eq:h-Bellman-kappa} where $P_s$ only has a unidirectional coupling (no $\zeta$). That is, $P_s = \lambda_P(P_y,F,P_x)$, where $P_x$ does not have mode parameters $\mc E$ in its domain. Also we define $F(\alpha|x,a,t)$ to be a \textit{homogeneous} multi-goal function when it is defined on $\mc A^{1} = \{\alpha_\dg,\alpha_{\epsilon}\}$, only having \textit{one} non-null goal, $\alpha_\dg$, in $\mc A^{1}$. 

Given that $\eta_{\pi_{g}}(\alpha_\dg,x_{g},t_f|x,t)$ specifies the time that it takes, $t_d=t_f-t$, to complete the goal variable $\alpha_\dg$ under $\pi_{e,g}$, the agent will produce null actions $\alpha_{\epsilon}$ from the homogeneous availability function $F(\alpha|x,a,t)$ for states in its trajectory until the goal $\alpha_\dg$ is induced at $(x_g,t_f)$.
Therefore, we can obtain an \textit{internal state prediction operator}, $\omega$, which returns the future internal state when the goal $\alpha_{+}$ is first activated, by evolving the null Markov chain on a given space for $t_d$ time-steps. Taking $\mc Y$ for example, if we let $P_{y\epsilon}(y'|y)$ be the Markov matrix by setting the action of $P_y(y'|y,\alpha)$ to $\alpha_{\epsilon}$ for all possible inputs $y$, then we can compute $\omega$ as a matrix power (see figure \ref{fig:Homeostatic Control 1} for illustration):
\begin{align*}
\underbrace{\omega_y(y_f|y_i,t_d)}_{\mathclap{\text{Prediction Operator}}}~~:=~~ \underbrace{P_{y\epsilon}^{t_f-t}(i,f)}_{\mathclap{\text{Null Markov chain}}},\quad\quad \underbrace{(t_f,t)\leftarrow\eta_{\pi}(\alpha_\dg,x_f,t_f|x,t)}_{\text{Time duration from STFF}} 
\end{align*}
where $(i,f)$ indexes the element of the evolved matrix chain matrix, corresponding to the initial state $y_i$ and final state $y_f$. Furthermore, if we have multiple independent high-level state-space transition operators, we can compute a set, $\Omega = \{\omega_w, \omega_y, \omega_z\}$, of internal prediction operators for each internal state-space, and use it to define an \textit{internal prediction operator} $\omega_{r}$ on the vector $\mathbf{r}$, 
\begin{align}
    \nonumber
    \omega_{r}(\mathbf{r}'|\mathbf{r},t_d):=\omega_w(w'|w,t_d)\omega_y(y'|y,t_d)\omega_z(z'|z,t_d),\quad \mathbf{r}:=[w,y,z],~ \mathbf{r}':=[w',y',z'],
\end{align}

We can then state the unidirectional decomposition. If $\bar{\mathscr{M}}=(\mc Y \times...\times \mc Z\times \mc X, \mc A, P_s,\bar{f}_{\dg})$ is a hierarchical TG-CMDP with $P_s = \lambda_P(P_y,...,P_z,F,P_x)$, homogeneous $F$, and where $\bar{f}_{\dg}(x,a,t)$ is not a function of $(y,...,z)$. Then the optimal STFF $\bar{\eta}^{**}_{\pi}$ is equivalent to:
$$ \bar{\eta}^{**}_{\pi}(\alpha_\dg,x_f,y_f,...,z_f,t_f|x,y,...,z,t)=\omega_y(y_f|y,t_f-t)...\omega_z(z_f|z,t_f-t)\eta^{**}_{\pi}(\alpha_\dg,x_f,t_f|x,t)$$
where (e.g.) $\omega_y(y_f|y,t_f-t)=P_{y,\epsilon}^{t_f-t}(y,y_f)$ where $P_{y,\epsilon}$ is the null-goal Markov chain derived from $P_y$, and $\eta^{**}_{\pi}$ (no bar) is the STFF for a flat TG-MDP $\mathscr{M}=(\mc X,\mc A,P_x,\bar{f}_{\dg})$. The equation can alternatively be written with $\omega_r$ and $\mathbf{r}=[y,...,z]$ as:
\begin{align}
    \bar{\eta}^{**}_{\pi}(\alpha_\dg,\mathbf{r}_f,x_f,t_f|\mathbf{r},x,t)=\omega_r(\mathbf{r}_f|\mathbf{r},t_f-t)\eta_{\pi}^{**}(\alpha_\dg,x_f,t_f|x,t).\label{eq:unidirectional-decomp}
\end{align}




\begin{figure}[t]
    \centering
    \includegraphics[scale=0.36]{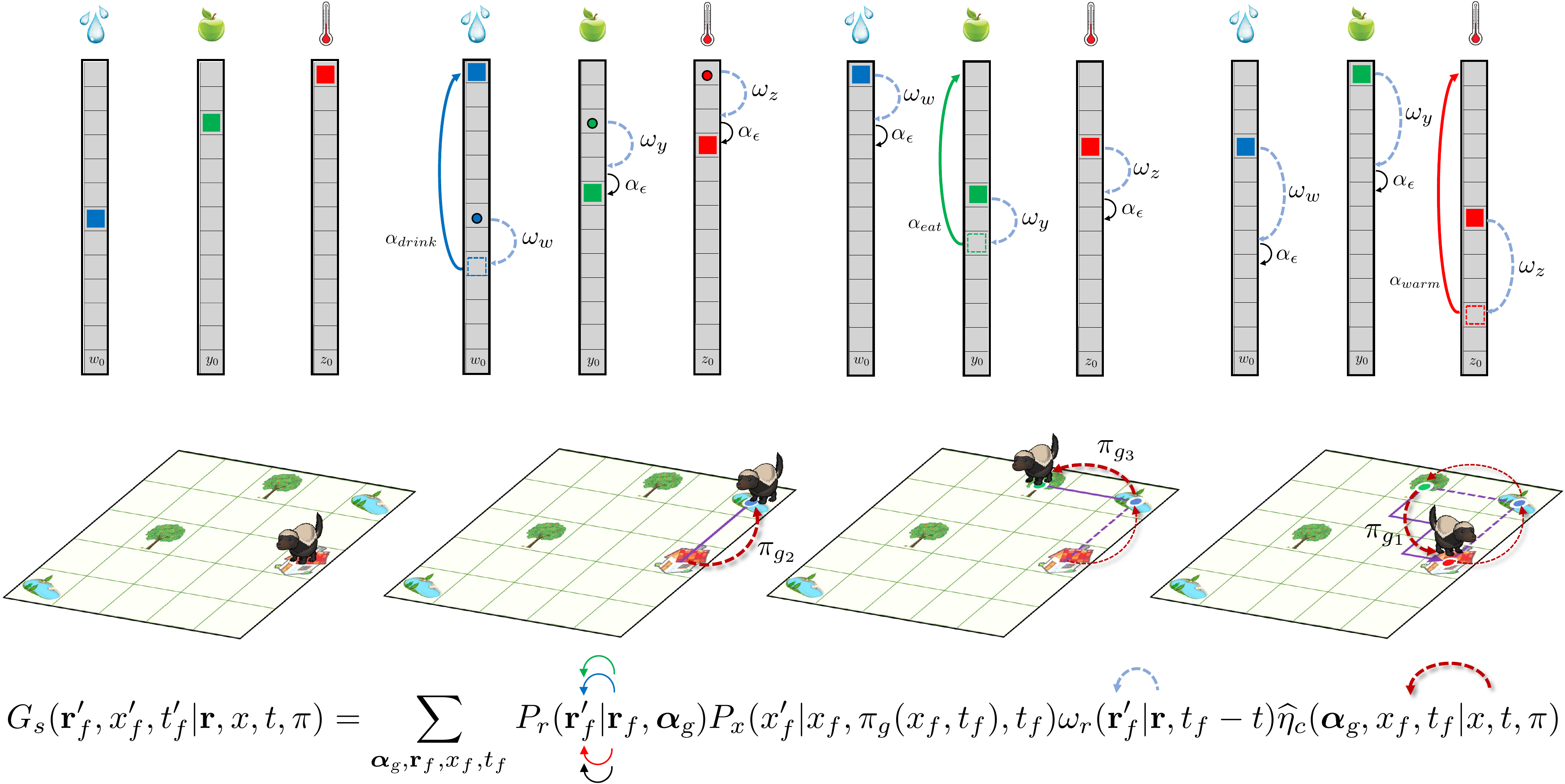}
    \caption{Stoffel uses the planning factorization to execute the plan $\rho_{1,3,2}=(\pi_{g_2},\pi_{g_3},\pi_{g_1})$ while updating the low-level states with $\widehat{\eta}$ and physiological states with $\omega$ predictions, each which map initial states to final states (and times) with an dashed arc, followed with a one-step update indicated by a solid arc.}
    \label{fig:TG-CMDP Examples}
\end{figure}

This result is a special case of the bidirectional TG-CMDP decomposition theorem (\ref{bidirectional-thm}), which we will state and prove shortly. The significance of this decomposition is that we only need to compute the goal-conditioned policy $\pi$ and STFF $\eta^{**}_{\pi}$ on the low-level state-space (not the product space), and we can compute the higher-level prediction operator $\omega_r$ separately on each individual high-level space. By forming this factorization, we can forward sample sequences of policies. In figure \ref{fig:TG-CMDP Examples} we show how this allows us to chain together sequences policies and predict the resulting state-vectors after following each policy. This paper will only consider scenarios where the homogeneity assumption holds, and theory for when it does not hold will be addressed in future work.


\subsubsection{Time-to-go Restriction for Bidirectional Coupling}

We introduced the planning factorization assuming we only had a unidirectional coupling, but our original construction had a bidirectional coupling where $\zeta$ mapped the state of $\textbf{r}$ down to a mode parameter $e$ which conditioned the operator $P_{x,e}$. That is, the dynamics of $\textbf{r}$ will determine a deadline on how long we can use $P_{x,e}$ before the mode changes. Therefore, solving an OBE restricted to the low-level space might seem improper in the bidirectional case because dynamics on $\mc X$ depend on $\mc R$, which would require us to track all of the internal states in order to index the correct mode. However, because the internal state-spaces evolve independently of $\mc X$ the policy will only produce null actions $\alpha_\epsilon$ for the intermediate states of the trajectory due to the homogeneity assumption. Thus, a computed policy which is restricted to a single mode will be \textit{valid} up to a time-constraint given as the amount of time before any component of the vector $\mathbf{r}$ becomes defective under the sequence of null goals. This is determined by the \textit{minimum} first-hit time over the set of ``null Markov chains", given by the linear systems \cite{bremaud2013markov},
\begin{align*}
    \mathbf{1}=(I-\bar{P}_{w\epsilon})\mathbf{t}_w,\quad\quad \mathbf{1}=(I-\bar{P}_{y\epsilon})\mathbf{t}_y,\quad\quad\mathbf{1}=(I-\bar{P}_{z\epsilon})\mathbf{t}_z,
\end{align*}
where (e.g.) $\bar{P}_{w\epsilon}$ is a square matrix formed by setting the actions of $P_w(w'|w,\alpha)$ to $\alpha_{\epsilon}$, and the overbar indicates that the rows and columns corresponding to the defective states of $\mc W$ are deleted from the matrix. The vector $\mathbf{1}$ is a column of ones, and $\mathbf{t}_w$ is the unknown vector of first hit times where $\mathbf{t}_w(w_i)$ is the time it takes for an agent at $w_i$ to hit a defective state $w_d\in \mc D$. Each of these linear systems represent a solution of the expected first-hit time under the Markov chain, where by rearranging terms things become more clear: we see that $\mathbf{t}_w=M_w\mathbf{1}$, where $M_w=(I-\bar{P}_{w\epsilon})^{-1}$ (notice that the vector of ones sums up the expected total state-occupancies of each row). Here, the matrix inverse is the Fundamental Matrix $M_w$ for the absorbing chain $\bar{P}_{w\epsilon}$, which represents the expected total state-occupancies in transit to a defective state given the agent starts at $w_i$, index by the rows in $M_w$ \cite{bremaud2013markov} (cf. the successor representation \cite{dayan1993improving}, which does not have a terminal state and is discounted in time).\footnote{In the case that the internal state-spaces are chains with an descending one-step increment null dynamics, and there is one defective state at the bottom of the chain, then the hitting time is given by the index in the state array (e.g. $y_2$ is $2$ time-steps away from the defective state $y_0$ under the null dynamics).  The linear system, however, covers all hitting-time cases with many possible defective states or state-space transition structures.} It should also be emphasized that $\textbf{t}$ represents the \textit{expected} defective-state hitting time, which under deterministic transitions is the exact hitting time. If the high-level space has stochastic dynamics, then one would need to construct the first-hitting time \textit{distribution} by forward evolving the null Markov chain, and use this distribution to set the feasibility function appropriately. We will only consider deterministic high-level dynamics here for simplicity.

Computing the time-to-mode-switch set, $\mc T_r := \{\mathbf{t}_w,\mathbf{t}_y,\mathbf{t}_z\}$,
allows us to make an important restriction on which unconstrained policies are valid to follow. 
We can now state the bidrectional TG-CMDP decomposition, which will allow us to construct a factorization that is constrained to polcies that are \textit{valid} under an given $\br$, respecting the time-to-go values in the set $\mc T_r$:

\begin{mytheo}{Bidirectional TG-CMDP Decomposition}{theoexample}

Let $\bar{\mathscr{M}}^b=(\mc Y \times...\times \mc Z\times \mc X, \mc A, P_s,\bar{f}_{\dg})$ be a TG-CMDP where $\mc R=\mc Y \times...\times \mc Z$, $P_s = \lambda_P^b(P_r,F,\zeta,P_x)$, homogeneous $F$, element-invariant $\zeta$, deterministic $P_r$, and where $\bar{f}_{\dg}(x,a,t)$ is not a function of $\textbf{r}=(y,...,z)$. Let the mode-switch time set be $\mc T_r = \{\mathbf{t}_w,\mathbf{t}_y,\mathbf{t}_z\}$. If $t_f-t\leq\min_{\mathbf{t}_j\in \mc T_r} \mathbf{t}_j(\br_j),$ then the optimal STFF $\bar{\eta}^{**}$ is equivalent to:
\begin{align}
    \bar{\eta}^{**}(\ba_\dg,\textbf{r}_f,x_f,a_f,t_f|\textbf{r},x,t)&=\omega_y(y_f|y,t_f-t)...\omega_z(z_f|z,t_f-t)\eta^{**}_{\pi,e_\br}(\ba_\dg,a_f,x_f,t_f|x,t)\label{eq:bidrec-eq}\\
    &=\omega_r(\textbf{r}_f|\textbf{r},t_f-t)\eta^{**}_{\pi,e_\br}(\ba_\dg,x_f,a_f,t_f|x,t),
\end{align}
where $\omega_r(\br_f|\br,t_d):=\omega_y(y_f|y,t_d)...\omega_z(z_f|z,t_d),$
and the STFF $\eta^{**}_{\pi,e_\br}$ is computed solely on $P_x(x'|x,a,t,e_\br)=P_{x,e_\br}(x'|x,a,t)$ fixed to the mode $e_\br \gets \zeta(\br)$.\label{bidirectional-thm}
\end{mytheo}

This theorem is proved in Appendix \eqref{appx:bidirectional_proof}, and it reduces to the unidirectional equation \ref{eq:unidirectional-decomp} as a special case where the mode-switching time-to-go condition can be dropped due to the absence of a $\zeta$ function. Here, $j$ on $\mathbf{r}_j$ indexes the corresponding state in the vector, e.g. $\mathbf{t}_w(\mathbf{r}_w)\equiv\mathbf{t}_w(w),~ w \in [w,y,z]=\mathbf{r}$. Note that the condition $t_f-t\leq\min_{\mathbf{t}_j\in \mc T_r} \mathbf{t}_j(\br_j)$ in the theorem means that this only holds if the agent can reach the goal before a mode-switching event. In figure \ref{fig:Modes-and-time} we illustrate this time-constraint: the policy call $\pi_{g_3}$ from state $x_{g_1}$ violates the time constraint dictated by $\mathbf{r}$, and the policy $\pi_{g_2}$ does not, and the transitions of $\widehat{\eta}$ are set accordingly. This theorem applies to one STFF for completing a single goal.
We will next show how we can aggregate many individual solutions into a larger operator and use it for multi-goal planning.

\begin{figure}[h]
\centering
\includegraphics[scale=0.5]{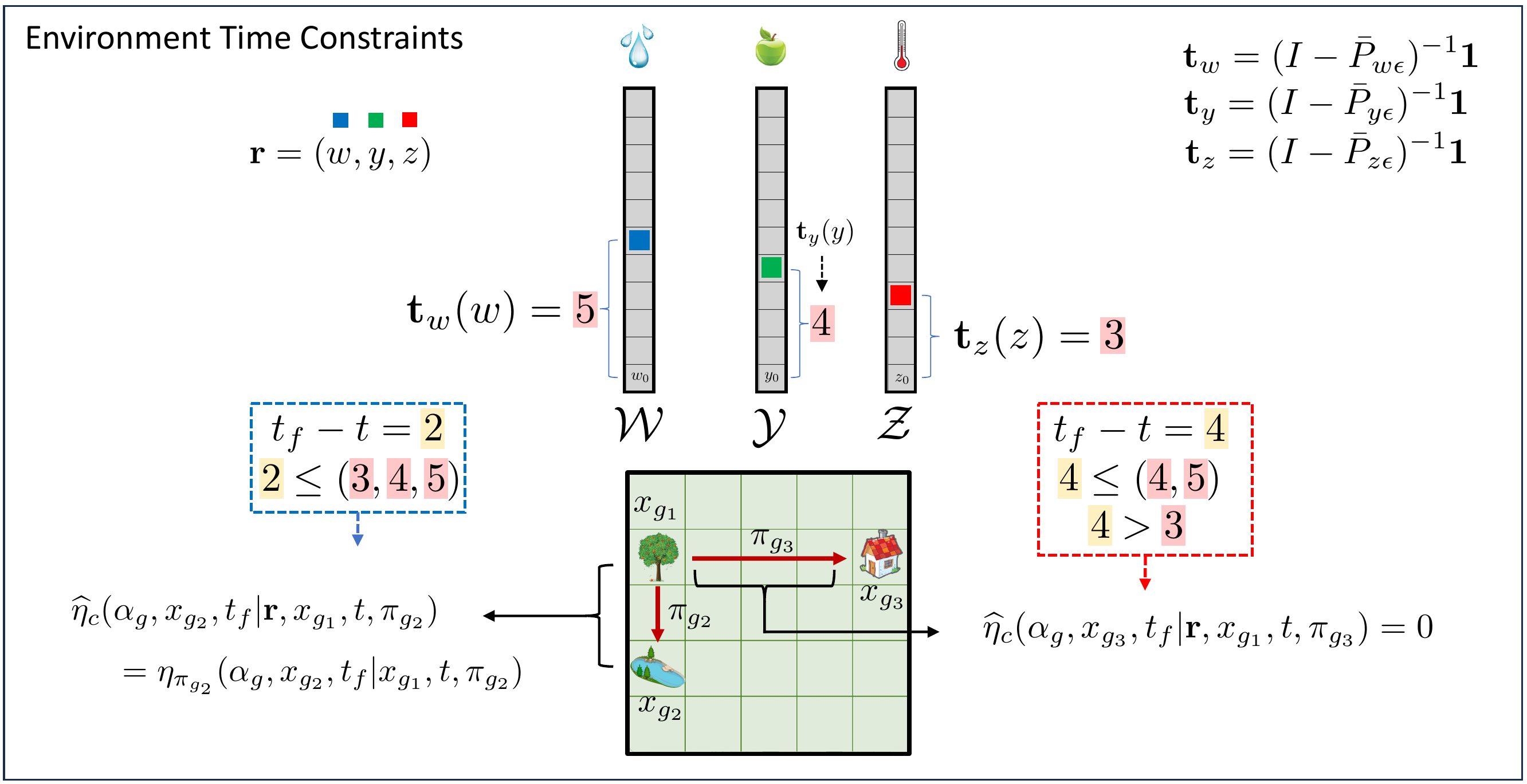}
\caption{Time Constraints: the time vectors $\textbf{t}_w$, $\textbf{t}_y$ and $\textbf{t}_z$ are computed and compared to the time it takes to navigate to goal states.  It takes 2 time steps to get to the lake from the tree, and so the agent can successfully get there without violating philosophical constraints. If the agent goes from the tree to the house, which takes 4 steps, it will not starve or die of dehydration, but it will will freeze to death, so the task of getting to the house is infeasible from the tree.}
\label{fig:Modes-and-time}
\end{figure}

\subsubsection{Forward Sampling With a Goal-conditioned Policy Set}\label{sec:forward}
Because the explicit operator $P_s$ is very large, computing solutions to equation \eqref{eq:h-Bellman-kappa} with either dynamic programming or forward-sampling low-level actions is not practical.  This is a significant problem for any agent which represents a hierarchy of transition operators. However, we can exploit the time-constrained TG-CMDP factorization discussed in the last two subsections to forward sample \textit{policies} from an ensemble of solutions for each goal (similar to \cite{ringstrom2019constraint}), where we compute policies and low-level STFFs separately from the higher-order transition operators and combine them with the higher-order state predictions $\omega$. 

Let,
\begin{align*}
    \pmb{\mathscr{M}}:=\{\mathscr{M}_{e_1,g_1},...,\mathscr{M}_{e_m,g_n}\},
\end{align*}
be an ensemble of TG-MDPs with an availability function for  each of the low-level goal states in $x$-component domain of $F$, i.e. $x_g \in \mc X_g = Dom_x(F)$, paired with each mode in $\mc E$. Each $\mathscr{M}_{e_j,g_i} = (\mc X,\mc A,P_{x e_j},f_{g_i})$ is a TG-MDP which is optimized using operator $P_{x e_j}$ and the availability function $f_{\dg}(x,a,t):=F(\alpha_\dg|x,a,t)$ is a restriction of the action-availability function $F$ to a single goal $\alpha_g\in \mc A^{1}$ corresponding to a single goal-state $x_g\in \mc X_g$, the set of states associated with task-goals. Solving the regular OBEs for each $\mathscr{M}_{e_j,g_i}$ in $\pmb{\mathscr{M}}$ will result in \textit{unconstrained} goal-cover sets, $\Pi_u := \{\pi_{e_1,g_1},...,\pi_{e_m,g_n}\}$ and $\mc H_u := \{\eta_{\pi_{e_1,g_1}},...,\eta_{\pi_{e_m,g_n}}\}.$
We can then define a \textit{constrained aggregate state-time feasibility function} $\widehat{\eta}_c$ using equation \eqref{eq:bidrec-eq} applied to the STFFs in $\mc H_u$:
\begin{align*}
    \widehat{\eta}_{c}(\boldsymbol{\alpha}_\dg,x_f,t_f|\mathbf{r},x,t,\pi_{\zeta(\mathbf{r}),\dg}) := \eta_{\pi_{e_\br},\dg}(\boldsymbol{\alpha}_\dg,x_f,t_f|x,t),\quad \forall \br : t_f-t\leq\min_{\mathbf{t}_j\in \mc T_r} \mathbf{t}_j(\br_j)
\end{align*}
Notice here that even though $\widehat{\eta}_{c}$ is parametrically defined on $\mc R \times \mc X \times \mc T$, it is defined to always output an STFF defined only on $\mc X \times \mc T$, and thus we are avoiding representing an object in the full product-space. We can now compose the aggregate operator $\widehat{\eta}_{c}$ with $\omega_{r}$ for the joint dynamics. Given that the STFF maps to an action (goal), in order to sequentially compose it we have to define the \textit{product-space goal-operator} $G_s:(\mc X \times \mc R \times \Pi \times \mc T)\times (\mc X \times \mc R \times \mc T)\rightarrow [0,1]$ defined with the policy-prediction composition operator $\lambda_G$ as,
\begin{align}
&G_s(\mathbf{r}_f',x_f',t_f'|\mathbf{r},x,t,\pi_{\zeta(\textbf{r})})=\lambda_G(P_r,P_x,\omega_r,\widehat{\eta}_c):= \label{eq:jump-op}\\
\nonumber&\sum_{\boldsymbol{\alpha},x_f,\mathbf{r}_f, t_f}\underbrace{P_r(\mathbf{r}_f'|\mathbf{r}_f,\boldsymbol{\alpha})P_x(x_f'|x_f,\pi(x_f,t_f),\zeta(\mathbf{r}_f),t_f)\omega_{r}(\mathbf{r}_f|\mathbf{r},t_f-t)\widehat{\eta}_{c}(\boldsymbol{\alpha},x_f,t_f|\mathbf{r},x,t,\pi_{\zeta(\textbf{r})})}_{\text{Goal-Operator Factorization}}
\end{align}
where $t_f'=t_f+1$, and we use $P_r$ and $P_x$ to evolve the dynamics one extra step so that the operator $G_s$ represents what happens \textit{after} the goal $\alpha_g$ is activated (hence the $t_f+1$) (See Fig.\ref{fig:Homeostatic Control 1} for operator illustrations). Also note that $\zeta$ and $\pi$ are implicit supporting functions in the arguments of $\lambda_G$, suppressed in the notation.

With $G_{s}$, we have a composite goal-operator which respects mode-switching constraints. Crucially, this operator has a memory-efficient factorization defining the agent's ontology, $$\mc O = \{\widehat{\eta}_{c},\omega_{w},\omega_{y},\omega_{z},P_w,P_y,P_z,P_x\},$$ that we can use to forward-sample state vectors without explicitly representing the Cartesian product space of $G$ in memory, which would be otherwise intractable. 

It is also important to note that if the higher-order state-space is \textit{static}, meaning that each null action induces an identity transition $y\xrightarrow{\alpha_{\epsilon}} y$, for all states $y$, then $\omega$ must have identity transitions for every state for all time $t_d$ provided as an input. This simplifies things, and it will be used when we introduce static binary states $\boldsymbol{\sigma}$ for logical tasks, or to parameterize the state of an environment. For example, these binary states and can be flipped under a transition operator $P_\sigma$ when the agent presses a button or picks up an item, and so the prediction operator must necessarily be given as, 
\begin{align*}
    \omega_{\sigma}(\boldsymbol{\sigma}'|\boldsymbol{\sigma},t_d)=\{1~~\text{if: } \boldsymbol{\sigma}=\boldsymbol{\sigma}', ~~0~~\text{otherwise}\},~ \forall t_d\in \mathbbm{N}_{0},
\end{align*}
due to the static property of the state-space.

Because the operator $G_s$ is indexed by polices, we can apply any tree-search algorithm, such as breadth-first search (BFS), to generate $m$-length \textit{plans} (open-loop sequences of closed-loop policies) $\rho = (\pi^{(1)},\pi^{(2)},...,\pi^{(m)})$ from the set $\Pi$, shown in algorithm \ref{alg:environment_update}\footnote{Note that algorithm (\ref{alg:environment_update}) is written assuming deterministic dynamics for simplicity. If stochastic dynamics are used, the variables have to be represented as probability distributions over states (e.g. $(\vec{\mathbf{t}},\vec{\mathbf{y}},...)$).}. Given an initial state-time $(\mathbf{s},t)_{i}$, we can initiate a root node, $\texttt{root} = \texttt{Node}(\mathbf{s}_{i},t_{i},\rho = (),\texttt{is\_leaf}=\texttt{False})$. Then executing, $BFS(\texttt{root}, \mc O, m)\rightarrow \texttt{leaves}$, will generate leaves by advancing the state with policies $\pi \in \Pi$ applied to the operator set $\mc O$ for every level in the tree. It is straightforward to see that a plan can be used to define a \textit{plan operator}, $Q$. For example, the entries of $Q$ for a length-2 plan $\rho_{1,2}=(\pi_1,\pi_2)$ is the composition,
\begin{align}
    Q(\mathbf{r}'',x'',t_{f_2}|\mathbf{r},x_i,t,\rho_{1,2}) := \sum_{\mathclap{\mathbf{r}',x',t_{f_1}}}G_s(\mathbf{r}'',x'',t_{f_2}|\mathbf{r}',x',t_{f_1},\pi_{2})G_s(\mathbf{r}',x',t_{f_1}|\mathbf{r},x_i,t,\pi_{1}),\label{eq:plan-jump-op}
\end{align}
where the definition extends for a $\rho$ of any length by sequentially multiplying $G_s$ and marginalizing over all of the intermediate state-times.
Thus, $Q_m:(\mc S \times \mc T \times \mc P_m)\times (\mc S \times \mc T)\rightarrow[0,1]$ can be computed by BFS where leaf nodes index $m$-length plans $\rho\in\mc P_m$ and the resulting state-times. We can see what an individual plan $\rho_{1,3,2}=(\pi_{g_1},\pi_{g_3},\pi_{g_2})$ looks like when executed in a simple gird-world with physiological states in figure \ref{fig:TG-CMDP Examples}.

Since the leaves contain the resulting state-time $(\mathbf{s}_\rho, t_\rho)$ of applying the open-loop plan $\rho$, we can evaluate the state-time with the availability function $\bar{f}_{\alpha}(\mathbf{s}_\rho, a^{*}, t_\rho)$. While this can be easily done for a given $\bar{f}_{\dg}$, a good method of defining this function is not obvious. An engineer might intuit that the best internal state that should be encoded as the goal is the fully satiated state $\mathbf{r}_{max}=(w_{max},y_{max},z_{max})$, however, this state \textit{is clearly not reachable}. How should one know, \textit{a priori}, which of the many possible vectors in a large Cartesian product space should or should not be encoded as a ``goal" in $\bar{f}_\dg$? 

\begin{algorithm2e}
\DontPrintSemicolon
\SetKw{return}{return}
\SetKwRepeat{Do}{do}{while}
\SetKwData{conflict}{conflict}
\SetKwData{safe}{safe}
\SetKwData{sat}{sat}
\SetKwData{unsafe}{unsafe}
\SetKwData{unknown}{unknown}
\SetKwData{true}{true}
\SetKwInOut{Input}{input}
\SetKwInOut{Output}{output}
\SetKwFor{Loop}{Loop}{}{}
\SetKw{KwNot}{not}
\begin{small}
	\Input{$\widehat{\eta}_{c}, \mc H_{c} = \{\eta_{e_1,g_1},\eta_{e_1,g_2},...\},\Omega_{r} = \{\omega_w,...,\omega_z\},\mc P_{r}=\{P_w,...,P_z\},P_x, P_\sigma$, $\Pi_{e,g}=\{\pi_{e_1,g_1},\pi_{e_1,g_2},...,\pi_{e_\ell,g_{k}}\}$, $\zeta$, $F$, $m$}
	\Output{tree}
	let $root\leftarrow Node(\boldsymbol{\sigma}_{init},r_{init}, x_{init}, t_0, \rho=(), leaf\leftarrow False, parent \leftarrow none)$ be an initial node\;
	$\mathscr{K}_{sub} \leftarrow \texttt{compute\_all\_sublimated\_feasibility\_functions}(\mc P_r,f)$\;
	$tree \leftarrow initialize\_tree(root)$\;
	$Queue.push(root)$\;
	\While{$not\_empty(Queue)$}{
	    $node \leftarrow Queue.pop()$\;
        $(\boldsymbol{\sigma},\mathbf{r},x,t) \leftarrow (node.\boldsymbol{\sigma},node.\br,node.x,node.t)$\;
        $e\leftarrow \zeta(\mathbf{r},\boldsymbol{\sigma})$\;
	    \For{$\pi_{e,g}$ in $\Pi_e$}{ \If{$(\kappa_{\pi_{e,g}}(x,t)> 0) \land (\texttt{sublimation\_check}((\boldsymbol{\sigma},\mathbf{r},x,t),\mathscr{K}_{sub})=True)$}
	    {
	            $(\boldsymbol{\alpha},x_f,t_f) \leftarrow \widehat{\eta}_c(\boldsymbol{\alpha},x_f,t_f|x,t,\mathbf{r},\pi)$\;
	            $\mathbf{r}' \leftarrow \texttt{update\_all\_internals}(\mathbf{r},t_f-t,\Omega_{r})$\;
	            $x'' \leftarrow P_x(x''|x_f,\pi(x_f,t_f),t_f)$\;
	            $\mathbf{r}'' \leftarrow \texttt{one\_step\_internal\_update}(\mathbf{r}',\boldsymbol{\alpha},\mc P_{r})$\;
	            $\boldsymbol{\sigma}''\leftarrow P_\sigma(\boldsymbol{\sigma}''|\boldsymbol{\sigma},\boldsymbol{\alpha}(\boldsymbol{\sigma}))$\;
	            $t'' \leftarrow t_f+1$\;
  
    	        \uIf{$len(node.\rho)<m$}{
        	       $new\_node \leftarrow Node(\boldsymbol{\sigma}'',\mathbf{r}'',x'',t'', leaf \leftarrow \texttt{False},parent\leftarrow node)$\;
        	       $tree \leftarrow add\_to\_tree(tree,new\_node)$\;
        	       $Queue.push(new\_node)$\;
    	        }
    	        \Else{
    	            $new\_node \leftarrow Node(\boldsymbol{\sigma}'',\mathbf{r}'',x'',t'', leaf \leftarrow \texttt{True},parent\leftarrow node)$\;
    	           $tree \leftarrow add\_to\_tree(tree,new\_node)$\;
        	        
    	        }
	        }
	    }
	    
	}
	$\texttt{Return}(tree)$\;
\end{small}
\caption{SPA$\_$Breadth$\_$First$\_$Search}
\label{alg:environment_update}
\end{algorithm2e}


\section{Self-Preserving Agent}
\subsection{Goal Setting with the Valence Function}\label{sec:self-preserving}
We now turn to the question: what is a good goal? Previously, we introduced OBEs to solve for policies that induce variables, $\alpha_\dg$, that act on a different space, i.e. a \textit{task goal}, but there also are many possible \textit{non-task goal-states} that do not emit a $\alpha_\dg$ variable. Our approach is to evaluate the quality of \textit{any} possible state-vector of the Cartesian product space with an intrinsic measure of the change of the agent's empowerment in order to side-step the problem of defining $\bar{f}_{\dg}$. We define the valence function $\mathfrak{V}_n$ as the n-step empowerment difference from following an $m$-step plan $\rho$ from initial state-time, $(\mathbf{s},t)$, and ending at $(\mathbf{s}_{\rho},t_{\rho})$:
\begin{align}
    \mathfrak{V}_{n}(G,G',Q_m,\mathbf{s},t,\rho) := \expec_{(\mathbf{s}_{\rho},t_{\rho})\sim Q_m(\cdot,\cdot|\mathbf{s},t,\rho)}\Big[\underbrace{\mathfrak{E}_n(G'|\mathbf{s}_{\rho},t_{\rho})}_{\text{Final Emp.}}-\underbrace{\mathfrak{E}_n(G|\mathbf{s},t)}_{\text{Initial Emp.}}\Big],\label{eq:valence}
\end{align}
where $G$ is an operator, $G'$ is an operator evaluated after following $\rho$ that shares the same domain as $G$ but can potentially have different transition dynamics, and $(\mathbf{s}_{\rho},t_{\rho})$ is the state vector and time after executing a plan $\rho$ from $(x,t)$ under the $m$-step plan operator $Q_m$ in equation \eqref{eq:plan-jump-op}.
The open-loop plan $\rho$ can stand for any kind of action sequence, e.g. either $\rho = (a^{(1)},...,a^{(m)})$ or (for the focus of this paper) $\rho = (\pi^{(1)},...,\pi^{(m)})$. 
Note that controlling with sequence of polices is a (deterministic) semi-Markov formulation somewhat similar to the Options framework \cite{sutton1999between}, however we do not use predefined or learned initiation- or termination-state sets, rather, terminations conditions are defined by goal-success or failure events dictated by the STFF.
The operator $G$ can be a fine-grained operator such as $P_s(\mathbf{s}'|\mathbf{s},a,t)$, 
or a hierarchical goal-operator $G_s(\mathbf{r}',x',t'|\mathbf{r},x,t,\pi)$. 
Computing empowerment on an operator is a measure of the agent's capacity on the domain of that operator. While the fine-grained operator for the full product space, $P_s$, is the most complete, it is challenging to compute empowerment at long horizons. Alternatively, the value of state-time goal-operators like $G_s$ is that one can compute empowerment \textit{at the level of the task}, which means that it operates over long state-time ranges in the sub-space of low-level states the task is defined on, called the \textit{grounded subspace} \cite{ringstrom2020jump}, and the resulting internal states. 
If we define the set of plans of length $m$ obtained from BFS as $\mc P$, then we can compute the best plan which maximizes valence from state-time $(\mathbf{s},t)$,
\begin{align}
    \nu^*_{\rho,\mathbf{s},t} = \max_{\rho \in \mc P_m}\big[\mathfrak{V}_n(G,G',Q_m,\mathbf{s},t,\rho)\big],\label{eq:max-plan}\\
    \rho^*_{\mathbf{s},t} = \argmax_{\rho \in \mc P_m}\big[\mathfrak{V}_n(G,G',Q_m,\mathbf{s},t,\rho)\big].\label{eq:argmax-plan}
\end{align}

We write the optimal valence as $\nu^*_{\rho,\mathbf{s},t}$, leaving the transition operators out as implicit arguments. Also, note that $\nu^*_{\rho,\mathbf{s},t}=\nu^*(\mathbf{s},t,\rho)$, where the appended subscripts indicate that it is a scalar for a \textit{valence-to-go function} $\nu$ computed only for one initial state-time. A valence-to-go function is simply valence in the context of a \textit{valence Bellman equation}, which we will discuss later on in Section \ref{sec:VBE}.

Computing empowerment in a product space can be done in principle with the Blahut-Arimato algorithm, however the row-space of the channel matrix will be $|\Pi|^n$ for a horizon $n$, and the column space will be $|\mc S \times \mc T|$, which scales exponentially with additional new state-spaces comprising $\mc S$.  However, with determinism in $P_s$ $F$ and $\bar{f}_{\dg}$ of the TG-CMDP, one can represent an product-space state vector $\textbf{s}$ as a set of (deterministic) marginal state vectors $(\textbf{x},\textbf{w},\textbf{y},...,\textbf{z})$ to avoid representing the full product space, and forward propagate these vectors through $\mc O$ for all actions to sum up the total number of reachable states after $n$ steps (see section \ref{sec:det-emp-product} for pseudocode).  We do not currently have a method for computing empowerment with a stochastic factorization without building the full channel, but it could be possible in future implementations to maximize a lower-bound on empowerment by fixing an action-sequence distribution (e.g. uniform) $p(\textbf{a}^{(n)}_t)$ constant and maximize the mutual information between action-sequences and states under this distribution. Therefore, stochastic operators and availability functions can be used for SPA, but determinism plays a major role in simplifying the computations.

\subsection{Self-Preserving Agent Definition}
We have defined all of the components for a self-preserving agent, which includes the base transition operators, abstract goal operators, and valence functions. We now formally define SPA as: 
\begin{mydef}{Self-Preserving Agent}
~ A self-preserving agent (SPA) is a 4-tuple $\mathbb{S}=(P_x,\mc P_r, \zeta, F)$, where $P_x:(\mc X \times \mc A \times \mc T \times \mc E) \times \mc X \rightarrow [0,1]$ is a low-level transition operator with a defective mode $e_-$ and normal mode $e_+$, $\mc P_r = \{P_w,P_y,...,P_z\}$ are secondary transition operators, $\zeta$ is a mode function which maps defective states-vectors to the defective mode, and $F$ is an action-availability function.
\end{mydef}

SPA uses these components to compute an ensemble of unconstrained state-time feasibility functions and their policies using the OBE equations (\ref{obs-C-Bellman-kappa}-\ref{graveyard-marg-eta}) to construct the unconstrained aggregate feasibility functions $\mc H_u$ from which it can derive the constrained aggregate $\widehat{\eta}_{c}$, the internal prediction operator $\omega_{r}$. Then, SPA runs BFS for $m$-length plans, forward sampling state-vectors with the factorization in equation \eqref{eq:jump-op}, implicitly forming $Q$ (equation \eqref{eq:plan-jump-op}) by enumerating the leaves. After BFS, SPA computes empowerment over all the leaf state-times and the initial state-time in order to calculate the valence of each plan with equation \eqref{eq:valence}, and selects the best plan $\rho^*$ using equation \eqref{eq:argmax-plan}, and follows the plan it until completion (see Algorithm \ref{alg:SPA} for pseudocode). 

We will now provide two theoretical examples of the valence optimization, one simple one in which the environment does not change \ref{sec:example_1}, and then in the context of affordances (\ref{sec:example_2}) where we compute it in task space when the environment can change.

\begin{algorithm2e}[h]
\DontPrintSemicolon
\SetKw{return}{return}
\SetKwRepeat{Do}{do}{while}
\SetKwData{conflict}{conflict}
\SetKwData{safe}{safe}
\SetKwData{sat}{sat}
\SetKwData{unsafe}{unsafe}
\SetKwData{unknown}{unknown}
\SetKwData{true}{true}
\SetKwInOut{Input}{input}
\SetKwInOut{Output}{output}
\SetKwFor{Loop}{Loop}{}{}
\SetKw{KwNot}{not}
\begin{small}
	\Input{$\mc P_e = \{P_{x,e_0},P_{x,e_1},...\},\mc P_r = \{P_w,P_y,...,P_z\}, \mc D,\mc E, \zeta, F,n,m,T_f$}
	$\mc F \leftarrow \{f_{\dg_1},f_{\dg_2},...\}$, set of goal-restricted availability functions of $F$\;
	$\mc T_{istc} \leftarrow \{\}$,  internal state-space time constraint set (ISTC)\;
	\For{$e \in \mc E$}{
    	\For{$f_{g_i}\in \mc F$}{
        	$(\kappa_{e,g_{i}},\pi_{e,g_{i}},\eta_{e,g_{i}}) \leftarrow \texttt{feasibility\_iteration}(P_{x,e},f_{g_i},T_f)$\;
        	$(\mathscr{K},\Pi,\mc H)\leftarrow \texttt{append\_to\_sets}((\mathscr{K},\Pi,\mc H),(\kappa_{e,g_{i}},\pi_{e,g_{i}},\eta_{e,g_{i}}))$\;
    	}
	}
	\For{$P_{\ell}\in \mc P_r$}{
	    define function $\omega_\ell(\ell_j|\ell_i,t_d) := P_{\ell,\epsilon}^{t_d}(i,j)$\;
	    $\Omega.\texttt{append}(\omega_\ell)$\;
	    $\bar{P}_{\ell,\epsilon} \leftarrow \texttt{remove\_states}(P_{\ell,\epsilon},\mc D)$\;
	    $\mathbf{t}_\ell\leftarrow\texttt{lin\_sys}(I-\bar{P}_{\ell,\epsilon},\mathbf{1})$\;
	    $\mc T_{istc}.\texttt{append}(\mathbf{t}_\ell)$\;
	}
	$\widehat{\eta}_c \leftarrow \texttt{time\_constrain}(\mc H, T_{istc})$\;
	$\mc O \leftarrow \{\widehat{\eta}_c,\Omega,\mc \mc P_r,P_x\}$\;
	$tree\leftarrow\texttt{SPA\_breadth\_first\_search}(\mc O,\Pi,F,\zeta,m)$\;
	$init\_emp \leftarrow \texttt{empowerment}(\mc O,root.s,root.t,n)$\;
	$best\_valence = 0$\;
	\For{$leaf \in tree.leaves$}{
	    $final\_emp \leftarrow \texttt{empowerment}(\mc O_{leaf.\rho},leaf.s,leaf.t,n)$\;
	    $cur\_valence \leftarrow final\_emp - init\_emp$\;
	    \If{$cur\_valence > best\_valence$}{
	        $best\_valence \leftarrow cur\_valence$\;
	        $best\_\rho \leftarrow leaf.\rho$\;
	    }
	}
	$\texttt{execute\_plan}(best\_\rho)$\;
\end{small}
\caption{Self-Preserving Agent (Deterministic)}
\label{alg:SPA}
\end{algorithm2e}

\subsubsection{Example 1: Valence from change in state}\label{sec:example_1}
\begin{figure}[h]
\centering
\includegraphics[width=\linewidth]{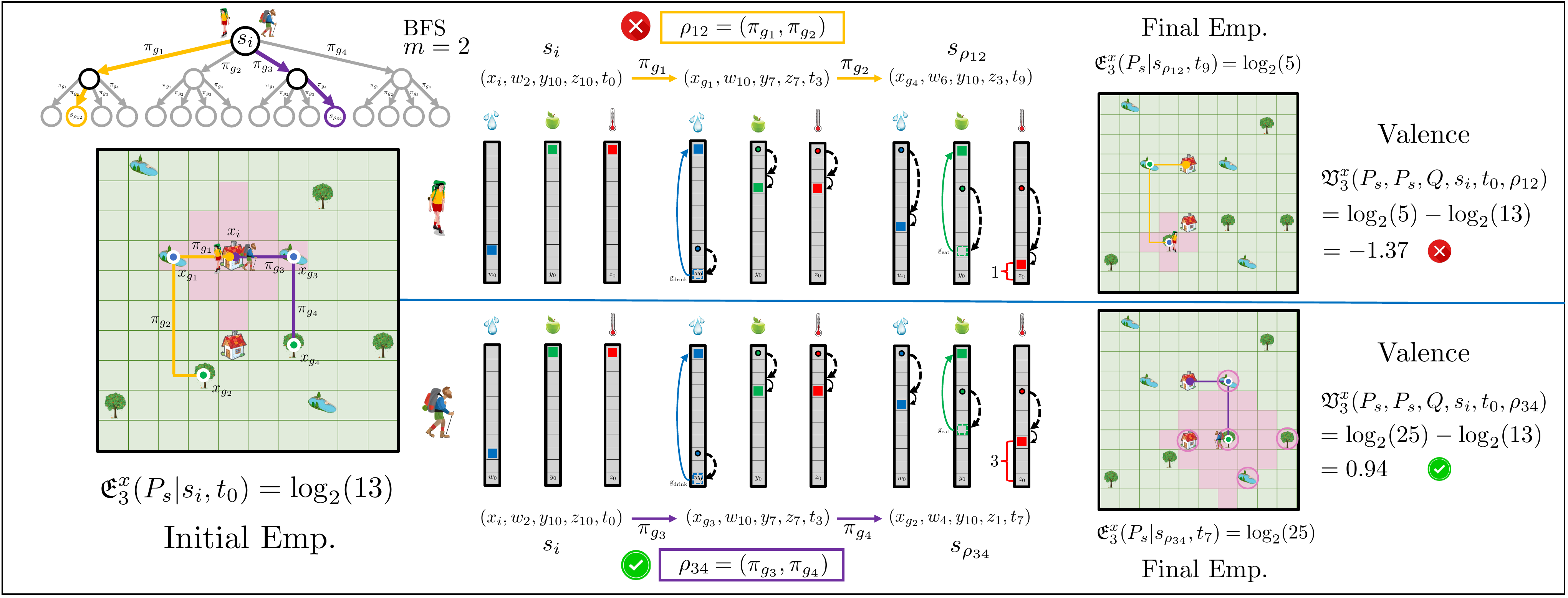}
\caption{Valence evaluation for a change in state:  Two hikers, represent two possible plans generated by SPA. We show the evolution of the state vectors under the plans and compute the valence.  Hiker 1, follows a plan which leads to a decrease in empowerment for a negative valence, and Hiker 2 follows a plan which increases its empowerment for a positive valence.}
\label{fig:Valcence_comb}
\end{figure}

Now that we have developed the mechanics of SPA we can being to illustrate some theoretical examples.
Figure \ref{fig:Valcence_comb} shows the BFS plan prediction and valence computation of SPA for two plans, using two hikers for the purpose of illustration, where each start at the same low-level state $\mathbf{s}_i$. Empowerment on product spaces is challenging to visualize, so for illustrative purposes we use marginal empowerment, $\mathfrak{E}_n^x$ (see A.\ref{appx.marg_emp}), which shows the contribution from the space $\mc X$ given a physiological state which bounds the hiker's movement. Initially, both hikers cannot move more than 2 states without dying, thus both Hikers have an initial empowerment (setting $n=3$ for simplicity) $\mathfrak{E}_3^{x}(P_s|\mathbf{s}_{i},t_0) = \log_2(13)$. Hiker 1 (Yellow Shirt) executes $\rho_{12}=(\pi_1,\pi_2)$ ending at state $x_{g_2}$, with an empowerment of $\log_2(5)$ which produces a negative valence (-1.37).  Notice that none of the reachable states (shaded pink) include the house, so Hiker 1 will inevitably die because he cannot return home to get warm. Notice that the temperature state effectively dictates the final empowerment for $P_{s}$ for both plans, as the $z$ internal state which will dictate the switch to the defective mode the soonest (in $1$ step for Hiker 1 and $3$ steps for Hiker 2). The horizon parameter $n$ matters here, because for $n=1$ the final empowerment on $P_s$ will be the same for $\rho_{12}$ and $\rho_{34}$ but as $n$ is increased greater to $2$ or $3$ the discrepancies between $\rho_{12}$ and $\rho_{34}$ become apparent. Hiker 2 (blue shirt) executes $\rho_{34}=(\pi_3,\pi_4)$ ending at state $x_{g_4}$, with an empowerment of $\log_2(25)$ and a positive valence (0.94), thus, SPA will select $\rho_{34}$ as the better plan. Agent 2 is also also capable of performing four other possible tasks within the empowerment horizon, which includes the cabin unlike Hiker 1. These tasks, as we will discuss shortly, can be formalized as Gibsonian affordances \cite{gibson1977theory}, which we will later show can be directly changed to increase valence.

\subsection{Affordances}\label{sec:afford}

In the previous example we suggested that the tasks available to the agent are Gibsonian affordances, which are environmental opportunities to achieve an internal goal \cite{gibson1977theory, roli2022organisms}.  Affordances, in Gibson's original conception, is what an environment \textit{offers} \textit{to an agent}. Affordances are not just controllable aspects of the environment, they must incorporate an agent's capacities in relation to the environment; feasibility functions naturally represent such capacities.

Affordances have been formalized in RL \cite{abel2014toward, abel2015goal, khetarpal2020can} where Khetarpal et al. define them as $(x,a)$-pairs satisfying an set of intended distributions over future \textit{states}, where the intention of an action is a distribution over next-states given an initial state. However, we can also think about affordances as being the intended high-level \textit{goal-actions} which induce state-transformations on a space. This is a useful notion because it is the state-transformation itself that we care about learning (which we discuss in section \ref{sec:lifelong}), and also because there could be many distinct (high-level) states we might want to apply a transformation to, and so we might only be interested in the means by which we induce the transformation to those states. With SPA goals are afforded contingent on the agent's internal states and environment, which restricts the agents movement to realize goals. If the empowerment horizon is set to $n=1$ and the dynamics are deterministic, then the agent is computing the logarithm of the number of ways in which it can feasibility influence other systems of interest under $G_s$, which is a sensible way of understanding an affordance. For our purposes, a similar definition to Khetarpal et al. can be given as, $(\mathbf{s},\pi)$\textit{-pairs that satisfy an intended higher-order action $\alpha_g$}: 
\begin{align*}
    \mc{AF}_{\alpha_g}^t = \{(\mathbf{s},\pi): \sum_{\mathbf{s}_f,t_f}\widehat{\eta}(\alpha_g,\mathbf{s}_f,t_f|\mathbf{s},t,\pi)>0, \forall(\mathbf{s},\pi)\in \mc S \times \Pi\}.
\end{align*}
An affordance set $\mc{AF}_{\alpha_g}^t$ is notation introduced to establish a connection between our work and the work of Khetarpal et al. and Roli et al. \cite{khetarpal2020can,roli2022organisms}, 
and it simply reflects the structure of $G_s$ since both are defined by $\widehat{\eta}$. Thus, changes in $\mc{AF}_{\alpha_\dg}^t$ indicate changes to the structure of $G_s$. 

Gibson imagined the activity of changing the environment to create affordances as a primary activity of organisms. In his words:

\begin{quote}
    ``Why has man changed the shapes and substances of his environment? To change what it affords him. He has made more available what benefits him and less pressing what injures him. [...] Over the millennia, he has made it easier for himself to get food, easier to keep warm, easier to see at night, easier to get about, and easier to train his offspring." p.122 \cite{gibson1977theory}
\end{quote}

We can see that implicit in Gibson's original conception is the idea that affordances should be actively changed to suit the \textit{needs} of an agent. Suiting one's needs implies a normative evaluation, and we argue in this paper that valence is a good criterion for life-long agents because we can estimate the impact of an affordance on an agent's hierarchical control architecture. In the next section we will show how SPA can compute valence when it alters its own affordances, and we can begin to see how the (colloquial) \textit{value} of an item in the world is not a fixed and static property, but can emerge from the effect the item has an agent's ontology.

\subsubsection{Example 2: Valence from change in environmental affordances}\label{sec:example_2}

\begin{figure}[h]
\centering
\includegraphics[width=\linewidth]{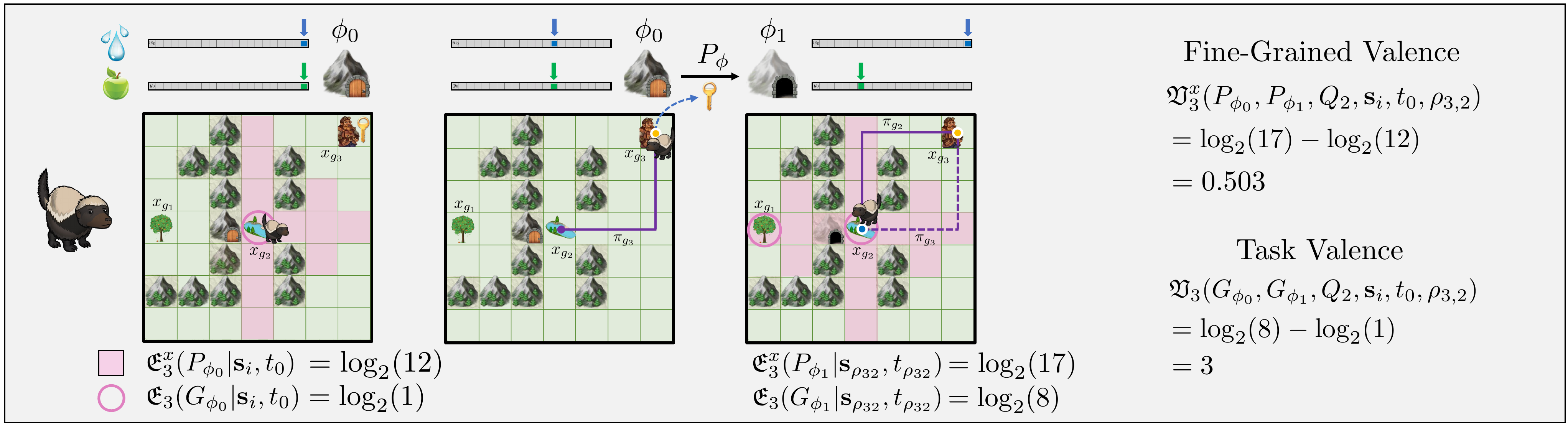}
\caption{Valence evaluation for a change in state and environment: Stoffel starts from a lake with an initial empowerment, and then travels to the Dwarf to obtain a key.  This key allows Stoffel to travel through the mountain-pass by reparameterizing the base transition model to $P_{x,key}$ and the resulting state-time feasibility functions derived from the base model. Thus, if the he returns to the lake then there is an expansion of empowerment both in the fine-grained transition operator $P_s$, and in the goal-operator $G_s$ which allows him to compute empowerment in task-space. With the key, he can now travel back and forth between the lake and tree in order to survive, which is reflected in the valence computation.}
\label{fig:Valcence_comb_2}
\end{figure}

The valence function definition allows for the possibility that there can be two operators for evaluation. Figure \ref{fig:Valcence_comb_2} shows a scenario how this can occur if the agent actively changes the structure of the environment. Here, Stoffel needs food and water to survive, however, he only has access to water on one side of an impassable mountain range. However, if Stoffel obtains a key from the Dwarf, represented by a binary state $\phi$, he can unlock the mountain pass door. This switches the dynamics from $P_{x,e_{\phi_0}}$ to $P_{x,e_{\phi_1}}$ with an additional operator $P_{\phi}(\phi'|\phi,\alpha_{\phi})$, creating a shortcut through the mountain range to an apple tree.  With the door open, he can cycle between water and food fast enough to avoid death. 

The two low-level operators $P_{x,e_{\phi_0}}$ to $P_{x,e_{\phi_1}}$ parameterize different $\widehat{\eta}_{e_{\phi}}$ functions when computing STFF ensembles, and therefore, changes in the environment can induce changes in affordances reflected in $G_{s,e_\phi}$. We evaluate empowerment on fine-grained operators $P_{s,e_{\phi_0}}$ and $P_{s,e_{\phi_1}}$ or goal-operators $G_{s,e_{\phi_0}}$ and $G_{s,e_{\phi_1}}$, each indexed by the environmental modes\footnote{It's important to recognize that we could always include the state-space $\mc E$ in the operator $P_s$ and use a single operator evaluated at two states, however this increases the size product space operator.}. The figure shows an executed plan $\rho_{32}=(\pi_3, \pi_2)$ ($m=2$), followed by an evaluation of empowerment for $n=3$. For this example we assume that the agent computes an ensemble $\mc H_u$ which includes the key as a goal for the plan operator $Q$, but the operator $G_{s}$ that we compute empowerment on is only defined on the physiological goals (see A. for full details). Notice that if Stoffel does not retrieve the key, he will eventually die of starvation. However, if he obtains the key, he can open the mountain pass door and eat from the apple tree.  After following the plan $\rho_{32}$, he returns to the lake to drink water, however, now that the mountain pass is open he can reach extra squares on the other side, leading to a positive fine-grained valence.  Additionally, another task-goal $\alpha_{eat}$ is \textit{afforded}, $\mc{AF}^{e_{\phi_0}}_{\alpha_{eat}} \rightarrow \mathcal{AF}^{e_{\phi_1}}_{\alpha_{eat}}$, under the re-parameterized goal-operator $G_{s,e_{\phi_1}}$. With the key, Stoffel can now follow many other possible plans, such as $\rho = (\pi_1,\pi_2,\stackrel{n-2}{...},\pi_1,\pi_2)$ which repeats $(\pi_1,\pi_2)$ for $n$ iterations, in which he can avoid hitting a defective state; this is captured by n-step final-empowerment. What started out as zero task-empowerment increases to an empowerment of $3$ (with $n=3$ there are $2^3=8$ reachable state-times under $G_{s,e_{\phi_1}}$). Thus, the agent can use feasibility functions (defining $G_s$) to reason across the product space in order to forecast the long-term term consequences arising from changes to the environment structure and affordances.
\subsubsection{Quantifying the value of an item with valence}\label{sec:item value}
Interestingly, we can define a computation in which the agent can attribute value to the key in figure \ref{fig:Valcence_comb_2} with a function $V_n$. The agent can compare its current state $\mathbf{s}_{\phi_1} = (x_i,y_j,...,\phi_1)$ (with the key) to the state it would be in if it did not have the key $\mathbf{s}_{\phi_0} = (x_i,y_j,...,\phi_0)$ with all other states in the vector being equivalent to $\mathbf{s}_{\phi_1}$, where $G_s^{\phi}$ is the operator conditioned on environment state $e_{\phi}$: 
\begin{align*}
V_n(\phi_1|\phi_0,G_s,\textbf{s},t) = \underbrace{\mathfrak{E}_n(G_s^{\phi_1}|\mathbf{s}_{\phi_1},t)}_{\text{With Key}}-\underbrace{\mathfrak{E}_n(G_s^{\phi_0}|\mathbf{s}_{\phi_0},t)}_{\text{Without Key}}.
\end{align*}
The value of the key in figure \ref{fig:Valcence_comb_2} with respect to $(\textbf{s}_i,t_0,n=3)$ is $V_3(\phi_1|\phi_0,G_s,\textbf{s}_i,t_0)=3$. This is the same amount as the valence of the plan in the example, though it does not follow that the value of the item must equal the valence of the plan that obtains it. Also, note that this computation assumes that $G_s$ as an operator does not include the task of obtaining the key in its domain, which would defeat the purpose of the calculation---we want to know the value of the key conditioned on it existing or not existing in the world.

Notice that the value of the key is relative to the agent's state, time, operators, and the horizon $n$. Therefore, a key could become valued or valueless to the extent that these parameters change.  What we have presented is a mechanism in which value can be directly bestowed upon an item by an agent. Notice that this is very similar to a utility function, however, the agent is not committed to static values for items like keys, rather, if an open-ended agent were to learn new structure of the world, it valuations can change as the agent's planning architecture $G$ or state-vector $\mathbf{s}$ change. That being said, an agent could store and recall the value of an item it has calculated in the past as if it were like a utility. This would be an important mechanism for maximizing valence without explicitly computing it, much in the same way that a person might act on an instinct to pick up a twenty dollar bill on the ground without computing the empowerment-gain of increasing the state of his or her bank account by twenty dollars.





\subsection{Empirical Comparison}
\begin{figure}[h]
    \centering
    \includegraphics[width = \columnwidth]{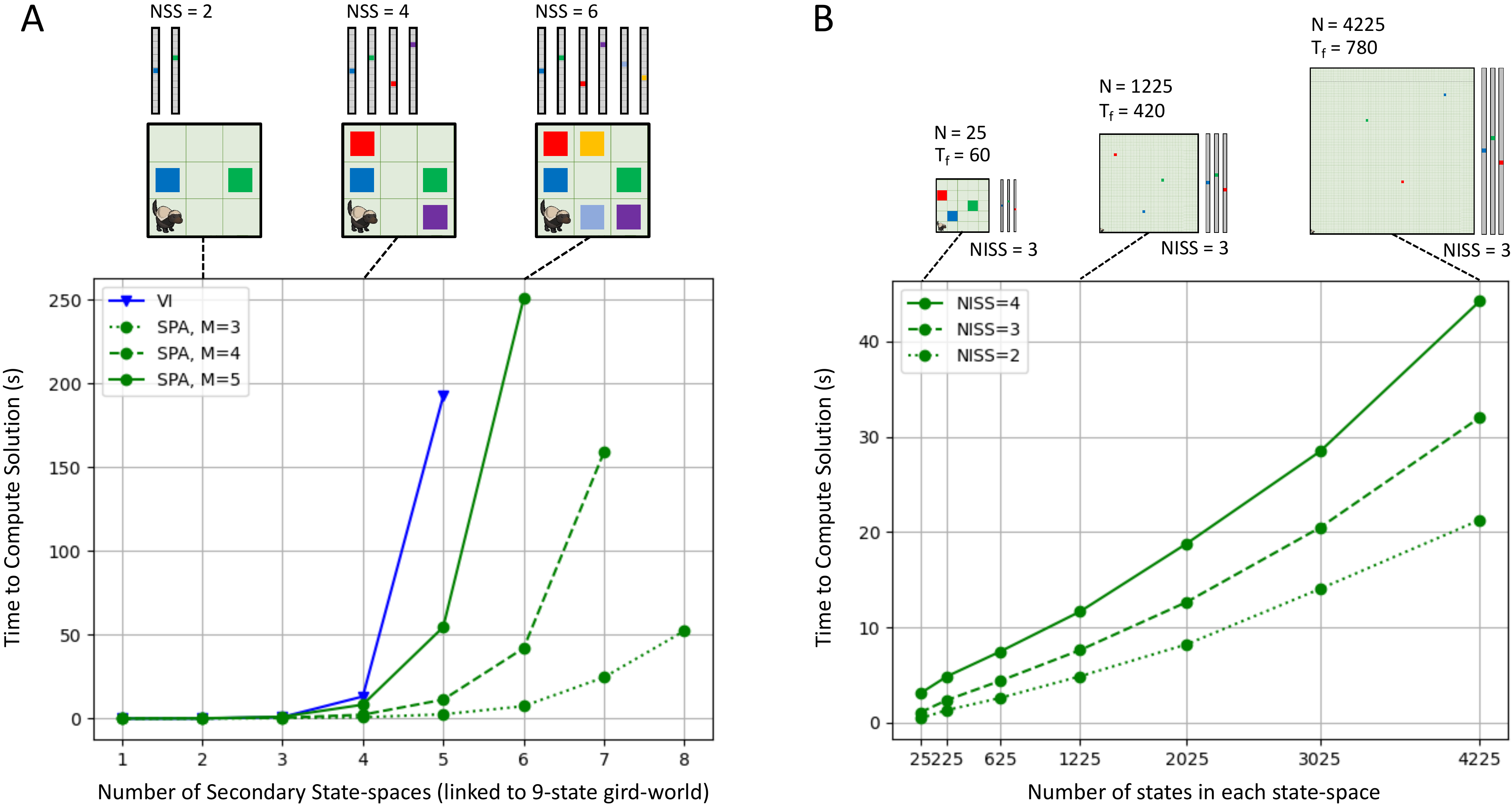}
    \caption{(A) The time to compute the solution is plotted while varying the number of secondary state-spaces linked to a $3\times 3$ grid-world. The number of states in each secondary state-space is fixed at $13$. The blue line is the solution times for IHDR, which exponentiates when we use value-iteration (VI) in the full product space as we add more state-spaces. The green lines are SPAs solution times as we add more state-spaces. The empowerment horizon was fixed to $n=3$, and each green line corresponds to a different BFS depth, $M$. (B) We demonstrate the time-complexity of SPA when increasing the number of states in each state-space far beyond what IHDR can manage, holding the number of internal (physiological) secondary state-spaces (NISS) constant for each line plot and fixing the BFS depth, $M=3$. The number of states, $N$, applies to both the grid-world \textit{and} each of the secondary state-spaces. The problems were solved on an Apple M1 Max, 32 GB RAM.}
    \label{fig:experiments}
\end{figure}

We present, in Figure \ref{fig:experiments}, empirical demonstrations of SPA's time-complexity, and how it compares to IHDR value-iteration solutions for problems in the full product-space. By progressively adding new internal state-spaces, such hunger, hydration, temperature, etc. the problem becomes combinatorially larger, so we show how various parameters of the SPA algorithm change the algorithm's scalability. The IHDR problem assumes that physiological death-states are penalized with an arbitrarily high penalty ($-1000$), and fully satiated states are rewarded with a reward ($+10$), and both IHDR and SPA problems will assume determinism in the transition operators. 

In Figure \ref{fig:experiments}.A we fix the size of the nine-state gird-world ($3\times 3$) and vary the number of secondary state-spaces (e.g. physiological state-spaces) which are linked to randomized goal states. The number of states in each secondary state-space is fixed to $13$. The blue line shows how the solution time exponentiates when we solve the solution with value-iteration in the full product space as we add on more state-spaces. The green lines plot how SPAs solution time scales as we add additional state-spaces to the product-space, where the empowerment horizon was fixed to $3$. We collected results for 10 runs, but the variance was negligible so we simply plotted the mean. Each green line varies the depth of BFS. The exponential growth for SPA is caused by adding more state-spaces and corresponding goal-conditioned policies (goals being colors). BFS with fixed depth $M$ grows polynomially in the number of policies $P$ at the rate $P^M$, where $P$ is also equal to the number of secondary state-spaces. 

The $3\times 3$ gridworld was chosen in \ref{fig:experiments}.A because larger grid-worlds become intractable too quickly for IHDR. To demonstrate how SPA scales beyond the $3\times 3$ gird-world, in Figure \ref{fig:experiments}.B we demonstrate the time-complexity when increasing the size of \textit{all} state-spaces, holding the number of secondary state-spaces constant for each line plot. Note that the number of states, $N$, applies to both the grid-world \textit{and} each of the secondary state-spaces. Each point on the graph quantifies the total time of all aspects of the SPA algorithm, which includes computing all STFFs $\eta$, prediction operators $\omega$, $M$-depth BFS, and empowerment for each leaf of the BFS tree. The empowerment horizon was set to $n=2$, BFS depth was set to $M=3$, and the time horizon $T_f$ for each problem is set to be $12\sqrt{N}$ to guarantee the agent will not exceed the time limit under BFS and empowerment computations. Therefore, the size of the full product-space of state-vectors and times that the factorization of $G_s$ is defined on is $12\sqrt{N}N^S$ where $S$ is the total number of state-spaces (gird-world and physiological). Both IHDR value iteration and SPA were optimized by representing the hierarchical state-space transition operators into sparse matrices for all relevant aspects of the problem when it is computationally advantageous, whereas dense multi-dimensional arrays (tensors) can be used for simplicity, but do not scale well. For smaller problems solved in figure \ref{fig:experiments}.A, sparse matrices for SPA are slower compared to dense tensors, but sparse matrices are necessary for the larger grid-worlds in \ref{fig:experiments}.B due to memory constraints. 

With SPA we can compute solutions to much larger problems because we 1) never need to form the explicit hierarchical operator like IHDR and instead compute a space efficient factorization of $G_s$, and 2) we avoid dynamic programming to compute a value function defined on the full product-space and instead forward-compute a sequence of policies with tree-search. Note that tree-search can become arbitrarily expensive depending on the depth of the roll-out, and thus this example is meant to highlight how valence maximization tree-search can solve problems in a given regime that are simply intractable from the perspective of IHDR. These examples are qualitative comparisons because one objective function is not meant to approximate the solution to another, rather the examples are comparing how different formalizations scale. The central qualitative departure between the two methods is that we never need to justify the arbitrary reward function provided for the IHDR problem, and instead the problem only requires optimizations that depend the hierarchical transition operator. Additionally, SPA was optimized serially on a single core, however, the policies, STFFs, prediction operators $\omega$, and empowerment computations could all be parallelized on multiple cores.

\section{Valence Bellman Equation}\label{sec:VBE}
We will now address the Bellman foundation of SPA. SPA optimizes the valence of an open-loop plan via forward sampling, but we developed the theory from the ground up by first constructing the components of the sampling factorization (built on TG-MDP formalisms) and then we used it to define the valence optimization. However, it is natural to ask the more fundamental question of whether or not there is a formalization for the valence optimization written as a Bellman equation. Indeed, there is; since valence is a scalar, a one-action valence function could be used in a Bellman equation in place of a reward function, constituting a finite-horizon \textit{Valence Bellman Equation} (VBE), here written in a hierarchical form (hVBE) with the hierarchical transition operator $P_s$,
\begin{align}
    &\nu_n^*(\mathbf{s},t)=\max_{a}\left[\mathfrak{V}_n(P_s,P_s,P_s,\mathbf{s},t,a)+ \sum_{\mathbf{s}'}P_s(\mathbf{s}'|\mathbf{s},t,a)\nu_n^*(\mathbf{s}',t+1)\right],\label{eq:VBE1}\\
    &\pi^*(\mathbf{s},t)=\argmax_{a}\left[\mathfrak{V}_n(P_s,P_s,P_s,\mathbf{s},t,a)+ \sum_{\mathbf{s}'}P_s(\mathbf{s}'|\mathbf{s},t,a)\nu_n^*(\mathbf{s}',t+1)\right],\\
    &Q_{\pi}^*(\mathbf{s}_{T_f},T_f|\mathbf{s},t)= \sum_{\mathbf{s}'}Q_{\pi}^*(\mathbf{s}_{T_f},T_f|\mathbf{s}',t+1)P_s(\mathbf{s}'|\mathbf{s},t,\pi^*(\mathbf{s},t)),
\end{align}
Notice that in the VBE, the operator that moves the agent around is the same operator being measured with valence---in this equation there is only the transition operator and an intrinsic measure on the operator, no external quantity. The optimal \textit{valence-to-go function} $\nu$, solved for with value iteration, has the important property that it represents the cumulative empowerment gain of the agent from its initial state to the distribution over states at the final time, $Q_{\pi}^*(\mathbf{s}_{T_f},T_f|\mathbf{s},t)$, under a policy $\pi^*$ (the boundary condition at $t=T_f$ is $Q_{\pi}^*(\mathbf{s}_{T_f},T_f|\mathbf{s}_{T_f},T_f)=1$ and $\nu_n^*(\mathbf{s},T_f)=0$). That is to say, the cumulative empowerment gain of the valence-to-go function can itself be expressed as an abstract valence function $\mathfrak{V}_{n,\pi}$.
This is true because if we unroll the VBE recursion (equation \eqref{eq:VBE1}) into its series form and substitute in the empowerment difference definition of valence for each state (equation \eqref{eq:valence}), then all of the intermediate empowerment values (where $\mathfrak{E}_{\mathbf{s}_{t}} = \mathfrak{E}_n(P_s|\mathbf{s}_{t},t)$) cancel out, leaving only the final empowerment evaluation minus the initial empowerment evaluation, meaning the solution is \textit{path-independent} (see Section \ref{appx:valence-BE} for full derivation):
\begin{align}
    &\nu_{\pi}^*(\mathbf{s}_{t_0},t_0)= \left(\cancel{\expec_{\mathbf{s}_{t_1}}\left[ \mathfrak{E}_{\mathbf{s}_{t_1}}\right]}-\mathfrak{E}_{\mathbf{s}_{t_0}}\right) + \left( \cancel{\expec_{\mathbf{s}_{t_2}}\expec_{\mathbf{s}_{t_1}} \left[\mathfrak{E}_{\mathbf{s}_{t_2}}\right]}-\cancel{\expec_{\mathbf{s}_{t_1}}\left[\mathfrak{E}_{\mathbf{s}_{t_1}}\right]} \right)\\
    &\nonumber\quad+ \left( \cancel{\expec_{\mathbf{s}_{t_3}}\expec_{\mathbf{s}_{t_2}}\expec_{\mathbf{s}_{t_1}} \left[\mathfrak{E}_{\mathbf{s}_{t_3}}\right]}-\cancel{\expec_{\mathbf{s}_{t_2}}\expec_{\mathbf{s}_{t_1}}\left[\mathfrak{E}_{\mathbf{s}_{t_2}}\right]}\right) +...+ \left( \expec_{\mathbf{s}_{T_f}}...\expec_{\mathbf{s}_{t_1}} \left[\mathfrak{E}_{\mathbf{s}_{T_f}}\right]-\cancel{\expec_{\mathbf{s}_{T_f-1}}...\expec_{\mathbf{s}_{t_1}}\left[\mathfrak{E}_{\mathbf{s}_{T_f-1}}\right]}\right),\label{eq:init-to-final}\\
    &\quad\quad= \expec_{\mathbf{s}_{T_f}}...\expec_{\mathbf{s}_{t_1}} \left[\mathfrak{E}_{\mathbf{s}_{T_f}}\right] -\mathfrak{E}_{\mathbf{s}_{t_0}}=  \expec_{(\mathbf{s}_{T_f},T_f)\sim Q_{\pi}^*}\left[\mathfrak{E}_n(P_s|\mathbf{s}_{T_f},T_f)\right] -\mathfrak{E}_n(P_s|\mathbf{s}_{t_0},t_0)
    \\
    &\nu_{\pi}^*(\mathbf{s}_{t_0},t_0)  = \mathfrak{V}_{n,\pi}(P_s,P_s,Q_{\pi}^*,\mathbf{s}_{t_0},t_0,\pi^*).\label{eq:init-to-final-2}
\end{align}
Notice that line \eqref{eq:init-to-final} is the definition of valence, but now for the abstract operator $Q_{\pi}^*:(\mc S \times \mc T)\times (\mc S \times \{T_f\})\rightarrow [0,1]$ for the policy $\pi$, which summarizes the final sequence of expectations $\expec_{Q_{\pi}^*}=\expec_{\mathbf{s}_{T_f}}...\expec_{\mathbf{s}_{t_1}}$, where each $\expec_{\mathbf{s}_{t}}:=\expec_{\mathbf{s}_t\sim P_s(\cdot|\mathbf{s}_{t-1},a,t-1)}$. Note that $\{T_f\}$ is a singleton dimension included for clarity. An agent acting under the optimal policy of this equation will seek out the most empowered states reachable under the horizon $T_f$, in expectation, from the initial state-time. 

As we can see, VBEs have a self-similar structure: if reward is replaced with valence for low-level state-actions, then the valence-to-go function $\nu^*_{n,\pi}$ is interpreted as a valence function at the level of state-policies, $(\textbf{s},\pi)$, under an operator $Q_\pi^*$. In other words: valence in, valence out. Because of equation \eqref{eq:init-to-final}, we can see that a VBE agent (which includes SPA) does not optimize the path that it takes to get to the final state (distribution) of highest empowerment, it only optimizes for reaching high-empowerment states under the horizon $T_f$, so all policies which have the same termination distribution are valence-equivalent from a given starting state-time. This means that VBE's can be naturally extended for hierarchical abstraction, as we could always create a higher-order semi-Markov hVBE with abstract operators which map from initial-to-final states, where the abstract operators are synthesized via function composition. This property makes it well-suited for incorporating OBE STFF solutions into the definition of a semi-Markov VBE, since OBEs are an optimization based on reachability rather than reward-maximization. 

Solving a VBE with dynamic programming should be avoided, especially in a hierarchical state-space, since it first requires calculating empowerment difference for every product-space state-action vector just to define the one-step valence function, then followed by solving the VBE for a valence-to-go function in the product space, both of which are computationally prohibitive. However, it is useful and illustrative to formalize the VBE because it defines the \textit{class} of model-based optimization SPA contained within and motivates the approach of forward sampling from a single state-time. Full solutions to this class would require representing feasibility function on a huge space, and therefore forward solutions from a single given starting state are ideal for VBEs because we do not need to evaluate empowerment for intermediate states. 


SPA, however, is a semi-Markov formulation that uses open-loop sequences of closed-loop policies, and so the canonical VBE above does not take this into account. Also, we defined SPA so that it has a constant number of polices for the BFS roll-out, which means that there is not a fixed time-horizon $T_f$. However, SPA fits in the VBE framework if we formalize it as a hierarchical semi-Markov VBE (hsmVBE). This equation is parameterized by the goal-operator $G_s$, produced by OBEs, and with the horizon $K_f$ which is the open-loop plan length and number of abstract time-steps $\mc K=\{k_0,k_1,...,K_f\}$. The equations are given as:
\begin{align*}
    &\nu_n^*(\textbf{s},t,k)=\max_{\pi}\left[\mathfrak{V}_n(G_s,G_s,G_s,\textbf{s},t,\pi)+ \sum_{\textbf{s}',t'}\overbrace{G_s(\textbf{s}',t'|\textbf{s},t,\pi)}^{\mathclap{\text{Constructed from OBE solutions}}}\nu_n^*(\textbf{s}',t',k+1)\right],\\
    &\mu^*(\textbf{s},t,k)=\argmax_{\pi}\left[\mathfrak{V}_n(G_s,G_s,G_s,\textbf{s},t,\pi)+ \sum_{\textbf{s}',t'}G_s(\textbf{s}',t'|\textbf{s},t,\pi)\nu_n^*(\textbf{s}',t',k+1)\right],\\
    &Q_{\mu}^*(\textbf{s}_{K_f},t_{K_f},K_f|\textbf{s},t,k)= \sum_{\textbf{s}',t'}Q_{\mu}^*(\textbf{s}_{K_f},t_{K_f},K_f|\textbf{s}',t',k+1)G_s(\textbf{s}',t'|\textbf{s},t,\mu^*(\textbf{s},t,k)),
\end{align*}
where,
\begin{align*}
    \nu_{n,\mu}^*(\mathbf{s},t,k) =  \expec_{(\mathbf{s}_{K_f},t_{K_f})\sim Q_{\mu}^*}\left[\mathfrak{E}_n(G_s|\mathbf{s}_{K_f},t_{K_f})\right] -\mathfrak{E}_n(G_s|\mathbf{s},t) = \mathfrak{V}_{n,\mu}(G_s,G_s,Q_{\mu}^*,\mathbf{s},t,k,\mu^*),
\end{align*}
is the semi-Markov valence-to-go function, $\mu$ is a closed-loop meta-policy, and $Q_{\mu}^*:(\mc S \times \mc T\times \mc K)\times (\mc S \times \mc T \times \{K_f\})\rightarrow[0,1]$ is the final state-time prediction at abstract time (i.e. policy count) $K_f$ (again, $\{K_f\}$ is a singleton dimension for clarity). Also, $G_s$ used in the first two arguments of the valence function can be substituted for $P_s$, depending on the choice over operator the agent uses for the empowerment calculation. The abstract time-steps $k$ are non-standard, but this is a well-formed equation where one can think of the state-time $(\textbf{s},t)$ as acting as a single state and $k$ acts as the (abstract) time variable in a standard finite-horizon MDP problem. Notice that $Q_{\mu}^*$ is simply the closed-loop version of the open-loop forward sampling plan operator $Q_m$ defined by equation \ref{eq:plan-jump-op}. SPA, on the other hand, computes open-loop meta-policies (plans) $\rho \in \mc P_m$ to construct $Q_m$. 

For the exact same reason that intermediate empowerment calculations cancel out for the original VBE, it is easy to see that SPA's valence evaluation of a plan, where it evaluates initial and final empowerment, is equivalent to summing the empowerment difference at every step along the way. Therefore, given that the hsmVBE above has the same horizon, we can relate SPA's open-loop sampling procedure to the closed loop valence-to-go function with the inequality,
$$\nu_{\rho,\textbf{s},t}^* \leq \nu_{\mu,n}^*(\textbf{s},t,k_0),$$
due to the fact that any open-loop policy can easily be cast as a closed-loop policy by assigning each action in a sequence to a time-index that parameterizes a non-stationary policy. Therefore, any open-loop policy exists within the set of all possible closed-loop non-stationary policies and so the optimal closed-loop policy bounds any open-loop policy from above. Equality between these two quantities will be achieved if the dynamics ($P_x, P_y, ...$) and action-availability function $F$ are deterministic because these conditions imply that the closed-loop meta-policy $\mu^*$ will produce a single deterministic sequence of policies when sampled, and this sequence of policies must therefore be an optimal valence-maximizing open-loop plan. 

We developed TG-CMDP as a formalism for solving specific tasks defined by $f_{\dg}$, which produces feasibility functions for abstract planning through the OBE. The VBE, on the other hand, is for controlling to any state which is determined to be in the interest of the agent under empowerment-gain as a metric. To summarize the interplay between the two Bellman equations: The OBEs play a functional role in making forward solutions to an intractable hierarchical VBE possible by creating feasibility function transition operators that connect the ``contact points" (states associated with non-null actions of $F$) of the individual component transition operators of a hierarchical transition operator $P_s$ (equation \eqref{eq:hierarchical_trans_op}) which parameterizes the VBE.  This is done by turning a Hierarchical VBE into a semi-Markov variant, the hsmVBE, and forward sampling plans.





\section{Reusable Modular Representations and Compositionality}\label{sec:compositionality}

Having established the connection between the OBE and VBE, we will now orient our discussion towards the topic of representational reusability and compositionality. The SPA formalism is compositional because the STFFs compose with themselves on the low-level state-space under sequential policy calls.  Composability implies that for policies $\pi_1$ and $\pi_2$ the corresponding STFFs can be represented as matrices $H_{\pi_1}$ and $H_{\pi_2}$ such that $H_{\pi_1}H_{\pi_2}=H_{\rho_{12}}$, which is an STFF matrix for the plan $\rho_{12}=(\pi_1,\pi_2)$. Thus, the matrix multiplications preserve the STFF interpretation, and so STFFs are objects that exhibit closure under matrix multiplication. This is in contrast to other representations like the successor representation (SR) from the infinite horizon discounted reward Bellman equation, which are matrices representing weighted state-occupancy that do not exhibit closure or compose across hierarchies \cite{dayan1993improving, gershman2018successor}. Furthermore, the SPA formalism also supports reusability because we can always expand our set of policy and STFF solutions for new problems and remap those solutions to new higher-level structure (refer back to figure \ref{fig:composition} for illustration).  Taken together, this means we can create flexibly composable multi-level transition systems for solving difficult problems, which we will now discuss.

\subsubsection{Vertical Composition: Goals as Non-stationary Non-Markovian Tasks}

Previously, we have shown how we can plan in a hierarchical state-space by forward sampling.  The hierarchical state-space consisted of a low-level state-space $\mc X$ and physiological state-space $\mc R$.  However, we can include other intermediate state-spaces, such as a binary vector task state-space $\Sigma$ with a transition operator $P_\sigma$, where we can define non-Markovian \textit{Boolean Ordered Goal} (BOG) tasks to drive the dynamics of higher-order state-spaces, such as the physiological state-space \cite{ringstrom2020jump}. When we say non-Markovian, we mean that the problem is non-Markovian with respect to the base state-space $\mc X$, where the higher-levels of abstraction, such as a binary vector space of a BOG task, can record the agent's progress in a task on $\mc X$. It is straightforward to map tasks defined on intermediate level state-spaces to induce dynamics in a higher-level space.

\begin{figure}[t]
\centering
\includegraphics[scale=0.40]{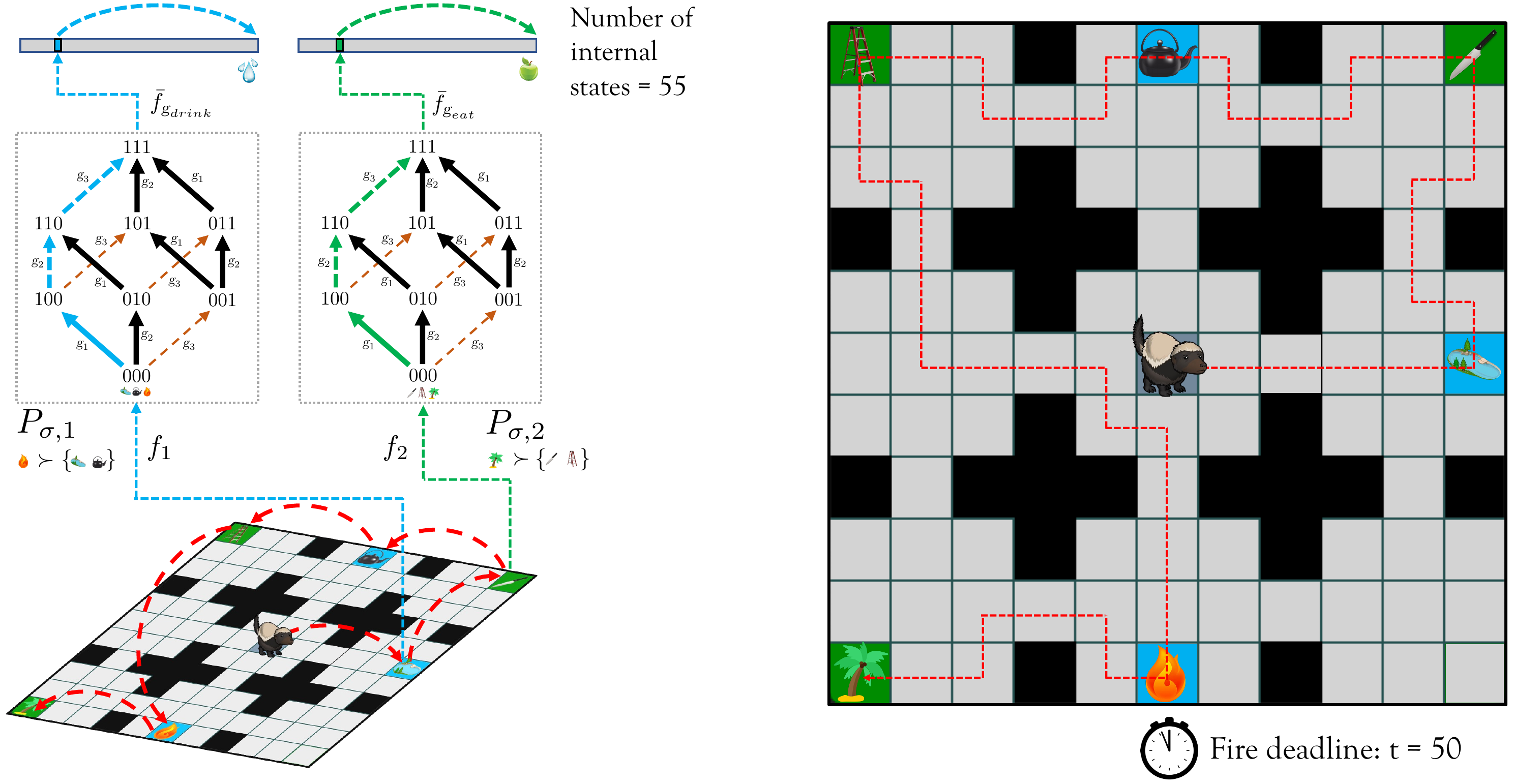}
\caption{Two different BOG tasks $\mathscr{B}_1$ and $\mathscr{B}_2$ can be completed in order to satisfy physiological needs. Each BOG task will only have one accepting state, which is the state $\boldsymbol{\sigma} = 111$. $\mathscr{B}_1$ requires SPA to collect a knife and a ladder in order to retrieve and break open a coconut for food. $\mathscr{B}_2$, requires SPA to collect pond water and a kettle in order to boil the water at a fire for safe drinking. Both tasks induce physiological jumps at the accepting state, and both tasks have precedence constraints encoded into the task space (red dashed arrows indicated disallowed transitions in $P_\sigma$). SPA starts the task full satiated in both physiological state-spaces and forward samples polices using state-time feasibility functions to flip bits (on the left, two dark red dashed arcs denote $\eta$-function state-time transitions to flip the first bits of the binary vector for each task). Each logical task is encoded in the $2^{nd}$-order action-availability function $\bar{F}$ with two task-goals $\bar{\alpha}_{drink}$ and $\bar{\alpha}_{eat}$.  The $1^{st}$-order action-availability function $f_\dg$ encodes all individual goals for each feature on $\mc X$ to induce dynamics on each $\Sigma$ space, and also encoded in $F$ is a time deadline on the fire, causing it to extinguish after $50$ time-steps ($F(\alpha_{boil}|x_{fire},a_0,t)$ equals $0$ for all $t>50$, $1$ otherwise), making the task non-stationary non-Markovian. The state-space on the right shows the optimal path (red dotted arrows). The physiological state-spaces have 55 states, so the agent cannot solve one task and then the other, rather, SPA must interleave goals for each task in order to satisfy both tasks before starving or dying of thirst. Each sub-goal is assumed to take one time-step, and we will ignore that, in reality, coconuts have thirst-quenching liquid inside, and assume it is only used for calories.}
\label{fig:vertical}
\end{figure}

To compose levels of abstraction, let $\Sigma$ be a binary vector state-space, $F:(\mc X \times \mc A \times \mc T) \times \mc A^{1} \rightarrow [0,1]$ be an action-availability function for first-order goals $\mc A^{1}$, and $\bar{F}:(\Sigma \times \mc T) \times \mc A^{2} \rightarrow [0,1]$ be the second-order multi-goal function for second-order goals $\mc A^{2}$ produced when achieving an accepting state-time for a non-stationary non-Markovian task.  Then the full hierarchical operator is given as:
\begin{align*}
    P_s(\textbf{s}'|\textbf{s},a,t)=\sum_{\bar{\boldsymbol{\alpha}},\alpha}P_r(\textbf{r}'|\textbf{r},\bar{\boldsymbol{\alpha}},t)\bar{F}(\bar{\boldsymbol{\alpha}}|\boldsymbol{\sigma},t)P_\sigma(\boldsymbol{\sigma}'|\boldsymbol{\sigma},\alpha)F(\alpha|x,a,t)P_x(x'|x,a,\zeta(\textbf{r}),t),
\end{align*}
where $\textbf{s}=(\textbf{r},\boldsymbol{\sigma},x)$. 
Note that the operator $P_\sigma$ can encode multiple binary tasks into one space. For example, in figure \ref{fig:vertical} we have two binary vector spaces $\Sigma_1$ and $\Sigma_2$ with corresponding operators $P_{\boldsymbol{\sigma},1}$ and $P_{\boldsymbol{\sigma},2}$ which can be combined into one binary vector space: $P_\sigma(\boldsymbol{\sigma}'|\boldsymbol{\sigma},\boldsymbol{\alpha}) = P_{\boldsymbol{\sigma},1}(\hat{\boldsymbol{\sigma}}'|\hat{\boldsymbol{\sigma}},\alpha)P_{\boldsymbol{\sigma},2}(\tilde{\boldsymbol{\sigma}}'|\tilde{\boldsymbol{\sigma}},\alpha)$, where the full vector $\boldsymbol{\sigma} = \hat{\boldsymbol{\sigma}} \circ \tilde{\boldsymbol{\sigma}}$ is the concatenated binary vector. Thus, we can have multiple tasks each corresponding to different goal variables in $\mc A^{2}$ driving different higher-level dynamics.

We can encode the task-acceptability conditions into $\bar{f}_{\dg}$. For our tasks, we will use the Boolean Ordered Goal (BOG) task formalism developed in Ringstrom et. al \citeyear{ringstrom2020jump}.
For a logical bit-vector state-space like $\Sigma$, a BOG task $\mathscr{B}$ is defined as a set of accepting states (the task-states $\Sigma_B$) and a set of precedence rules $\mc C$,
\begin{align*}
    \mathscr{B} = (\Sigma_B, \mc C), 
\end{align*}
which is parameterized by a Boolean statement $B$ written in disjunctive normal form, e.g. $B=(\sigma_0^T \land \sigma_1^T) \lor (\sigma_0^T \land \sigma_2^T)$. Each conjunction encodes the state (T or F in the superscript denotes values $1$ or $0$) of a bit in $\boldsymbol{\sigma}$ (denoted by the subscript $i$, e.g. $\sigma_i$ is the $i^{th}$ bit of $\boldsymbol{\sigma}$) for $\bar{f}_{\dg}$ to be non-negative evaluated on a given $\boldsymbol{\sigma}$. The task-states $\Sigma_B \subseteq \Sigma$ is the set of accepting task states corresponding to conjunctions of $B$ encoded in $\bar{f}_{\dg}$, and the set $\mc C$ are the set of precedence rules encoded in the structure of $P_\sigma$ (dashed lines on the cube in figures \ref{fig:vertical} and \ref{fig:subimation} denote that the transition has been disallowed). 
For example, $\sigma_1^T \prec \sigma_2^T$ means the first bit cannot be flipped to $1$ if the second bit is already $1$. 
For a TG-CMDP, it is not necessary that the intermediate state-space is a logical task (it could be an automata, for instance), nor is it a requirement that the highest-level space is a physiological state-space, however, we will use these state-spaces in this section to demonstrate vertical transition operator composition. We will also use the convention provided by Ringstrom \citeyear{ringstrom2020jump} that goals have \textit{types} such as \texttt{axe}, or \texttt{red}, and therefore precedence rules can be expressed over the types in order to enrich the semantics for the figure illustrations, e.g. $\texttt{water} \prec \texttt{fire}$ implies that the bit corresponding to water must be flipped before the bit corresponding to fire.

The operator $P_s$ defines the full three-level system. SPA can use the exact same policy and feasibility function decomposition of earlier sections and compute a state-time feasibility function set on $\mc X$, and forward sample policies to optimize a plan. If we have two or more tasks, naturally one might wonder if we can and plan at an even higher-level with \textit{second-order plans}, i.e. sequences of first-order plans $\bar{\rho}=(\rho_2,\rho_1,...)$, where a first-order plan $\rho$ is a sequence of closed-loop polices. This is indeed possible, but it introduces questions of optimality when using second-order plans, and we will not pursue this further. One can construct a hierarchical problem with two tasks that need to be completed in order to complete a second-order task, however, the optimal solution could require the agent to interleave policies for each task in order to complete both on time, due to how the two tasks are embedded in the world. If we optimize a first-order plan for each of two tasks in the hierarchical problem using only a set of low-level policies, then interleaving progress between the two tasks is possible. When using precomputed first-order plans for each task optimized separately, interleaving is not possible. Abstraction in this sense is would be clearly beneficial though for very large hierarchies, considering that chunking sequences would decrease the tree roll-out to find a solution. Future work could address the benefits of using second-order plans and the conditions for their optimality.

We demonstrate this interleaving task problem in Figure \ref{fig:vertical}, where SPA has two possible tasks that it can perform: BOG Task 1, $\mathscr{B}_1$ (green squares), require the agent to collect a knife and step-ladder in order to obtain and eat a coconut from a palm tree; BOG Task 2, $\mathscr{B}_2$ (blue squares), requires the agent to collect water and a kettle, to boil water at the fire (for sanitizing the water through boiling it in the tea-pot) so that the agent can drink clean water. Task 1 has a precedence constraint $\texttt{knife}\prec \texttt{coconut}$ and $\texttt{ladder}\prec \texttt{coconut}$, and Task 2 has precedence constraints $\texttt{water}\prec \texttt{fire}$ and $\texttt{kettle}\prec \texttt{fire}$. We can see that SPA's initial internal state $\textbf{r}_i$ is such that any trajectory that does Task 1 first and then Task 2, or vice-versa, will cause it to die of starvation or dehydration, however, if it interleaves the tasks, it can satisfy both tasks before dying.  Note that the items can have their own time availability windows, and we model the fire as having a finite temporal availability before it extinguishes, making the problem non-stationary non-Markovian\footnote{On a technical note, there are alternative ways this problem could be formalized. The tasks could be repeated depending on how the operators $P_{\boldsymbol{\sigma},1}$ and $P_{\boldsymbol{\sigma},2}$ are defined.  If the agent completes a task by collecting, boiling, and drinking water, then the fire and kettle will still exist, but the water will be consumed, and so completing the task should reset the bit corresponding to water to zero, but not the kettle and fire, one-step after it is achieved. The same would apply the knife and ladder as items, but the coconut is consumed and thus the bit must flip back to zero. The fire can, alternatively, not be defined as an ``item'', but rather as a static feature of the environment to boil the water. In this case, one would not represent fire as a bit, rather one could encode the task so that it could only be completed at the fire state, defined in $\bar{f}_{\dg_2}$.}.

To conclude, an agent built on a foundation of VBEs and OBEs, therefore, has two major kinds of decision process that it could engage in at any given moment to optimize a policy; should an agent choose to optimize a new policy and feasibility function for an unknown feature of the world, or should it simply optimize a policy to increase valence with the representations it already has? This leads us to an interesting conclusion: while a TG-CMDP task can be designed and specified by an engineer, it can also, in principle, be motivated by the \textit{potential} valence \textit{if the agent were to learn the mapping between low- and high-level state-spaces or learn new state-space transition dynamics}. We will return to this point in section \ref{sec:lifelong} when we discuss life-long learning, and the incentive to learn new high-level transition dynamics.

\section{High-level Reasoning with Sublimated TG-CMDP Solutions} \label{sec:sublimation}

SPA, being built on a the foundation of a TG-CMDP, is also capable of high-level \textit{reasoning}. Here by reasoning, we are simply referring to a process of applying computation to a part of the problem to constrain or direct the search for a solution to the entire problem. One important feature of the TG-CMDP is that it has a property called \textit{sublimation}. A \textit{sublimated problem} is when we take a TG-CMDP and only solve it on one of the state-spaces with a \textit{sublimated availability function} that is only a function of variables of that space. That is, if the original problem is $\mathscr{M}=(\Sigma \times \mc X, \mc A, P_s,\bar{f}_{\dg})$, the sublimated problem for space $\Sigma$ is $\mathscr{M}_{sub}=(\Sigma, \mc A^{1}, P_\sigma,\bar{f}_{\dg,sub})$, where the sublimated availability function $\bar{f}_{\dg,sub}(\boldsymbol{\sigma},t):=\max_{x,a}\bar{f}_{\dg}(\boldsymbol{\sigma},x,a,t)$ discards all other variables other than $\boldsymbol{\sigma}$ and $t$ (here, we just drop $x$ and $a$) from the original function $\bar{f}_{\dg}$ and retains the maximum possible feasibility over the discarded variables. We can then prove a theorem that the cumulative feasibility function of a sublimated problem must bound the full hierarchical problem from above. Put simply: if a problem is abstractly impossible from a high-level state, then it cannot be achieved from that high-level state when embedded in the world. Thus, sublimation exploits the fact that the OBE feasibility functions communicate reachability, rather than path-dependent cumulative quantities along a policy trajectory. If $\Sigma$ is a higher-level state space, then a sublimated feasibility function $\kappa_{sub,\boldsymbol{\sigma}}$ solved only on must bound the full hierarchical feasibility function $\bar{\kappa}$.  The theorem is stated as:

\begin{mytheo}{Sublimation Theorem}{theoexample}
If $\bar{\mathscr{M}}=(\Sigma \times \mc X, \mc A, P_s,\bar{f}_{\dg})$ is a full hierarchical TG-CMDP, and $\mathscr{M}_{sub}=(\Sigma, \mc A^{1}, P_\sigma,\bar{f}_{\dg,sub})$ is a sublimated TG-MDP using $P_\sigma$ on space $\Sigma$ derived from $\bar{\mathscr{M}}$, where $\bar{f}_{\dg,sub}(\boldsymbol{\sigma},t) := \max_{x,a} \bar{f}_{\dg}(\boldsymbol{\sigma},x,a,t)$, then:
$$\bar{\kappa}(\boldsymbol{\sigma},x,t)\leq \kappa_{sub}(\boldsymbol{\sigma},t),$$
where $\bar{\kappa}$ is the full hierarchical cumulative feasibility function and $\kappa_{sub}$ is the sublimated cumulative feasibility function.
\end{mytheo}

\begin{figure}[h]
\centering
\includegraphics[width=\linewidth]{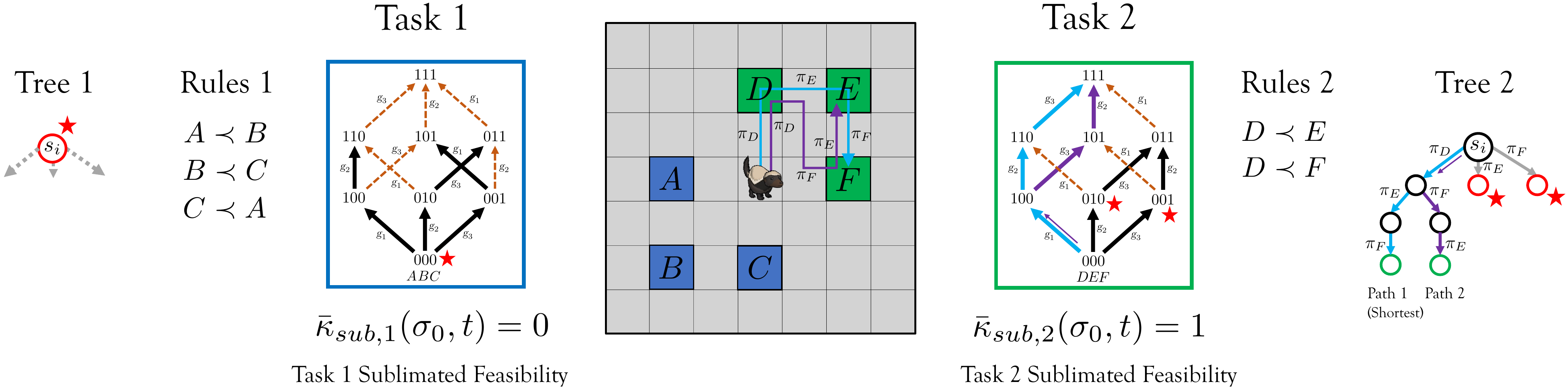}
\caption{Sublimation: 2 tasks each require the completion of three goals, recorded in a bit vector space where the agent starts at $\boldsymbol{\sigma}_0=000$. A rule (encoded by a brown dashed line) $A \prec B$ is interpreted as: flipping $A$ to $1$ requires $B$ to be $0$. Reverse transitions are not allowed by the transition operator $P_\sigma$. Task 1 encodes 3 precedence constraints which make the task impossible solely in the binary vector space as indicated by the red star at $\boldsymbol{\sigma}_0$, therefore the agent does not need to sample low-level polices $\pi_A$, $\pi_B$, or $\pi_C$ which drive the dynamics of $P_{\boldsymbol{\sigma},1}$. Task 2 only has 2 precedence constraints, and thus $P_{\boldsymbol{\sigma},2}$ has feasible paths from $000$ to $111$, so we can forward sample low-level polices $\pi_D$, $\pi_E$, or $\pi_F$ to find a sequence that completes the task. Notice that if the agent samples $\pi_E$, or $\pi_F$ first, transitioning $\boldsymbol{\sigma}$ for Task 2 from $000$ to $010$ or $001$ (red stars), then the sublimated feasibility from those states is $0$ and forward sampling will be terminated, as shown in the tree, where each node is a full hierarchical state vector $\textbf{s}$. The plan $\rho_{DEF}=(\pi_D\pi_E\pi_F)$ is feasible, as well as $\rho_{DFE}=(\pi_D\pi_F\pi_E)$, where $\rho_{DEF}$ can be selected as the time-minimizing plan.}
\label{fig:subimation}
\end{figure}
With this theorem, we can see that an agent can quickly rule out policies to sample by solving a partial problem in a higher-level state-space, as shown in figure \ref{fig:subimation}. This figure shows two tasks: Task 1 is impossible to complete, and so the agent never needs to try to forward sample any policies because the entire task is infeasible, $\bar{\kappa}_{sub,1}(\boldsymbol{\sigma}_0,t)=0$. Task 2 can be solved, but the sublimated feasibility function $\bar{\kappa}_{sub,2}$ can help constrain the roll-out of policies, where the red circles in the tree indicate that the task is infeasible from these states (and thus roll-out is terminated). The agent can select the time-minimizing policy that traverses the states in the order DEF.

It is important to In RL there is a practice of defining pseudo-rewards on sub-goals in order to create incentives to complete a task \cite{sutton1998introduction}. This could run into problems where an impossible task will still have rewards that the agent can accumulate, which will cause the agent to pursue it anyways. However, this does not occur with the TG-CMDP if the agent can reasons abstractly about sublimated feasibility. We see sublimation as an important tool for abstract and symbolic reasoning, which has been called for as a necessary ingredient in artificial intelligence for more general agents \cite{marcus2018deep, marcus2020next}. If natural or artificial intelligence operates in a compositional model-based manner, then computing sublimated solutions can act to refute or justify the relevance of a task or policy. 

\section{Relationship between TG-MDPs, VBEs and Multi-Objective RL}\label{sec:MORL}

\begin{figure}[h]
\centering
\includegraphics[scale=0.3]{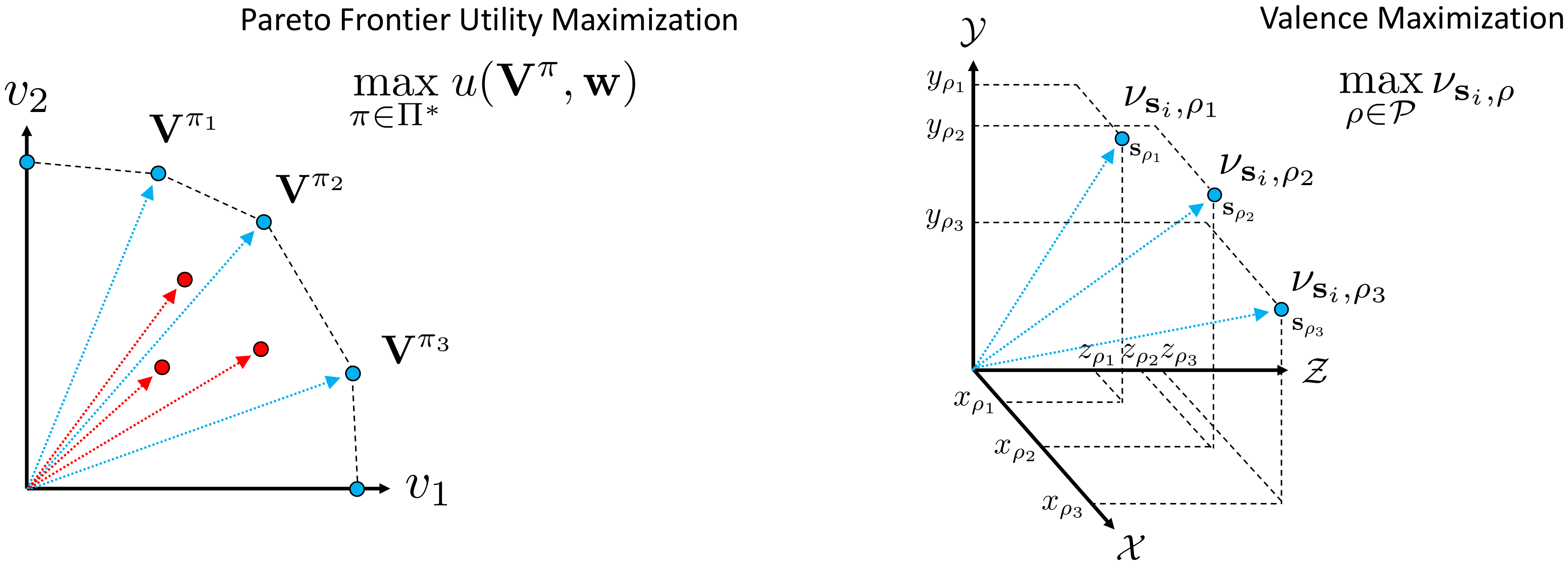}
\caption{Pareto Frontier Optimization vs. Valence Optimization. Left: An optimal policy can be computed from a set of policies, $\Pi^*$, lying on a Pareto Frontier, (blue points connected by dashed lines; red indicates interior ``dominated'' points) by evaluating the polices with respect to a utility function $u$ which is a function of weights $\mathbf{w}$. Right: product-space valence optimization is multi-dimensional (and multi-goal) but single-objective optimization, so it does not require a Pareto frontier or weights, and instead we optimize the valence of a plan, which transports the agent from $\textbf{s}_i$ to a final state vector $\textbf{s}_\rho = (x_{\rho},y_{\rho},z_{\rho})$ (or vector probability distribution). Notice that the axes represent objectives as \textit{value} on the left, and \textit{state} on the right.  Thus, on the right, a point farther away from the origin does not necessarily have higher valence, rather it depends on the \textit{structure} of a product space transition operator. Life-long valence-maximizing agents build up knowledge in the form of a product-space transition operator, and justify which state vectors in the product-space are best given knowledge of their internal functional organization.}
\label{fig:Pareto}
\end{figure}

It is natural to compare SPA to a Multi-objective RL (MORL) agent, given that the physiological dimensions of SPA can be viewed as multiple objectives. MORL is a multi-objective paradigm where agents have a vectorized value function $\mathbf{V}^{\pi}:\mc X \rightarrow \mathbbm{R}^n$ corresponding to a reward function vector $\mathbf{R}$, where $\mathbf{V}^\pi$ can be considered a set of individual value functions corresponding to the $n$ dimensions \cite{hayes2021practical}. It is typical to obtain a scalarized value function $V^{\pi}_u$ by using a utility function $u$ with a weight vector $\textbf{w}$, $V^{\pi}_u =u(\mathbf{V}^\pi,\textbf{w})=\mathbf{w}^T\mathbf{V}^\pi$ (where non-linear utilities are also possible). This means that an agent can compute a policy under a change in the agent's priorities, represented by the weights. Policies can be plotted in multi-dimensional value-space and a utility function can be constructed to choose a policy off of a Pareto frontier. The Pareto frontier is the set of value function vectors and associated policies that cannot be improved along a given dimension without sacrificing improvement along a different axis. The frontier represents a space of trade-offs where every \textit{dominated} point off of the frontier is clearly worse in all dimensions. Naturally, to select a policy from the frontier, one must \textit{justify} which point is preferable over all others. This justification is given by the utility function (though, Pareto frontiers are not always used in MORL, as they can be difficult to enumerate especially in continuous spaces). Of course, when using preference weights we run into a similar problem faced in standard RL, the problem of explaining where the preferences are derived from. In figure \ref{fig:Pareto}, we can see how SPA, being built on the foundation of the VBE, does not require weights or a frontier. SPA's valence maximization optimization is multi-goal, but single objective, and the preferences for various outcomes are dictated by the valence of the entire product space of the agent. 

An important difference between MORL and TG-CMDP agents is that MORL agents represent the alternative objectives value functions for policies on $\mc X$, whereas TG-CMDP agents represent these objectives as state-spaces in $\mc R = \mc W \times \mc Y \times ... \times \mc Z$ controlled by policies on $\mc X$ and framed as a reachability problem. This difference between value-objective and state-objective (state-space) is significant because in MORL value functions are derived from Bellman equations that have an additive form that results in the accumulation of a quantity over time without any imposed structure on the accumulation (other than discount factors). Alternatively, the TG-CMDP represents the effect of following some policy in other state-spaces \textit{with their own transition dynamics} that are not necessarily required to have a chain-like structure. This means that, while TG-CMDPs could use infinite chain-like state-space to record unbounded reward accumulation just like a value function, they can also use a state-space to record any other kind of process that cannot be modeled as reward accumulation. For example, in some video-games, an agent may have a capacity (such as $999$) of how much money it can carry. The designer of a MORL agent may model wealth as one of many possible objectives using a standard reward approach, representing the accumulation of wealth independent of limits to the amount of wealth that the agent can effectively posses (or physically carry and transport). Alternatively, the TG-CMDP can represent processes such as (but not limited to) wealth or energy dynamics with floor or ceiling limits, or bit-vector dynamics for recording object possession, making the TG-CMDP qualitatively different, but more expressive. Also, because alternative objectives are represented as state-spaces, all high-level states of a TG-CMDP agent can be used to condition its policy, and so they can also be used for new tasks (spending money from a previous task to obtain an item in a new task) whereas the value functions of MORL agent do not serve this role.  This is an important property for open-ended life-long agents which are constantly updating the states of many state-spaces.

\section{A Minimal and Ideal Model of Valence-based Life-long Learning}\label{chapter:lifelong}

Up until this point, optimizing a plan to increase valence is a process that has taken place with known dynamics.  However, the world affords many opportunities to learn and control new state-spaces. 
We argue that information seeking should be in the intrinsic interest of the agent, and that the TG-CMDP formalism allows for the accumulation of skills and knowledge because of its compositional and reusable feasibility functions, as we discussed in chapter \ref{sec:compositionality}. Such knowledge of newly observed dynamics in other spaces can be integrated into a planning architecture over time. We conclude this paper by demonstrating a minimal and idealized life-long learning algorithm supported by valence and TG-CMDPs. 

SPA has a number of functions which could be learned, such as the low-level dynamics $P_x$, action-availability function $F$, or any higher-order transition operator ($P_{\boldsymbol{\sigma}}$, $P_y$, etc.). We emphasize that our focus here will be to highlight the relevant aspects of our theory as it pertains to valence-based life-long learning to set the stage for future algorithms, \textit{not} to provide a comprehensive theory which address all problems of learning.
Thus, our life-long algorithm assumes a fixed discrete state-space model constituting the world and it assumes that the agent has perfect observations of the features on the underlying state-space---the state-space of the world will not expand over time, which is of course necessary for a true life-long algorithm. Instead, we present a simple learning algorithm to demonstrate the important properties of learning at a high-level of abstraction. We will assume that the low-level transition operator $P_x$ is known to SPA (including environment variable $e$) along with the action-availability function $F$ and the null dynamics of all high-level spaces. We also assume that SPA has knowledge of environmental features in the state-space, and can compute an aggregate state-time feasibility function for the features. Features $\theta$ will simply be lists of attributes associated with the goal variable ``visible" at the given states in $\mc X$, e.g. $\theta=\{\text{green},\text{apple}\}$ would be a green apple at a state, and eating the green apple would have a consistent state-transformation. Thus, any $\alpha$ will have a feature set $\theta$, including the null action $\alpha_{\epsilon}$, which is an empty set $\theta_{\epsilon}=\{\}$.  Given that the availability function maps the state-action-times to a goal variable, it also implicitly maps these tuples to the features (through the one-to-one correspondence). We will refer to the feature associated with state $x$ as $\theta_x$, and $\Theta$ is the set of all known features.

We also assume that when SPA learns something about a higher-level transition operator $P_\sigma$, that it will directly learn a function $\psi_{\alpha}$, called the \textit{state-transformation} function, that parameterizes $P_\sigma$.  For example, if $P_\sigma$ is a binary vector transition operator, a state-transformation $\psi_{\alpha_i}$ can be defined as an $\texttt{XOR}$ ($\oplus$) function which compares the $i^{th}$ bit (denoted as a superscript) in $\boldsymbol{\sigma}_j^i$ to another vector $\boldsymbol{\sigma}_k^i$ and applies logical \texttt{AND} ($\land$) to all other bits. That is, $\psi_{\alpha_i}$ returns $\texttt{True}$ if only the $i^{th}$ bit differs in the vector, and so it formalizes a bit-flip operation, 
\begin{align*}
    \psi_{\alpha_i}(\boldsymbol{\sigma}_j,\boldsymbol{\sigma}_k) := (\boldsymbol{\sigma}^i_j\oplus \boldsymbol{\sigma}^i_k)\bigwedge_{\ell \neq i}(\boldsymbol{\sigma}^\ell_j\land \boldsymbol{\sigma}^\ell_k).
\end{align*}
If we have a set $\boldsymbol{\psi} =\{\psi_{\alpha_1},\psi_{\alpha_2},...,\psi_{\alpha_n},\psi_{\alpha_\epsilon}\}$\footnote{The null state-transformation $\psi_{\alpha_\epsilon}$ is defined without $\oplus$ in order to induce null dynamics.} of state-transformations associated with each goal-action $\alpha_\dg \in \mc A^{1}$, then the bit-flipping operator $P_\sigma^{\boldsymbol{\psi}}$ will be defined as:
\begin{align*}
    P^{\boldsymbol{\psi}}_{\boldsymbol{\sigma}}(\boldsymbol{\sigma}_k|\boldsymbol{\sigma}_j,\alpha_i) = \{1 ~~\text{if}~~\psi_{\alpha_i}(\boldsymbol{\sigma}_j,\boldsymbol{\sigma}_k)=\texttt{True},~~0~~\text{otherwise}\},\quad \forall \psi_{\alpha_i}\in \boldsymbol{\psi}.
\end{align*} 

Alternatively, as we will use in an example, $\psi_{\alpha_{eat}}$ can be a function that acts on a physiological state-space $P_y^{\boldsymbol{\psi}}$ defined by $\boldsymbol{\psi}=\{\psi_{\alpha_{eat}},\psi_{\alpha_{\epsilon}}\}$:
\begin{align*}
    &\psi_{\alpha_{eat}}(y_j,y_k) := (y_k = y_{max}),\\
    &\psi_{\alpha_{\epsilon}}(y_j,y_k) := (y_{\text{max}(j-1,0)} = y_{k}),\\
    &P_{y}^{\boldsymbol{\psi}}(y_k|y_j,\alpha) := \{1 ~~\text{if}~~\psi_{\alpha}(y_j,y_k)=\texttt{True},~~0~~\text{otherwise}\},\quad \forall \psi_{\alpha}\in \boldsymbol{\psi},
\end{align*}
where $\delta$ is a Kronecker delta function used to compare states, and the resulting $P_y$ will map any state $y_j$ up to the fully satiated state $y_{max}$ (other functions are possible that increment the physiological state up or down by a fixed amount).

By providing an observation function $O:\mc A^{1} \rightarrow \Psi$, the agent can directly learn $\psi_{\alpha_i}$ from activated goal variables:
\begin{align*}
    O(\alpha_i)\rightarrow \psi_{\alpha_i}.
\end{align*}
If the agent \textit{observes} $\psi_{\alpha_i}$, then it means that it can expand the domain of $P$ to the goal-action $\alpha_i$ and set all state-transitions of $P$ in accordance with $\psi_{\alpha_i}$. Even though the function $\psi_{\alpha_i}$ is information that is generally not accessible to a real world agent, observing it, rather than inferring it, will serve to demonstrate a broader capacities of SPA in life-long learning, which is our primary interest here. Since $\psi_{\alpha}$ is a function, inferring it from observed state-transitions would be akin to program induction \cite{shaw1975inferring,lake2015human}, which could be employed for a more realistic process of learning. However we will leave inference over these functions from state observations for future work. We will also assume that the observation function is deterministic and $\psi$ functions are consistent across different environments. 

If the agent has any new $\boldsymbol{\psi}$ set, we can construct the operator $G_s^{\boldsymbol{\psi}}$ using the definition of $P^{\boldsymbol{\psi}}$. For example, if the agent's only high-level space is $\Sigma$, we would have:
\begin{align*}
\small&\xoverbrace{G_s^{\boldsymbol{\psi}}(\boldsymbol{\sigma}_f',x_f',t_f'|\boldsymbol{\sigma},x,t,\pi)}^{\text{Updated Goal Operator}}\\
&\quad\quad\quad\quad\quad := \sum_{\mathclap{\boldsymbol{\sigma}_f,x_f,t_f,\alpha}}\xoverbrace{P_\sigma^{\boldsymbol{\psi}}(\boldsymbol{\sigma}_f'|\boldsymbol{\sigma}_f,\alpha)}^{\mathclap{\text{Updated High-level Dynamics}}}P_x(x_f'|x_f,a_{\pi},\zeta(\boldsymbol{\sigma}_f),t_f)\omega_{\sigma}(\boldsymbol{\sigma}_f|\boldsymbol{\sigma},t_d)\widehat{\eta}_{c}(\alpha,x_f,t_f|x,t,\pi),
\end{align*}
where $t_f'=t_f+1$ and $t_d=t_f-t$. 

The algorithm we present will be an $m$-policy roll-out of a single policy in order to see if it can gather information about a feature under physiological constraints.  That is, we do not perform long roll-outs which condition on a potential belief updates from our observations, as one might in a Bayesian-RL framework \cite{ghavamzadeh2015bayesian} (as the agent will not know what the dynamics are until it observes them). Nevertheless, this simple heuristic algorithm demonstrates how information seeking behavior can be guided by the expected valence of learning. We will assume that for each policy $\pi$ in $\rho$, that the agent computes the expected valence $\nu_{\textbf{s},t,\pi}$ of arriving at the state with a feature, which is the valence of arriving at that state given the agent's knowledge of the effect of the feature on the product-space dynamics, plus a prior over the valence of the feature in the absence of knowledge:
\begin{align}
    \nu^{*}_{\textbf{s},t,\Theta}=
    \max_{\rho\in \mc P_m}\Bigg(\mathfrak{V}_n(G_s^{\boldsymbol{\psi}},G_s^{\boldsymbol{\psi}},Q^{\boldsymbol{\psi}}_m,\textbf{s},t,\rho)+ \xunderbrace{\sum_{\pi^i_g\in \rho_{uni}}\mathbbm{1}_{\bar{\Theta}}(\theta_{x_\pi})\expec_{x_g\sim Q_{s,\rho_i}}\expec_{\nu\sim p_{\nu}}\nu}_{\mathclap{\text{Sum of valence for unknown features.}}}\Bigg), \label{eq:lifelongplan}
\end{align}
where $\expec_{x_g\sim Q_{s,\rho_i}}$ is the plan-operator expectation over the probability that the agent will be at the goal state $x_g$ under the (partial) plan $\rho_{i}=(\pi^{1},...,\pi^{i})$ sequentially sampled with $G_s^{\boldsymbol{\psi}}$, and $\expec_{\nu\sim p_{\nu}}=\expec_{\nu\sim p_{\nu}(\cdot|\theta_{x_g})}$. This equation is the same as a standard valence computation (equation \ref{eq:max-plan}) except there is an additive valence prior for when the agent does not know what transformation the feature $\theta_{x_\pi}$ will induce from state $x_{\pi}$, the final state of following $\pi$. The sum is over $\rho_{uni}$ which are the unique policies in $\rho$, so that we do not count the same valence multiple times if the agent repeats the same policy. 

Notice that when the agent does not know the feature effect, it still computes the resulting valence of traveling to the state under its current understanding of $\boldsymbol{\psi}$ respecting the relevant physiological and external state changes---the prior over valence is the agent's anticipated valence given new knowledge. The set $\Theta$ is the set of features $\theta$ corresponding to the current knowledge of dynamics $\boldsymbol{\psi}$, and $\bar{\Theta}$ is the set of unknown features. $\mathbbm{1}_{\bar{\Theta}}$ is an indicator function that returns $1$ when feature $\theta_{x_{\pi}}$ is in the unknown feature set $\bar{\Theta}$, and therefore acts to add anticipated valence for an unknown feature (which can be positive or negative, depending on the definition of $p_{\nu}$). The operator $G_s^{\boldsymbol{\psi}}$ is the current product-space goal-operator under the  parameterization $\boldsymbol{\psi}$ of known state-transformations. 

One could make the case that the introduction of a prior over feature-specific valence is counter to the spirit of this paper, which has sought to derive value signals solely from the intrinsic structure of the agent.  While this is true, we acknowledge that real world biological agents may have in-born curiosity signals or preferences for shapes, colors, sounds, and tastes, among other features, and we maintain that signals of this nature can be combined with product-space valence. Furthermore, it also might be possible for the agent to infer its valence-priors for new features using past observations of similar features and their observed valence. 

\subsection{A Simple SPA Learning Algorithm and Remapping Example}\label{sec:lifelong}

\begin{figure}[!h]
\centering
\includegraphics[scale=0.38]{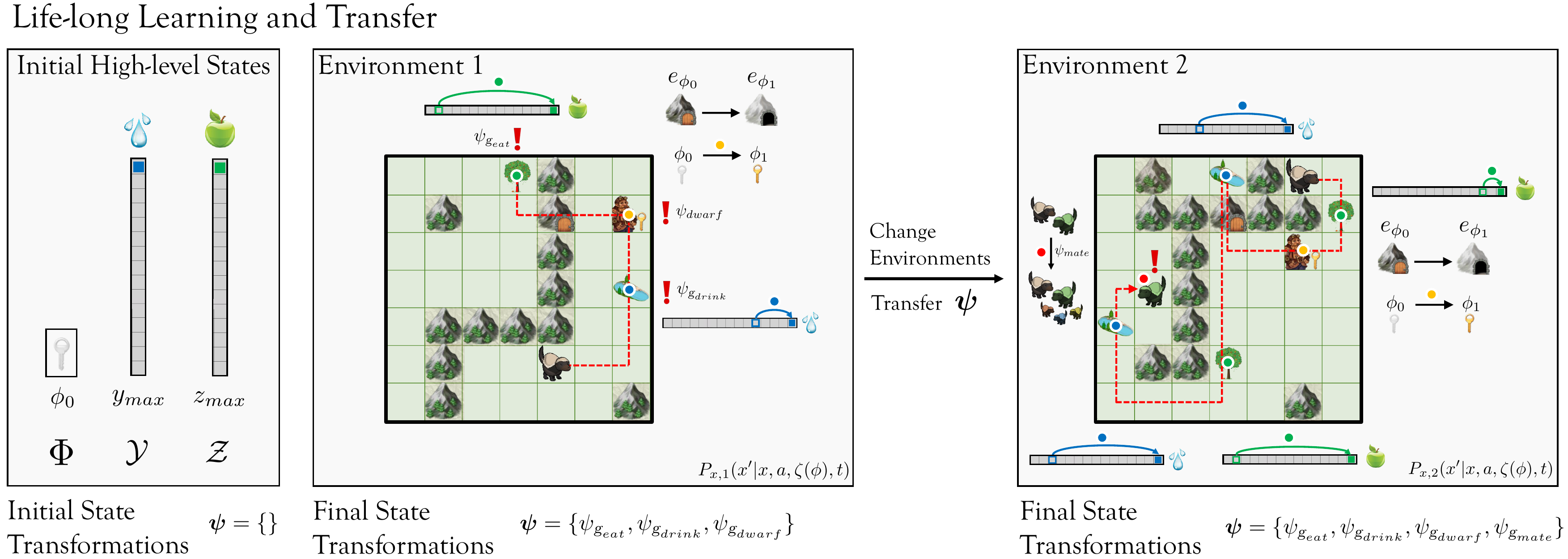}
\caption{Learning, Transfer and Remapping: Here we show an agent employing the life-long algorithm. We assume Stoffel knows the null dynamics and can generate low-level state-time feasibility functions for the known goals in the map to produce $\widehat{\eta}$. In environment 1, Stoffel is fully satiated and optimizes a plan to first go the lake. The red exclamation points indicate that the function $\psi_{\dg}$ has been observed and the agent updates $G_s$ and re-plans. Stoffel then travels to the dwarf and learns that he can get a key (changing the object state to $\phi_1$ and mode to $e_{\phi_1}$), and then he travels to the tree an learns that he can eat an apple. After the episode Stoffel is transferred to a new environment with his state-transformation set, in which he can readily plan without learning. Knowing he can get a key from the dwarf, he can chain $6$ policies together to reach Hammie, in which he learns a new state-transformation $\psi_{mate}$. Stoffel could not be confident that he can reach Hammie without the knowledge that he can obtain a mountain-pass key from the Dwarf, which he needs in order to drink from the first lake to avoid dehydration.}
\label{fig:life-long}
\end{figure}

The Life-long SPA algorithm (algorithm \ref{alg:open-ended-spa}) presented here is fully model-based. It assumes knowledge of all transition operators and environmental features (trees, lakes, etc.), but it does not assume knowledge of the feature's effect on the higher-level state-space. The algorithm is formalized to accommodate program in future instantiations, but these complexities are avoided for now by the use of the observation function. At the beginning of the algorithm, SPA will only have knowledge of $P_x$ and the null dynamics for all other operators. If we assume a prior $p_{\nu}$, then the agent then computes the best plan with equation \eqref{eq:lifelongplan} and executes the plan. If the agent learns a new $\psi_{\dg}$ from observation (red exclamation point in figure \ref{fig:life-long}), then we can update $\boldsymbol{\psi}$ and $P^{\boldsymbol{\psi}}_s$ and $G^{\boldsymbol{\psi}}_s$, and allow the agent to interrupt its previous plan to re-plan in light of the updated information. 

\begin{algorithm2e}[h]
\DontPrintSemicolon
\SetKw{return}{return}
\SetKwRepeat{Do}{do}{while}
\SetKwData{conflict}{conflict}
\SetKwData{safe}{safe}
\SetKwData{sat}{sat}
\SetKwData{unsafe}{unsafe}
\SetKwData{unknown}{unknown}
\SetKwData{true}{true}
\SetKwInOut{Input}{input}
\SetKwInOut{Output}{output}
\SetKwFor{Loop}{Loop}{}{}
\SetKw{KwNot}{not}
\begin{small}
	\Input{Dynamics $P_x$, High-level operators $\mc Q = \{P_{\phi}, P_y, P_z,...\}$ action-availability function $F$, mode-function $\zeta$, final time $t_f$, horizon $m$, prior $p_{\nu}$}
    $\widehat{\eta}_c \leftarrow \texttt{create\_aggregate\_STFF}(P_x,\mc Q,F,\zeta)$\;
    $\Omega \leftarrow \texttt{create\_omegas}(\mc Q)$\;
    $\Theta \gets \emptyset$\;
    $t\leftarrow 0$\;
    $\boldsymbol{\psi}\leftarrow \{\psi_{\alpha_{\epsilon}}^y,\psi_{\alpha_{\epsilon}}^z,...\}$\quad\quad\#\text{Initialize null dynamics for all spaces}\;
	\While{$t<\inf$}
            {
	    $\rho \leftarrow \argmax_{\rho\in \mc P_m}\big(\mathfrak{V}_n(G_s^{\boldsymbol{\psi}},G_s^{\boldsymbol{\psi}},Q^{\boldsymbol{\psi}}_m,\textbf{s},t,\rho)+\sum_{\pi\in \rho_{uni}} \mathbbm{1}_{\bar{\Theta}}(\theta_{x_\pi})\expec_{x_g\sim Q_{s,\rho_i}}\expec_{\nu\sim p_{\nu}}\nu\big)$\;
	    \For{$\pi\in \rho$}{
	        $(\alpha_{\pi},\mathbf{r}_\pi,x_{\pi},t_{\pi})\leftarrow \omega_r(\cdot|\mathbf{r},t_\pi-t)\widehat{\eta}_{c}(\cdot,\cdot,\cdot|x,t,\pi)\quad\quad\#\text{Update State}$\;
    	    \If{$\theta_{x_\pi}\notin \Theta$}
    	    {
	        $\psi_{\alpha_{\pi}}\leftarrow O(\alpha_{\pi})\quad\quad\#\text{Observe State-Transformation}$\; 
	        $\boldsymbol{\psi}.append(\psi_{\alpha_{\pi}})$\;
	        $\Theta.append(\theta_{x_\pi})$\;
	        $\texttt{break\_for\_loop}\quad\quad\#\text{Re-plan after learning new dynamics}$\;
	        
    	    }
    	    $(\textbf{s}',t') \leftarrow \texttt{one\_step\_state\_update}(\alpha_{\pi},\mathbf{r}_\pi,x_{\pi},t_{\pi},\pi,\mathcal{Q}^{\boldsymbol{\psi}},P_x)$\;
    	    $(\textbf{s},t)\leftarrow (\textbf{s}',t')\quad\quad\#\text{Update state and time variables to latest}$\;
	    }
	}
\end{small}
\caption{Life$\_$Long$\_$SPA}
\label{alg:open-ended-spa}
\end{algorithm2e}

Figure \ref{fig:life-long} shows a simple example of life-long Stoffel running algorithm \ref{alg:open-ended-spa}, and then transfers to a new environment to plan with the learned knowledge in order to reach Hammie without any additional learning, which could not be achieved without prior learning, as Stoffel would die of dehydration and hunger. The example assumes that Stoffel can "see" all of the features of each environment, knows the dynamics model $P_{x}$ for each environment, and can compute feasibility functions for each environment with feasibility iteration. Stoffel has two physiological state-spaces with $15$ states, meaning he has an effective range of $14$ before he will die, and interacting with features takes one time-step. There are two features of physiological significance, the apple tree and the lake, that can be found in the state-space, along with a Dwarf who can offer a key, the possession of which is represented by the bit state $\phi$ parameterizing the environment variable $e_{\phi}$. Stoffel initially knows the null dynamics of all spaces, but does not understand that he can drink from the lake, eat from the tree, or obtain the key from the Dwarf. Stoffel starts out fully satiated and decides to obtain the food first, since the prior $p_{\nu}$ encodes new features as equally good, then he travels to the Dwarf and learns that he can open the door, which means he can then obtain water instead.  After Stoffel has learned the three operators in environment $1$, it is then transferred over to environment $2$ and the high-level transition operator actions are remapped to new state-time feasibility functions, and since Stoffel knows the map between features and transformations, he already knows how to act in environment $2$. Therefore, Stoffel can generate a plan to reach a new feature, Hammie, without needing to learn anything. If Stoffel did not learn the dynamics, he could not reach Hammie without dying of thirst, as he would not know how to reach the first lake to obtain water using the key to pass through the door.

\section{Discussion}
We have introduced SPA, a reward-free model-based autotelic agent that can dynamically plan in a large product space of variables in order to stay alive, and uses empowerment-gain to evaluate the quality of states produced by policy sequences. We showed how SPA can reason at a high-level of abstraction via sublimation, plan across a vertical hierarchy of logical tasks connected to physiological state-spaces, and how SPA could use the anticipated valence of obtaining new information in order to obtain new knowledge of transition dynamics. We intend that this framework will inspire lines of research, in both AI and computational neuroscience, centered around compositional hierarchical planning and empowerment.

A few questions will require further investigation. First, we use a time-minimizing policy basis for a set of goals, however, it remains to be proven whether or not this basis is sufficient to form a plan operator that has maximum empowerment with respect to its domain. Second, the hyper-parameters, $m$, for generating plans, and $n$ for evaluating empowerment are fixed and could be optimized, future work should address whether there is a good theory for setting the parameters. Third, product-space empowerment in our theory was limited to the case of deterministic operators because this allows us to compute empowerment via reachable-state counting without using the Blahut–Arimoto algorithm. While Blahut–Arimoto can be applied to our algorithm in the stochastic case, this is not ideal due to the size of the channel matrix which has to be constructed. Future research could be directed towards the question of whether or not there are methods for computing empowerment (or computing a bound for it) in the stochastic case using our operator factorization to avoid constructing the full channel's joint distribution. Lastly, in our high-level dynamics learning example, we simplified the problem by making the assumption that the agent directly learns the function that parameterizes the high-level transition operator. We did this for simplicity, but future work could infer the underlying transition structure directly from state observations. 

A few technical and philosophical questions arise with respect to SPA.  Some readers might wonder if SPA can be \textit{designed} to do specific tasks, such as cooking a meal, run errands, or completing a video-game.  While the TG-CMDP is a general decision process for computing solutions to non-stationary non-Markovian tasks and appropriate for such engineering problems, SPA's decision theoretic foundation is based on empowerment gain. If SPA is truly autonomous, then it could be incentivized to complete specific tasks so long as completing the task is worth it to SPA, which is determined by assessing its own hierarchical structure. A person could withhold basic resources from SPA unless a task is completed to the benefit of the person, perhaps contingent on SPA's capacity to possess awareness and theory-of-mind---this is an act many people may rightly consider to be immoral in real-world scenarios. Keep in mind, the validity of SPA's empowerment-gain computation hinges on the knowledge that it is derived from a measure of its own planning representations. A deceitful person (such as a computer scientist) could intervene and feed SPA an artificially adjusted high final-empowerment estimation for a state-vector predicted under a plan, but the person would be committing an act of deception and would be undermining SPA's agency. Presumably, only an agent which has access to the knowledge that it carried the computation out on itself could be said to be confident in its own motivations. We propose that agency necessarily depends on the potential to interrogate any external or internal signal to evaluate its significance with respect to the integrity of the agent's hierarchical architecture. Inhibiting such a process undermines the effort of creating agents with true autonomy. 

Our framework has several implications for the debate around the reward-hypothesis.  First, we agree with the position of Silver et al. that intelligence can be framed as optimizing the accumulation of a scalar quantity, however, in our framework we do not use a reward-function in either the policy optimization or the goal justification. The scalar, valence, is derived from computational work applied to the representations of an agent which correspond to a complex world-model of itself and the external world, not as an externally or internally received quantity like a reward function. Indeed, one could simply define reward as valence---in fact, as we have shown, a finite-horizon Bellman equation with this definition will simply have optimal polices that try to reach the most empowered state, in expectation (see A.\ref{appx:valence-BE}). However, we believe our framework should avoid the theoretical commitments (discount factors, value functions) of IHDR or other reward-based optimizations. There is no guarantee that IHDR captures all aspects of intelligence, and to the contrary, the factorization of our architecture is possible by exploiting the properties of a Bellman equation of a different form---the model-based foundation defining the optimization is crucial, and it is not clear how IHDR could achieve a similar result with the same or better computational complexity. Secondly, if we were to define reward as a valence, we see no reason for storing a value function that represents the accumulated total empowerment gain since the gain already reflects changes in the agent's hierarchical organization and state; many problems are non-stationary and it might be that for most problems of consequence, the agent will likely never return to the same state vector in large product space. Lastly, Silver et al. have argued that ``an effective agent may make use of additional experiential signals to facilitate the maximisation of future reward" \cite{silver2021reward}. Valence could be a kind of experiential signal, however it is unclear why valence, which is a well-defined quantity derived from an agent's acquired representations, must act in support of a reward function, which lacks an explanation of origin for \textit{any} given product space an agent might represent and compute its control policies on. 




It is also worth revisiting the point that while the valence function outputs a scalar, it is not a utility function like we find in MORL, where utility is defined as a weighted combination of value functions for multiple tasks \cite{hayes2021practical}. 
The bidirectional coupling between the internal and external transition operators renders one large operator; therefore, valence unifies the contribution of many seemingly disparate drives which are often regarded as distinct factors that need to be weighted together, as was suggested as an objection to RIEH \cite{vamplew2021scalar}.  Furthermore, since the internal states depend on the base state-space to be controlled, SPA incorporates environmental context into the evaluation of the quality an internal state, e.g. being in a state of low hydration is not intrinsically bad if there is a water source nearby to avoid the dehydration state; the same low-hydration state could be catastrophic if the agent is located in a desert 30 miles from an oasis---this contextual information is incorporated into hierarchical empowerment. Therefore, product-space valence sidesteps the problem of defining reward functions in RL and defining weighting functions on a set of value functions in MORL, with the benefit that the question of explaining the origin of such functions can in principle be disregarded depending on how the problem is framed. Product-space valence is not a linear or non-linear function of individual factors lying on a Pareto frontier. However, we see this not as a shortcoming, but as a reality. Agents must investigate and relieve the tensions arising from the competing needs of their many interconnected subsystems through reasoning, and an outcome of successful reasoning is \textit{explainable motivations} for the unity and expressiveness of the system, whereas utility from rewards on each subsystem will always require further explanation: where do the scalar weights come from, and what do they mean? 

On a closing note, we would like suggest that the coupled operators of SPA instantiate something akin to the concept of \textit{organizational closure}, which is a relationship of mutually dependent sub-systems, each enabling and conditioning each other's dynamics, where the full system, sustained by the ongoing activity of the components, serves as a locus of normativity to be maintained for self-preservation \cite{varela1979principles}. Typically, organizational closure refers to coupled systems that \textit{materially} support each other through the production of physical substrate (e.g. chemical reagents), which is not occurring in our state-space control models, and thus the analogy does not hold in that respect, however, states can represent relevant objects or quantities. Organizational closure has been argued to be necessary for the existence of normativity in autopoietic systems \cite{mossio2015biological}, and (as we mentioned in the introduction) for a theory of what Roli et al. calls an ``organismal agent," an agent capable of perceiving environmental affordances and ``select[ing] from a repertoire of alternative actions when responding to environmental circumstances based on its internal organization," as opposed to a mere input-output processing system \cite{roli2022organisms}. Indeed, our agent satisfies this aim and directs computational work towards the assessment of possible states of the world in order to make choices which promote the integrity of the agent as a whole, exhibiting a form of decision making which is mechanistically normative, but relativistic to an idiosyncratic planning architecture which has developed over time. This is possible under a particular ``intellectual phenotype" where the OBEs generating the agent's feasibility functions are formalized on the objective of reachability over reward-maximization, permitting the agent to both forward sample high-dimensional state vectors and evaluate goal states with the same compositional representations. We submit that intelligent systems which organize their behavior around the realization of states that are conceived of and rationalized, but yet to be experienced, possess agency of a teleological nature. Of course, since control-theoretic algorithms produce end-directed dynamics, they have a teleological quality \cite{rosenblueth1943behavior}. However, many have argued that a full account behavior in biology and neuroscience must explain how goal states are set, which requires a normative standard \cite{juechems2019does,yin2020crisis, roli2022organisms, jaeger2021fourth, mossio2017makes,kiverstein2022problem}. Here, we invoke the concept of teleology to convey the idea that a potential state in an agent's model of the world can play a causal role in bringing about its own realization with the normative consequence of enhancing the agent's capacity to act in a teleological manner in the future, a conception of self-determination that is consistent with arguments made by philosophers of biology and agency \cite{deacon2011incomplete,mossio2015biological,mossio2017makes}. Indeed, by setting goals with empowerment-gain, the expressiveness of SPA's planning representations (STFF factorization) from considered goal state-vectors is causally determining the behavior that brings about the state-vector.
This is why feasibility functions, which allow an agent to reason across the different state-spaces that define the agent, play such an important role in our theory, and by extension, why the Bellman optimization principles (e.g. reachability vs. reward-maximization) that underpin an agent's representations cannot be ignored---the structure of the availability signal (as opposed to a reward signal) determines the interpretation of the representation that is produced. With this considered, we anticipate that further investigation of the questions arising from our framework will advance our understanding of intelligent living systems, the nature of agency, and how to design life-long agents which exhibit these remarkable capacities.

\section*{Acknowledgements}
Thank you to Paul R. Schrater, Tatyana Matveeva, and Igor Krawczuk for helpful discussions and commentary on the content of this manuscript.

\newpage
\bibliography{references}
\bibliographystyle{apacite}

\clearpage

\newpage
\section{Appendix}
\subsection{Notation}\label{appx:notation}

For the Operator Bellman Equations in this paper, we will often refer to the "task" goal variable as $\alpha_\dg$, but in the context of the equations, we will often drop the $+$ subscript to avoid notational clutter. The OBEs only optimize the feasibility of achieving the task goal. The null-goal variable will be referred to as $\alpha_{\epsilon}$.  The time $T_f$ will refer to the fixed finite horizon parameter, and $t_f$ will refer to any final time of completing as task within $[0:T_f]$.

\subsection{Operator Bellman Equations}

Let $\mathscr{M}=(\mc X,\mc A,P_x,f_\dg)$ be a TG-MDP with horizon $T_f$.  The Operator Bellman Equations are given as:
\begin{align}
    &\kappa_{\pi_g}^*(x,t) = \max_{a\in \mc A} \left[f_{\dg}(x,a,t) + (1-f_{\dg}(x,a,t)) \sum_{x'} P_x(x'|x,a,t)\kappa_{\pi_g}^*(x',t+1) \right], \label{appx:obs-C-Bellman-kappa}\\
    &\pi^{**}_{\dg}(x,t) = \argmin_{a\in \mc A^*_{x,t}}\left[tf_{\dg}(x,a,t)+(1-f_{\dg}(x,a,t))~\sum_{\mathclap{x',x_f,t_f}}~ t_f~ \eta^{**}_{\pi_g}(\alpha_{\dg},x_f,t_f|x',t+1)P_x(x'|x,a,t) \right],\label{appx:obs-c-Bellman-pol}\\
    &\eta_{\pi_g}^{**}(\alpha_\dg,x_f,t_f|x,t) = (1-f_{\dg}(x,\pi_{\dg}^{**}(x,t),t))\sum_{x'} \eta_{\pi_g}^{**}(\alpha_{\dg},x_f,t_f|x',t+1)P_x(x'|x,\pi_{\dg}^{**}(x,t),t),\label{appx:graveyard-marg-eta}
\end{align}
where $\kappa^*:\mc X\times \mc T\rightarrow[0,1]$ is the optimal cumulative probability of solving the task, $\pi^{**}: \mc X\times \mc T\rightarrow \mc A$ is the optimal policy, and $\mc A^*_{x,t}$ is the $\argmax$ set of cumulative-maximizing actions from equation \eqref{appx:obs-C-Bellman-kappa}. 

Equation \eqref{appx:graveyard-marg-eta} is technically defined as:

\begin{align}
    \small\eta_{\pi_g}^{**}(\alpha_\dg,x_f,t_f|x,t) = \begin{cases}
        (1-f_{\dg}(x,\pi_{\dg}^{**}(x,t),t))\expec_{P_x(\cdot|x,a^{**}_{xt},t)}\eta_{\pi_g}^{**}(\alpha_\dg,x_f,t_f|x',t+1),& \text{If: } t_f>t\\
        f_g(x_f,\pi_\dg^{**}(x_f,t_f),t_f)& \text{If: } t_f=t\\
        0 & \text{If: } t_f<t
    \end{cases},
\end{align}
but this is omitted in the main OBE equation for compactness. Formally, the state-time feasibility function, $\eta_{\pi_g}^{**}$, is defined as: $$\eta_{\pi_g}^{**}: (\mc X \times \mc T)\times(\mc X \times \mc T \times \{\alpha_\dg,\alpha_{\epsilon}\})\rightarrow [0,1],$$ where $\alpha_{\epsilon}$ is a "null goal" which represents a goal variable induced when the agent fails the task.  However, solving the above equations will only compute $\eta_{\pi_{g}}$ for the task-goal $\alpha_\dg$ which defines the task.  Computing $\eta_{\pi_{g}}$ for the task failure event will be discussed in section \ref{sec:task_failure}.

\subsubsection{Boundary conditions}\label{appx:boundary}
Given that this is a finite-horizon problem, future feasibility after $t\geq T_f+1$ is defined as zero, i.e. $\kappa(x,T_f+\tau)=0, \forall \tau \in \mathbb{N}_+$.  Therefore, the state-time boundary conditions for \eqref{appx:obs-C-Bellman-kappa}, \eqref{appx:obs-c-Bellman-pol}, and \eqref{appx:graveyard-marg-eta} are given as:
\begin{align}
    \kappa_\dg^*(x_f,T_f)&=\max_a \left[f_g(x_f,a,T_f)\right]= f_g(x_f,a^*_t,T_f), &&\forall x_f \in \mc X, \\
    \pi_\dg^{**}(x,T_f) &= \argmax_a \left[f_g(x_f,a,T_f)\right]= a^*_t, && a\in \mc A^*_{x,T_f},\\
    \eta^{**}_{\pi_{\dg}}(\alpha_\dg,x_f,T_f|x_f,T_f) &= f_g(x_f,\pi_\dg^{**}(x_f,T_f),T_f)= f_g(x_f,a^{**}_t,T_f), &&\forall x_f \in \mc X, \label{appx:eta-boundary}
\end{align}

where $\mc T = [0,1,...,T_f]$, and $\mc A^*_{x,T_f}$ is the set of maximizing arguments for the $\kappa_\dg^*(x,T_f)$ (which will be a singleton set). It should be noted that when we say "boundary conditions" we mean all state-times pairs which include the final time $T_f$. 

\subsubsection{Relationship Between $\kappa$ and $\eta$}\label{appx:kappaeta}
Assume that we compute an optimal $\kappa^*$, $\pi^{**}$, and $\eta^{**}$. Using the boundary condition, \eqref{appx:eta-boundary}, we can substitute $\eta$ in for $f_{\dg}$ in the recursion \eqref{appx:obs-C-Bellman-kappa} but with the maximization dropped (since we have already optimized the three main equations), resulting in:
\begin{align}
    \kappa_{\pi_g}^*(x_t,t) = \eta^{**}_{\pi_{\dg}}(\alpha,x_t,t|x_t,t) + (1-f_{\dg}(x_t,a,t)) \sum_{x_{t+1}} P_x(x_{t+1}|x_t,a,t)\kappa_{\pi_g}^*(x_{t+1},t+1)\label{appx:kappa_sub_f_eta}
\end{align}

Focusing on the right side of the addition, we can unroll $\kappa$ another time-step, again substituting in $\eta$ in for $f_{\dg}$:
\begin{align}
    (1-f_{\dg}(x_t,a,t)) &\sum_{x'} P_x(x_{t+1}|x_t,a,t)\Bigg[ \eta^{**}_{\pi_{\dg}}(\alpha,x_{t+1},t+1|x_{t+1},t+1) \\
    \nonumber &+ (1-f_{\dg}(x_{t+1},a,t+1)) \sum_{x'} P_x(x_{t+2}|x_{t+1},a,t+1)\kappa_{\pi_g}^*(x_{t+2},t+2)\Bigg].\label{appx:kappa_expand}
\end{align}

After distributing we have,
\begin{align}
    (1-f_{\dg}(x_t&,a,t))\sum_{x_{t+1}} P_x(x_{t+1}|x_t,a,t) \eta^{**}_{\pi_{\dg}}(\alpha_\dg,x_{t+1},t+1|x_{t+1},t+1) \\
    \nonumber &+ (1-f_{\dg}(x_t,a,t))\expec_{x_{t+1}\sim P_x} (1-f_{\dg}(x_{t+1},a,t+1)) \expec_{x_{t+2}\sim P_x}\kappa_{\pi_g}^*(x_{t+2},t+2),
\end{align}
where the first term is equal to the definition of the feasibility function $\eta(x_{t+1},t+1|x_t,t)$ defined by the OBE equation \ref{appx:graveyard-marg-eta}. Substituting, we have:
\begin{align}
    =\eta(\alpha_\dg,x_{t+1},t+1|&x_t,t)\label{appx:kappa_unroll_2}\\
    \nonumber &+(1-f_{\dg}(x_t,a,t))\expec_{x_{t+1}\sim P_x}(1-f_{\dg}(x_{t+1},a,t+1)) \expec_{x_{t+2}\sim P_x}\kappa_{\pi_g}^*(x_{t+2},t+2).
\end{align}
Note that \eqref{appx:graveyard-marg-eta} implies that for any start state-time $(x,t)$ and final state-time $(x_{t_f},t_f)$ we can expand $\eta$ into a sequence of expectations of not achieving the goal under the policy, multiplied by the probability of achieving the goal:
\begin{align}
    \nonumber\eta_{\pi_g}(\alpha_\dg,x_{t_f},t_f|x_{t},t)=(1-f_{\dg}(x_t,a_{\pi},t))\expec_{x_{t+1}\sim P_x}(1-f_{\dg}(x_{t+1},a_{\pi},t+1))~...\\ ...~\expec_{x_{t_f-2}\sim P_x}(1-f_{\dg}(x_{t_f-2},a_{\pi},t_f-2))\expec_{x_{t_f-1}\sim P_x}\eta_{\pi_g}(\alpha_\dg,x_{t_f},t_f|x_{t_f-1},t_{f}-1).\label{appx:eta_expand}
\end{align}
Recall that \eqref{appx:kappa_unroll_2} is the right side of the addition in \eqref{appx:kappa_sub_f_eta}, and every time we expand $\kappa$ and distribute, we get a term which can be written as an extended $\eta$ function, plus the expectation of a future $\kappa$ term. Therefore, we can keep unrolling \eqref{appx:kappa_unroll_2} out until $T_f$, and apply \eqref{appx:eta_expand} each time, resulting in the sequence:
\begin{align*}
    \kappa_{\dg}^*(x,t) = \eta_{\pi_g}^{**}(\alpha_\dg,x_t,t|x,t)+\eta_{\pi_g}^{**}(\alpha_\dg,x_{t+1},t+1|x,t)+...+\eta_{\pi_g}^{**}(\alpha_\dg,x_{T_f},T_f|x,t),
\end{align*}

which can be written as a sum over the final state-times, leading to the relationship between $\kappa$ and $\eta$:
\begin{align*}
    \kappa_{\dg}^*(x,t) = \sum_{x_f,t_f}\eta_{\pi_g}^{**}(\alpha_\dg,x_f,t_f|x,t).
\end{align*}
The cumulative feasibility function (which is the cumulative probability of completing the task) is simply the sum of the first goal-satisfaction state-time events of the state-time feasibility function for a policy $\pi_\dg$.

\subsubsection{Task Failure Probabilities}\label{sec:task_failure}

The feasibility function computed by \eqref{appx:graveyard-marg-eta} represents the probability of finishing the task at a given state-time (inducing the task-goal variable $\alpha_\dg$, with the $+$ dropped in the OBE equations for clarity).  We can also compute the first task-failure probability, $\eta(\alpha_\epsilon,x_f,t_f|x,t)$, which is the probability that a given state-time is the last state-time that an agent can achieve the goal under the optimized policy. Formally, for any given initial $(x,t)$, the probability of task failure at $(x_f,t_f)$ under the policy is given as:

\begin{align*}
    \eta(\alpha_{\epsilon},x_f,t_f|x,t) = \begin{cases}
    Pr(x_f,t_f|x_i,t,\pi) & \text{if } (\sum_{x'}\kappa(x',t_f+1)P_x(x'|x_f,a_{\pi_{xt}},t_f)=0)\\ &\quad\quad\quad\land (\kappa(x_f,t_f)>0)\\
    0 & \text{otherwise}
    \end{cases},
\end{align*}
where,
\begin{align*}
    Pr(x_f,t_f|x_i,t,\pi) =\sum_{\mathclap{x_{t},...,x_{t_f}}}P_x(x_{t_f}|x_{t_f-1},\pi(x_{t_f-1},t_f-1),t_f-1)...P_x(x_{t+1}|x_{t},\pi(x_{t},t),t),
\end{align*}

Computing the probability $Pr(x_f,t_f|x_i,t,\pi)$ like this requires forward unrolling of the controlled transition dynamics, however the situation is greatly simplified under the assumption of deterministic dynamics.

\subsubsection*{Task-Failure State-time Feasibility Function under Deterministic Transition Operator}

In the case that $P_x$ is a deterministic operator, the policy will only produce one trajectory for any given final time $t_f$ under the availability probability $f_{\dg}(x,a,t_f)$, $$\mathbf{x}_{\pi}=(x_{t},x_{t+1},...,x_{t_f^+}).$$ Given that the cumulative feasibility function has the information of the final time that we can possibly achieve the task under the policy, we automatically have a state-prediction model of the task-failure event in $\mathbf{x}_{\pi}$.  This means that the final state of task-failure must either be at the goal state if the task has any positive probability of being achievable, or the cumulative feasibility function evaluates to zero and the task is impossible (meaning the policy is degenerate and will not be used). Formally, we can compute the failure state-time feasibility as:
\begin{align*}
    \eta^{**}_{\pi_\dg}(\alpha_{\epsilon},x_{f},t_f|x_{i},t_s,\pi)=1-\kappa^{*}_{\dg}(x_{i},t_s), \quad \forall(x_i,t_s)~~ s.t.~~ \kappa^{*}_{\pi_\dg}(x_{i},t_s)>0.
\end{align*}
where $t_f^+$ is the last possible final time the task will be achieved under the policy, and $x_g$ is the single goal state associated with the availability function $f_{\dg}$.  


\subsubsection{Transition Operator over Termination Events}

The termination conditions for the policy are given by the goal success and failure events we previously mentioned, where success is achieved by inducing the goal variable, and the failure event is defined as the first time that the goal can no longer be achieved. When we say that a policy \textit{terminates} at state-time $(x,t)$, we mean that this is the last state-time the policy will be called.  We could use the convention that policy termination refers to the first time that the policy is \textit{not} used (which would be one time-step later than the previous definition), however with the transition operator interpretation of the feasibility function, it is more natural to use the former definition in order to say that the state-time feasibility function \textit{maps to} the final (terminating) state-time with probability $p$ (note that the operator $G_s$ in the main text is the one-step evolution of the feasibility function, and therefore maps to the first state-time \textit{after} termination).  Therefore, because 
\begin{align*}
    &\sum_{x_f,t_f}\eta(\alpha_\dg,x_f,t_f|x,t)=\kappa_{\dg}(x,t),\\
    \text{and}\quad &\sum_{x_f,t_f}\eta(\alpha_{\epsilon},x_f,t_f|x,t)=1-\kappa_{\dg}(x,t),
\end{align*}
then the summation over goals, states, and times is,
\begin{align*}
    \sum_{\alpha,x_f,t_f}\eta_{\pi_g}(\alpha,x_f,t_f|x,t)=1-\kappa_{\dg}(x,t)+\kappa_{\dg}(x,t)=1.
\end{align*}
Therefore the operator $\eta_{\pi_g}$ is a distribution over termination events.

When we have an ensemble of state-time feasibility functions $\Pi_u = \{\pi_{e_j,g_1},...,\pi_{e_m,g_n}\}$ with the ensemble of policies, $\mc H_u = \{\eta_{\pi_{e_1,g_1}},...,\eta_{\pi_{e_m,g_n}}\}$, then because each individual $\eta_{\pi}$ is a \textit{transition distribution}, we can aggregate these feasibility functions into $\widehat{\eta}$ which is an aggregate \textit{transition operator} where $\pi$ is an input to the function.  That is, the aggregate state-time feasibility function is an operator $\widehat{\eta}: (\mc X \times \mc T \times \Pi)\times (\mc X \times \mc T \times \{\alpha_\dg,\dg_-\})\rightarrow [0,1]$, defined as,
\begin{align*}
    \widehat{\eta}_u(\alpha,x_f,t_f|x,t,\pi_{e_j,g_i}):=\eta_{\pi_{e_j,g_i}}(\alpha,x_f,t_f|x,t),\quad \forall \eta_{\pi_{e_j,g_i}}\in \mc H_u.
\end{align*}

\newtcbtheorem[number within=section]{mytheo2}{Theorem}%
{colback=green!5,colframe=green!35!black,fonttitle=\bfseries}{th}

\subsection{Proof of Bidrectional TG-CMDP STFF Decomposition}\label{appx:bidirectional_proof}

\textit{Proof: } Remark: notice that this theorem gives an equivalence for the condition $\text{if: } t_d\leq\min_{\mathbf{t}_j\in \mc T_r} \mathbf{t}_j(\br_j),$ which is saying the time that it takes to complete the goal is less than the time that it takes for the environment variable to transition from $e$ to $e'$ under $P_r$. In this proof we will point out where we enforce this condition.

For compactness, let $e_\br \gets \zeta(\br)$. Recall, that at the horizon time boundary $\bar{\kappa}_{\dg}(\br,x,T_f)=\max_a\bar{f}_{\dg}(x,a,T_f)$.  Therefore, we can also define a new function without $\br$ as a parameter (no bar) $\kappa(x,T_f)=\max_a\bar{f}_{\dg}(x,a,T_f)$ which also equals the quantity $\bar{\kappa}_{\dg}(\br,x,T_f)$ at the boundary, and we can substitute it into the hierarchical OBE (line 2):
Let $\tau = T_f-1$
\begin{align}
    \bar{\kappa}_{\dg}^*&(\br,x,\tau)\label{eq:firstsub}\\
    &= \nonumber\max_{a} \Bigg[\bar{f}_\dg(x,a,\tau) + (1-\bar{f}_\dg(x,a,\tau)) \sum_{x'} P_s(\br',x'|x,a,e_\br,\tau)\bar{\kappa}_{\dg}^*(\br',x',T_f) \Bigg],\\
    &=\max_{a} \Bigg[\bar{f}_\dg(x,a,\tau) \\ 
    &\nonumber+ (1-\bar{f}_\dg(x,a,\tau)) \sum_{x',\br',\ba} P(\br'|\br,\ba)F(\ba|x,a,\tau)P_x(x'|x,a,e_\br,\tau)\kappa_{\dg}^*(x',T_f) \Bigg],\\
    &=\max_{a} \Bigg[\bar{f}_\dg(x,a,\tau) + (1-\bar{f}_\dg(x,a,\tau)) \sum_{x'} P_x(x'|x,a,e_\br,\tau)\kappa_{\dg}^*(x',T_f) \Bigg],\\
    \bar{\kappa}_{\dg}^*&(\br,x,\tau) =\kappa_{\dg}^*(x,\tau).
\end{align}
We can repeat these same steps iteratively backwards through time for each $\tau<T_f-1$, substituting the $\kappa(x,\tau+1)$ from the previous step in for $\bar{\kappa}(\br,x,\tau+1)$ on the R.H.S. of equation \eqref{eq:firstsub}, to establish that $\bar{\kappa}_{\dg}^*(\br,x,t) = \kappa_{\dg}^*(x,t)$ for all $t\in \mc T$.

We can then repeat the exact same steps to obtain to the optimal policy $\bar{\pi}^{**}(\br,x,t)=\pi^{**}(x,t)$ for all $\br$ (not shown). Given that the policy is established by the same steps, and the STFF equation is a function of the policy, we now address the equivalency of the product-space STFF and the factorization in the stated theorem. 

The horizon is defined as $\eta(\ba_\dg,\br_f,x_f,t_f|\br_f,x_f,t_f)=\bar{f}_{\dg}(\ba_\dg|x,\pi^{**}(x,t),t)$. Again, letting $\tau = T_f-1$, we can start by substituting the product-space $\bar{\eta}$ with the $\eta$ restricted to $mc X$ and the variable $e$ at the horizon:
\begin{align}
    \bar{\eta}_{\pi}^{**}&(\ba_\dg,\br_f,x_f,t_f|\br,x,\tau)\label{start} \\ \nonumber&= (1-\bar{f}_\dg(x,a,\tau)) \sum_{x',\br'} P_s(\br',x'|\br,x,a^{**}_{\pi},\tau)\bar{\eta}_{\pi}^{**}(\ba_\dg,\br_f,x_f,t_f|\br',x',\tau+1),\\
    \bar{\eta}_{\pi}^{**}&(\ba_\dg,\br_f,x_f,t_f|\br,x,\tau)\\ 
    \nonumber&=(1-\bar{f}_\dg(x,a,\tau)) \sum_{x',\br'} P_s(\br',x'|\br,x,a^{**}_{\pi},\tau)\eta_{\pi}^{**}(\ba_\dg,x_f,t_f|x',\tau+1),
\end{align}
Substituting $ P_s(\br',x'|\br,x,a,t) :=\sum_{\ba}P_r(\br'|\br,\ba)F(\ba|x,a,t)P_x(x'|x,a,t,e_\br)$ and pulling out the summations over $\ba$, we have:
\begin{align}
    &\bar{\eta}_{\pi}^{**}(\ba_\dg,\br_f,x_f,t_f|\br,x,\tau)= \left(\sum_{\br',\ba}P_r(\br'|\br_\tau,\ba)F(\ba|x,a,\tau)\right)\\
    &\nonumber \quad\quad*(1-\bar{f}_{\dg}(x,a,\tau)) \sum_{x'} P_x(x'|x,a^{**}_{\pi},\tau,e_\br)\eta_{\pi}^{**}(\ba_\dg,x_f,t_f|x',\tau+1),\\
    &\bar{\eta}_{\pi}^{**}(\ba_\dg,\br_f,x_f,t_f|\br,x,\tau)\label{markov-0}\\
    &\nonumber\quad\quad= P_{\br,\epsilon}(\br_f|\br_\tau)\left((1-\bar{f}_{\dg}(x,a,\tau)) \sum_{x'} P_x(x'|x,a^{**}_{\pi},e_\br,\tau)\eta_{\pi}^{**}(\ba_\dg,x_f,t_f|x',\tau+1)\right).
\end{align}
Let $e_{\br,f}$ be the environment variable at $\tau=T_f-1$, and $P^{e_{\br,f}}_x$ be $P_x$ conditioned by $e_{\br,f}$:
\begin{align}
    &\bar{\eta}_{\pi}^{**}(\ba_\dg,\br_f,x_f,t_f|\br,x,\tau)\label{markov-1}\\
    &\nonumber\quad\quad= P_{\br,\epsilon}(\br_f|\br_\tau)\underbrace{\left((1-\bar{f}_{\dg}(x,a,\tau)) \sum_{x'} P_x^{e_{\br,f}}(x'|x,a^{**}_{\pi},\tau)\eta_{\pi}^{**}(\ba_\dg,x_f,t_f|x',\tau+1)\right)}_{=~\eta_{e_{\br,f}}^{**}(\ba_\dg,x_f,t_f,|x,\tau)}.
\end{align}
The term in the large parenthesis of equation \eqref{markov-1} is the definition of $\eta_{e_{\br,f}}^{**}$ (which is an STFF computed only on $P^{e_{\br,f}}_x$ and $\bar{f}_{\dg}$) starting from $\tau$ using the same goal-availability function. Substituting it in we have:
\begin{align}
    &\bar{\eta}_{\pi}^{**}(\ba_\dg,\br_f,x_f,t_f|\br,x,\tau) = P_{\br,\epsilon}(\br_f|\br_\tau)\eta_{e_{\br,f}}^{**}(\ba_\dg,x_f,t_f,|x,\tau)\label{last-eq}.
\end{align}
Note that $\tau=T_f-1$ and so $\br' = \br_f$. Also note that equation $\eqref{markov-0}$ replaces $P_r(\br'|\br_\tau,\ba)F(\ba|x,a,t)$ with the transition operator $P_{\br,\epsilon}$ fixed to the input $\ba_\epsilon$ due to the homogeneity assumption (the variables $(x,a,t)$ from $F$ are independent and can be dropped). 

As we repeat these steps \eqref{start} to \eqref{last-eq} backwards in time from $\tau=T_f-2$ to $\tau=0$, we can continue to hold the operator $P^{e_{\br,f}}_x$ as a constant across all steps---this is the enforcement of the single environment variable $e_{\br,f}$ mentioned at opening remark. We can ask: under what conditions will enforcing this variable ($e_{\br,f}$) result in the true equivalency expressed in equation $\eqref{last-eq}$? The answer is that if the true environment variable for the time step $0 \leq \tau < T_f-1$ is $e_{\br,\tau}$, and $e_{\br,\tau} = e_{\br,f}$, the equation must be true.  

If we work backwards in time, we can ask the important question: when is the first time $\tau$ that the expression is \textit{not} true? This must be the \textit{first} time that $e_{\br,f}$ is different for any given $e_{\br,\tau}$. Since $e$ is directly determined by $\br$ via $e_\br\leftarrow\zeta(\br)$, we can compute the first time that an $\br$-vector has an element that hits a mode that induces a different $e$ than the final $e_f$.  Since, $$P_{\br,\epsilon}(\br'|\br) = P_{w,e}(w'|w)P_{y,e}(y'|y)P_{z,e}(z'|z),\quad \br=(w,y,z),\br'=(w',y',z')$$ is a Markov chain (considering only three spaces for $\mc R$, w.l.o.g.), we can compute the first hitting time of each one, taking $y$ as an example \cite{bremaud2013markov}:
\begin{align}
    \mathbf{t}_y:=(I-\bar{P}_y)^{-1}\mathbf{1},
\end{align}
where $\bar{P}_y$ is the Markov matrix with the rows and columns corresponding to elements of $y$ which would map to an environment variable other than $e_{\br,f}$. 
Recall, $\zeta$ is element-invariant, so the first time any element $v$ of $\br=(v,...)$ transitions out of the mode-restriction set $\mc V_{e_{\br,f}}$ is the first time that $e$ will transition. Therefore, if we compute a set of time-to-go vectors $\mc T_r=(\mathbf{t}_w,\mathbf{t}_y,\mathbf{t}_z,...)$, then the first time that the mode parameter transitions from $e\rightarrow e'$ (where $e\neq e'$) when starting from a given $\br_j$ is:
\begin{align*}
    \tau-t_f\leq t_{min}=\min_{\mathbf{t}_j\in \mc T_r} \mathbf{t}_j(\br_j)
\end{align*}
Therefore, for any given $\tau$ as we are stepping back time, if this inequality is true, equation \eqref{last-eq} holds: 
\begin{align}
    \bar{\eta}_{\pi}^{**}(\ba_\dg,\br_f,x_f,t_f|\br,x,\tau) &= \\
    \nonumber\Bigg(\sum_{\br_{f-1}}...&\sum_{\br_{\tau'}}P_{\br,\epsilon}(\br_f|\br_{f-1})...P_{\br,\epsilon}(\br_{\tau'}|\br_\tau)\Bigg)\eta_{e_{\br,f}}^{**}(\ba_\dg,x_f,t_f|x,\tau),
\end{align}
which can be written as:
\begin{align}
    \bar{\eta}_{\pi}^{**}(\ba_\dg,\br_f,x_f,t_f|\br,x,\tau) = P_{\br,\epsilon}^{t_f-\tau}(\br_f,\br_\tau)\eta_{e_{\br,f}}^{**}(\ba_\dg,x_f,t_f|x,\tau),\\
    \bar{\eta}_{\pi}^{**}(\ba_\dg,\br_f,x_f,t_f|\br,x,\tau) = \omega_r(\br_f|\br_\tau,t_f-\tau)\eta_{e_{\br,f}}^{**}(\ba_\dg,x_f,t_f|x,\tau),
\end{align}
where $P_{\br,\epsilon}^{t_f-\tau}(\br_f,\br_\tau)$ is the element $(\br_f,\br_\tau)$ of the Markov chain matrix $P_{\br,\epsilon}$ taken to the power $(t_f-\tau)$ (the probability of starting at $\br_\tau$ and ending up at $\br_f$ after $t_f-\tau$ time steps), and $\omega_r(\br_f|\br_\tau,t_f-\tau):=P_{\br,\epsilon}^{t_f-\tau}(\br_f,\br_\tau)$, where,
\begin{align}
    P_{\br,\epsilon}^{t_f-\tau}(\br_f,\br_\tau) = P_{w,e}^{t_f-\tau}(w_f|w_{\tau})P_{y,e}^{t_f-\tau}(y_f|y_{\tau})P_{z,e}^{t_f-\tau}(z_f|z_{\tau}),
\end{align}
due to the independence of the transition dynamics in the definition of $P_{r}$.

Since the entire proof can be done with for the goal-success distribution, one can apply the exact same steps to the goal-failure distribution (not shown), resulting in: $$\bar{\eta}_-(\bar{\ba}_-,\br_-,x_-,t_-|\br,x,t)=\omega_r(\br_-|\br,t_- -t)\eta_{e_{\br,f}}(\bar{\ba}_-,x_-,t_-|x,t).$$
This concludes the proof. $\hfill \square$

A couple things to note: This proof assumes $P_\br$ is deterministic, so the above calculation is for the average hitting time, which must also be the exact hitting time.  If one were to prove this for stochastic $P_\br$, one would have to compute the joint probability that any of the components change the environment variable, which would involve forward-evolving each the Markov chain dynamics. One would have to compute this probability for all combinations of high-level states, which would be possible, but computationally expensive.  Furthermore, if $\zeta$ were not element-invariant, then we would have to compute the hitting time and probabilities on the full Markov chain $\bar{P}_r$:
\begin{align*}
    \mathbf{t}_r:=(I-\bar{P}_r)^{-1}\mathbf{1},
\end{align*}
which is computationally prohibitive, unless alternative techniques were possible to be developed.  It is also worth noting that this proof assumed that the high-level states were deterministic.  A similar proof could potentially be given assuming stochasticity if one computes the probability of arriving at a specific state before a given time under the controlled Markov dynamics, but would likely not have a simple solution (e.g. in the form of a linear system) and would involve forward-evolving the probability distributions under the Markov chain.

\subsection{Marginal Empowerment}\label{appx.marg_emp}

Marginal empowerment $\mathfrak{E}_n^x(P_s|x_t)$ can be defined as the empowerment of a particular state random variable in a product space (here $X$, denoted in the superscript by $x$, which corresponds to $\mc X$ of the larger product space $\mc S$), by marginalizing out the other state-spaces.  For instance, if $P_s(x',y',z'|x,y,z,a)$ is our transition operator, then 
\begin{align}
    \mathfrak{E}_n^x(P_s|s_t) =\max_{p(\mathbf{a}_t^{\tau-1}|s_t)}I(A^{\tau-1}_t;X_\tau|s_t)=\max_{p(\mathbf{a}_t^{\tau-1}|s_t)}\left[H(X_\tau|s_t)- H(X_\tau|A^{\tau-1}_t,s_t)\right],
\end{align}

is the marginalized empowerment, where we do not care about the entire final state random variable $S_\tau$, rather we only compute mutual information with respect to $X_{\tau}$.  That means that for a given $\mathbf{a}_t^{\tau-1}=(a_{t},...,a_{\tau-1})$, the random variable $X_\tau$ has a distribution for the marginalized operator $P(x_\tau|\mathbf{s}_{t},\mathbf{a}_t^{\tau-1})$, given as:

\begin{align}
    P(x_\tau|\mathbf{s}_{t},\mathbf{a}_t^{\tau-1}) = \sum_{\mathclap{y_{\tau},z_{\tau},...,x_{t+1},y_{t+1},z_{t+1}}}P_s(x_{\tau},y_{\tau},z_{\tau}|x_{\tau-1},y_{\tau-1},z_{\tau-1},a_{\tau-1})...P_s(x_{t+1},y_{t+1},z_{t+1}|x_t,y_t,z_{t},a_{t}).
\end{align}

Again, when $P_s$ is deterministic, we can compute the marginal empowerment by counting the set of reachable states on $\mc X$ and taking the logarithm.

\subsection{Deterministic Transition Operator Empowerment and Forward Diffusion}\label{appx:det_emp}

This proof is done on state-time transition operators, but it also applies to stationary transition operators as a special case.




\begin{mytheo2}{Deterministic Empowerment by Forward Diffusion}{theoexample}
Let $P_{st}: (\mc S \times \mc T \times \mc A) \times (\mc S \times \mc T) \rightarrow \{0,1\}$ by a deterministic transition operator.  The $n$-step empowerment from a state-time $(s,t)$ is $$\mathfrak{E}_n(P_{st}|\mathbf{s},t)=\log_2(||\lceil \vec{v}_{\tau} \rceil ||_{1}),$$ where $||\cdot||_{1}$ is the 1-norm and $\vec{v}_\tau$ is the probability vector over $\mc S \times \mc T$ after a forward $n$-step ($n=\tau-t$) diffusion starting from $(s,t)$, with a uniform distribution over actions.
\end{mytheo2}

\textit{\textbf{Proof}: }Let $\mc C = \mc S \times \mc T$ be a set of state-times.  We redefine $P_{st}$ for convenience as $P: (\mc C \times \mc A) \times \mc C \rightarrow \{0,1\}$. Since
$$\mathfrak{E}_n(P|c) =\max_{p(A)_t^{\tau-1}|c}\left[H(C_\tau|c)- H(C_\tau|A^{\tau-1}_t,c)\right],$$ where $p$ is the distribution over action sequences, determinism in $P$ implies that the conditional entropy $H(S_\tau|A^{\tau-1}_t,c)$ equals $0$, and therefore empowerment is maximized at
\begin{align}
    \max_{p(c_\tau)}H(C_\tau|c)=\log_2(\mc C_{n})
\end{align}
where $\mc C_{n}$ is the set of possible reachable states under open-loop plans of length $n$, therefore, one only needs to compute $\mc C_{n}$.  If we let $\pi(a|c)$ be a uniform distribution over actions for every state $\mathbf{s}$, $v_{t_0}$ be a one-hot vector over $\mc S \times \mc T$ with a value of $1$ for $(c,t)$, and let $P_{\pi}(c'|c) = \sum_{a}P(c'|c,a)\pi(a|c)$ (where $P_{\pi}$ is the acting as the state-time Markov matrix), then we write the vector update after one forward step as:
\begin{align}
     v_{t_0}' P_{\pi}=v_{t_1}',
\end{align}
A non-zero entry in $v_{t_1}$ implies the existence of an action which can be used to deterministically reach the state-time associated with the entry with a probability of $1$. By extension,
\begin{align}
     v_{t_0}' P_{\pi}^{n}=v_{t_n}',
\end{align}
implies the existence of an $n$-length action sequence that can reach any state state with non-zero probability in $v_{t_n}$.  As such, the number of possible reachable states under open-loop policies is,
\begin{align}
    \mc C_n=||\lceil v_{t_n} \rceil ||_{1},
\end{align}
which is the count of states with a probability greater than $0$. Therefore, $$\mathfrak{E}_n(P|\mathbf{s},t)=\log_2(||\lceil v_{t_n} \rceil ||_{1}).$$

$\hfill \square$

\subsection{Valence Bellman Equation}\label{appx:valence-BE}

Here we show that if we define reward as valence, then the finite horizon Bellman equation simply optimizes a closed-loop policy which controls to the state with the highest empowerment value (in expectation). Starting with the finite-horizon Bellman equation with horizon $t_f$, we substitute in $\mathfrak{V}(P_x,P_x,P_x,x_t,t,a)$ in place of a reward function:
\begin{align*}
    \nu(x,t)=\max_{a}\left[\mathfrak{V}(P_x,P_x,P_x,x_t,t,a)+ \sum_{x'}P_x(x'|x,a)\nu(x',t+1)\right].
\end{align*}

We drop the empowerment horizon $n$ for generality. Note that we are using $P_x$ for the first and second empowerment evaluation and also using it for the \textit{plan distribution} $Q$ over the next states, i.e. $Q=P_x$.

If we optimize the Bellman equation to produce $v^*$ and $\pi^*$ then, by substituting the definition of reward, our optimal value function is defined as:
\begin{align*}
    \nu_{\pi}^*(x,t)=\mathfrak{V}_{x_t a_{xt}^*} + \expec_{x'\sim P_x(\cdot|x,a_{xt}^*)}\left[\nu_{\pi}^*(x',t+1)\right],
\end{align*}
where $a_{xt}^*=\pi^*(x,t)$. Expanding the recursive form into its series representation, we have:
\begin{align*}
    \nu_{\pi}^*(x_{t_0},t_0) = \mathfrak{V}_{x_{t_0} a_{xt_0}^*} + \expec_{x_{t_1}} \left[\mathfrak{V}_{x_{t_1} a_{xt_1}^*}\right] + \expec_{x_{t_2}}\expec_{x_{t_1}} \left[\mathfrak{V}_{x_{t_2} a_{xt_2}^*}\right] + ... + \expec_{x_{t_f}}...\expec_{x_{t_1}} \left[\mathfrak{V}_{x_{t_f} a_{xt_{f}}^*}\right].
\end{align*}

The definition of valence is,
\begin{align*}
    \mathfrak{V}(P_x,P_x,P_x,x_t,a_t) = \expec_{x_{t+1}\sim P_x(\cdot|x_t,a,t)}\left[\mathfrak{E}_{x_{t+1}}\right]-\mathfrak{E}_{x_t},
\end{align*}
where $\mathfrak{E}_{x_t} \equiv \mathfrak{E}(P_x|x_t)$. Substituting this into the series we have:
\begin{align*}
    &\nu_{\pi}^*(x_{t_0},t_0) = \left(\cancel{\expec_{x_{t_1}}\left[ \mathfrak{E}_{x_{t_1}}\right]}-\mathfrak{E}_{x_{t_0}}\right) + \left( \cancel{\expec_{x_{t_2}}\expec_{x_{t_1}} \left[\mathfrak{E}_{x_{t_2}}\right]}-\cancel{\expec_{x_{t_1}}\left[\mathfrak{E}_{x_{t_1}}\right]} \right)\\
    &\quad+ \left( \cancel{\expec_{x_{t_3}}\expec_{x_{t_2}}\expec_{x_{t_1}} \left[\mathfrak{E}_{x_{t_3}}\right]}-\cancel{\expec_{x_{t_2}}\expec_{x_{t_1}}\left[\mathfrak{E}_{x_{t_2}}\right]}\right) +...+ \left( \expec_{x_{t_f}}...\expec_{x_{t_1}} \left[\mathfrak{E}_{x_{t_f}}\right]-\cancel{\expec_{x_{t_f-1}}...\expec_{x_{t_1}}\left[\mathfrak{E}_{x_{t_f-1}}\right]}\right)
\end{align*}

Notice that for every intermediate empowerment evaluation there is a negative quantity for the current time and a corresponding positive quantity of the same magnitude for the subsequent time, so each intermediate empowerment evaluation cancels out, leaving us with:
\begin{align*}
    \nu_{\pi}^*(x_{t_0},t_0) &= \expec_{x_{t_f}}...\expec_{x_{t_1}} \left[\mathfrak{E}_{x_{t_f}}\right] -\mathfrak{E}_{x_{t_0}} \\
    &= \expec_{x_{t_f}\sim Q_{\pi}}\left[\mathfrak{E}_{x_{t_f}}\right] -\mathfrak{E}_{x_{t_0}}\\ &=\mathfrak{V}(P_x,P_x,Q_{\pi},x_{t_0},t_0,\pi^*),
\end{align*}
where $Q_{\pi}(x_{t_f},t_f|x,t,\pi^*) = \expec_{x_{t_f-1}}...\expec_{x_{t_1}} P_x(x_{t_f}|x_{t_f-1},\pi^*(x,t_f-1))$ is the distribution over final states, with each expectation defined as $\expec_{x_{\tau}\sim P_{x}(\cdot|x_{\tau-1},a_{\pi_{\tau-1}})}$. This means the value function is simply the expected valence of the final state $x_{t_f}$ after following the policy for $t_f$ time-steps.  As we can see, optimizing with respect to empowerment gain is equivalent to optimizing to reach the states with the highest empowerment value (in expectation over final states). 
\newpage
\subsection{Sublimation Inequality Proof}\label{appx:sublimation}

\begin{mytheo}{Sublimation Theorem}{theoexample}
If $\bar{\mathscr{M}}=(\Sigma \times \mc X, \mc A, P_s,\bar{f}_{\dg})$ is a full hierarchical TGMDP, and $\mathscr{M}_{sub}=(\Sigma, \mc A^{1}, P_\sigma,\bar{f}_{\dg,sub})$ is a sublimated TG-CMDP using $P_\sigma$ on space $\Sigma$ derived from $\bar{\mathscr{M}}$, where $\bar{f}_{\dg,sub}(\boldsymbol{\sigma},t) := \max_{x,a} \bar{f}_{\dg}(\boldsymbol{\sigma},x,a,t)$, then:
$$\bar{\kappa}(\boldsymbol{\sigma},x,t)\leq \kappa_{sub}(\boldsymbol{\sigma},t),$$
where $\bar{\kappa}$ is the full hierarchical cumulative feasibility function and $\kappa_{sub}$ is the sublimated cumulative feasibility function.
\end{mytheo}
\textbf{Proof}:

For notational compactness let $\bar{f}_{\delta} = 1-\bar{f}_{\dg}$.  We start the proof by proving the inequality at the time horizon $T_f$, where the condition $\bar{\kappa}(\boldsymbol{\sigma},x,T_f) = \bar{f}_\dg(\boldsymbol{\sigma}_k,x_i,T_f)$ and $\kappa_{sub}(\boldsymbol{\sigma},x,T_f) = \bar{f}_{\dg,sub}(\boldsymbol{\sigma},T_f)$ holds. Also, following from the definition of $\bar{f}_{\dg,sub}$, it must be the case that, $\bar{f}_\dg(\boldsymbol{\sigma},x,t)\leq \bar{f}_{\dg,sub}(\boldsymbol{\sigma},t),$ and therefore the horizon feasibility for the two problems must be, $\bar{\kappa}(\boldsymbol{\sigma},x,T_f)\leq \kappa_{sub}(\boldsymbol{\sigma},T_f).$

Starting with the definition of the OBE, and letting $\tau = T_f-1$:
\begin{align}
    &\bar{\kappa}^*(\boldsymbol{\sigma}_k,x_i,\tau)\label{eq:sublime-0-TG-MDP}\\
    \nonumber&= \max_{a} \left[\bar{f}_\dg(\boldsymbol{\sigma}_k,x_i,\tau) + \bar{f}_{\delta}(\boldsymbol{\sigma}_k,x_i,\tau) \sum_{\boldsymbol{\sigma}',x'} P_{\boldsymbol{\sigma} x}(\boldsymbol{\sigma}',x'|\boldsymbol{\sigma}_k,x_i,a,\tau)\bar{\kappa}^*(\boldsymbol{\sigma}',x',T_f) \right],\\
   &\leq \max_{a} \left[\bar{f}_\dg(\boldsymbol{\sigma}_k,x_i,\tau) + \bar{f}_{\delta}(\boldsymbol{\sigma}_k,x_i,\tau) \sum_{\boldsymbol{\sigma}',x'} P_{\boldsymbol{\sigma} x}(\boldsymbol{\sigma}',x'|\boldsymbol{\sigma}_k,x_i,a,\tau)\bar{\kappa}^*_{sub}(\boldsymbol{\sigma}',T_f) \right],\label{eq:sublime-1-TG-MDP}\\
   &\leq \max_{a} \left[\bar{f}_{\dg,sub}(\boldsymbol{\sigma}_k,\tau) + \bar{f}_{\delta,sub}(\boldsymbol{\sigma}_k,\tau) \sum_{\boldsymbol{\sigma}',x'} P_{\boldsymbol{\sigma} x}(\boldsymbol{\sigma}',x'|\boldsymbol{\sigma}_k,x_i,a,\tau)\bar{\kappa}^*_{sub}(\boldsymbol{\sigma}',T_f) \right],\label{eq:sublime-2-TG-MDP}\\
   &=\max_{a} \left[\bar{f}_{\dg,sub}(\boldsymbol{\sigma}_k,\tau) + \bar{f}_{\delta,sub}(\boldsymbol{\sigma}_k,\tau) \sum_{\mathclap{\alpha,\boldsymbol{\sigma}',x'}}P_\sigma(\boldsymbol{\sigma}'|\boldsymbol{\sigma},\alpha)F(\alpha|x_i,a,\tau)P_x(x'|x_i,\tau,a)\bar{\kappa}^*_{sub}(\boldsymbol{\sigma}',T_f) \right],\label{eq:sublime-3-TG-MDP}\\
   &=\max_{a} \left[\bar{f}_{\dg,sub}(\boldsymbol{\sigma}_k,\tau) + \bar{f}_{\delta,sub}(\boldsymbol{\sigma}_k,\tau) \sum_{\alpha,\boldsymbol{\sigma}'}P_\sigma(\boldsymbol{\sigma}'|\boldsymbol{\sigma},\alpha)F(\alpha|x_i,a,\tau)\bar{\kappa}^*_{sub}(\boldsymbol{\sigma}',T_f) \right],\\
   &= \bar{f}_{\dg,sub}(\boldsymbol{\sigma}_k,\tau) + \bar{f}_{\delta,sub}(\boldsymbol{\sigma}_k,\tau) \sum_{\alpha,\boldsymbol{\sigma}'}P_\sigma(\boldsymbol{\sigma}'|\boldsymbol{\sigma},\alpha)F(\alpha|x_i,\pi^{**}(x_i,\tau),\tau)
   \kappa^*(\boldsymbol{\sigma}',T_f),\label{eq:sublime-4-TG-MDP}\\
   &\leq\max_{\alpha} \left[\bar{f}_{\dg,sub}(\boldsymbol{\sigma}_k,\tau) + \bar{f}_{\delta,sub}(\boldsymbol{\sigma}_k,\tau) \sum_{\boldsymbol{\sigma}'} P_\sigma(\boldsymbol{\sigma}'|\boldsymbol{\sigma}_k,\alpha)\bar{\kappa}^*_{sub}(\boldsymbol{\sigma}',T_f) \right],\label{eq:sublime-4.5-TG-MDP}\\
   &=\bar{\kappa}^*_{sub}(\boldsymbol{\sigma}_k,\tau)\\
   &\implies \bar{\kappa}^*(\boldsymbol{\sigma}_k,x_i,\tau) \leq \bar{\kappa}^*_{sub}(\boldsymbol{\sigma}_k,\tau).
\end{align}

Because this condition holds at the time $\tau=T_f-1$, then by backwards induction we can apply the same steps for each previous time-step, and the cumulative feasibility function for any time $t<T_f-1$ must also have the same inequality. Therefore, for every input $(\boldsymbol{\sigma},x,t)$, we have,
\begin{align*}
    \bar{\kappa}^*(\boldsymbol{\sigma},x,t) \leq \bar{\kappa}^*_{sub}(\boldsymbol{\sigma},t),\quad \forall t\in \mc T, \quad \forall x\in \mc X,\quad \forall \boldsymbol{\sigma}\in \Sigma,
\end{align*}

concluding the proof.  $\hfill \square$

Note that this proof applies to all possible sub-state-spaces of the full hierarchical product space.  If $\bar{\mathscr{M}}=(\mc S=\Sigma \times \mc W \times \mc Y \times...\times \mc X, \mc A, P_s,\bar{f}_{\dg})$ is a full hierarchical TG-CMDP, and $\mathscr{M}_{sub}=(\Sigma, \mc A^{1}, P_\sigma,\bar{f}_{\dg,sub})$.  Then $\bar{f}_{\dg,sub}(\boldsymbol{\sigma},t) := \max_{\lnot \Sigma} \bar{f}_{\dg}(\boldsymbol{\sigma},w,y,...,x,a,t)$, where $\max_{\lnot \Sigma}$ maximizes over all state variables comprising $\mc S$ that are not in $\Sigma$, along with the action. If one formalizes a set of sublimated TG-MDPs for each individual space in $\mc S$, then a set of sublimated feasibility functions $\mathscr{K}_{sub}:=\{\bar{\kappa}_{sub,\boldsymbol{\sigma}},\bar{\kappa}_{sub,w},\bar{\kappa}_{sub,y},...,\bar{\kappa}_{sub,x}\}$ can be computed to constrain policy forward sampling.

\newpage

\subsection{Empowerment Pseudocode}
\begin{algorithm2e}[h]
\DontPrintSemicolon
\SetKw{return}{return}
\SetKwRepeat{Do}{do}{while}
\SetKwData{conflict}{conflict}
\SetKwData{safe}{safe}
\SetKwData{sat}{sat}
\SetKwData{unsafe}{unsafe}
\SetKwData{unknown}{unknown}
\SetKwData{true}{true}
\SetKwInOut{Input}{input}
\SetKwInOut{Output}{output}
\SetKwFor{Loop}{Loop}{}{}
\SetKw{KwNot}{not}
\begin{small}
	\Input{$\mc O, \mathbf{s},t,n$}
	\Output{$empwr$}
        $P \leftarrow \mc O.P \quad\#~~ P:(\mc S \times \mc T \times \mc A) \times (\mc S \times \mc T)\rightarrow [0,1]$\;
	define $ns$ as $|\mc S|$\;
	define $nt$ as $|\mc T|$\;
	define $\pi(a|s)$ as a uniform distribution over $a$ for every $s$\;
	$P_{\pi}\leftarrow \sum_a P(:,:,a,:,:)\pi(a|:)$\;
	$P_{\pi}\leftarrow reshape(P_{\pi},[ns*nt,ns*nt])$\;
    $\#~~ \text{Forward Diffusion for unfactorized transition operator}$\;
	\uIf{$\texttt{is\_deterministic}(P)\land is\_product\_space=\texttt{False}$}{
	    $v \leftarrow \texttt{one\_hot}(\texttt{linear\_index}(s,t),ns*nt)$\;
	    \For{$\tau = 0:n-1$}{
	        $v\leftarrow \texttt{dot}(v, P_{\pi})$\;
	    }
	    $empwr \leftarrow \log_2(||\lceil v \rceil||_1)$\;
	}
        \uElseIf{$\texttt{is\_deterministic}(P)\land is\_product\_space=\texttt{True}$ }{          $empwr \leftarrow\texttt{det\_factorized\_empowerment}(\mc O,(s,t),n)$
        }
	\Else{
	$empwr \leftarrow \texttt{blahut\_arimoto\_empowerment}(P,(s,t),n)$
	}
	
    \texttt{return~} $empwr$\;
\end{small}
\caption{Empowerment}
\label{alg:empowerment}
\end{algorithm2e}

\subsection{Deterministic Empowerment (Factorized) Pseudocode}\label{sec:det-emp-product}

This empowerment algorithm is for deterministic factorized product-space transition operators, and simply counts up the number of unique reachable state-times for a horizon of $n$.

\begin{algorithm2e}[h]
\DontPrintSemicolon
\SetKw{return}{return}
\SetKwRepeat{Do}{do}{while}
\SetKwData{conflict}{conflict}
\SetKwData{safe}{safe}
\SetKwData{sat}{sat}
\SetKwData{unsafe}{unsafe}
\SetKwData{unknown}{unknown}
\SetKwData{true}{true}
\SetKwInOut{Input}{input}
\SetKwInOut{Output}{output}
\SetKwFor{Loop}{Loop}{}{}
\SetKw{KwNot}{not}
\begin{small}
	\Input{\text{Low-level Operators}: $\mathscr{P} = \{P_x,P_y,P_z,...\},\widehat{\eta}_c,\Omega = \{\omega_y,\omega_z,...\},\text{state-vectors}: \mathbf{s}=(\mathbf{x},\mathbf{y},\mathbf{z},...),t,n$}
	\Output{$empwr$}
        $queue \leftarrow empty\_queue()$\;
        \For{$k ~~\text{from}~~ 0 ~~\text{to}~~n$}{
                $(\mathbf{s},t)\leftarrow queue.pop()$\;
                \For{$\pi\in\Pi$}{
                    $(\mathbf{s}',t')\leftarrow (\mathbf{x}',\mathbf{y}',\mathbf{z}',...,t')\leftarrow \texttt{advance\_state}(\mathscr{P},\widehat{\eta},\Omega,(\mathbf{x},\mathbf{y},\mathbf{z},...),t,\pi)$\;
                    \If{$(\mathbf{s}',t')\not\in queue$}{
                        $queue.push((\mathbf{s}',t'))$
                    }
                }
        }
        $empwr \leftarrow \log_2(queue.size)$\;

    \texttt{return~} $empwr$\;
\end{small}
\caption{Deterministic Empowerment (Factorized)}
\label{alg:product-space-empowerment}
\end{algorithm2e}

\end{document}